\documentclass{article}

\usepackage{microtype}
\usepackage{graphicx}
\usepackage{booktabs} 
\usepackage[T1]{fontenc}    
\usepackage[caption=false,font=normalsize,labelfont=sf,textfont=sf]{subfig}
\usepackage{stfloats}
\usepackage{multirow}

\usepackage{hyperref}

\usepackage[accepted]{icml2024}

\usepackage{amsmath}
\usepackage{amssymb}
\usepackage{amsfonts}
\usepackage{mathtools}
\usepackage{amsthm}

\usepackage[capitalize,noabbrev]{cleveref}

\theoremstyle{plain}
\newtheorem{theorem}{Theorem}[section]

\newtheorem{lemma}[theorem]{Lemma}

\theoremstyle{definition}
\newtheorem{definition}[theorem]{Definition}

\theoremstyle{remark}
\newtheorem{remark}[theorem]{Remark}

\usepackage[textsize=tiny]{todonotes}

\begin{document}

\twocolumn[
\icmltitle{Exploring Adversarial Attacks against Latent Diffusion Model from the Perspective of Adversarial Transferability}



\icmlsetsymbol{equal}{*}

\begin{icmlauthorlist}
	\icmlauthor{Junxi Chen}{sysu}
	\icmlauthor{Junhao Dong}{ny}
	\icmlauthor{Xiaohua Xie}{sysu}
\end{icmlauthorlist}

\icmlaffiliation{sysu}{School of Computer Science and Engineering, Sun Yat-sen University, Guangzhou, China}
\icmlaffiliation{ny}{School of Computer Science and Engineering, Nanyang Technological University, Singapore}

\icmlcorrespondingauthor{Xiaohua Xie}{xiexiaoh6@mail.sysu.edu.cn}

\icmlkeywords{Machine Learning, ICML}

\vskip 0.3in
]



\printAffiliationsAndNotice{}  

\begin{abstract}
Recently, many studies utilized adversarial examples (AEs) to raise the cost of malicious image editing and copyright violation powered by latent diffusion models (LDMs). Despite their successes, a few have studied the surrogate model they used to generate AEs. In this paper, from the perspective of adversarial transferability, we investigate how the surrogate model's property influences the performance of AEs for LDMs. Specifically, we view the time-step sampling in the Monte-Carlo-based (MC-based) adversarial attack as selecting surrogate models. We find that the smoothness of surrogate models at different time steps differs, and we substantially improve the performance of the MC-based AEs by selecting smoother surrogate models. In the light of the theoretical framework on adversarial transferability in image classification, we also conduct a theoretical analysis to explain why smooth surrogate models can also boost AEs for LDMs.
\end{abstract}

\section{Introduction}
Recently, latent diffusion models (LDMs)~\cite{LDM}, powered by the conditional mechanism and the low-cost diffusion process in low-dimensional latent space, have shown great flexibility in generating high-quality images with user-defined conditions and low computation overhead. However, just as every coin has two sides, the convenience and low budget of using LDMs also reduce the cost of malicious image generation. For example, an \emph{adversary} can mimic the composition of a copyrighted photo by using image variation or can change someone's behavior presented in the picture with the image inpainting.

To prohibit the malicious usage of LDMs, some recent works~\cite{AdvDM, photoGuard, benchmark, mist} turned to adversarial examples (AEs) to protect the image from being edited or serving as a condition for image generation. They either added adversarial perturbations $\delta$ to push the image $x + \delta$ out of the generated distribution $p_{\theta}(x)$ \cite{AdvDM} or disturbed the output \cite{photoGuard} and intermediate latent \cite{mist, benchmark} of LDMs in order to degrade the quality of the generated images. 

In this paper, observing the inevitable discrepancy between the surrogate model used to manufacture AEs and the target model to attack, we model existing adversarial attacks to LDMs as transfer-based adversarial attacks. We investigate how the smoothness of the surrogate model, a significant factor affecting the adversarial transferability, influences the performance of AEs against LDMs. 

Specifically, when generating AEs for LDMs, some adversarial attacks~\cite{AdvDM, mist} use surrogate models different from the target model. Moreover, due to the variety of generation setups (e.g., guidance scale, prompt, and mask), the protected image faces multiple target models even if the adversary only uses one LDM for image editing. Consequently, adversarial transferability, meaning that AEs generated on one model can also crack other models, plays an important role. As a line of works improved the transferability empirically by modifying the surrogate model~\cite{ghostNet,advtrainImprove,LGVtrans} and analyzed the relationship between the surrogate model and the transferability theoretically~\cite{transferTheoryNIPS, transferTheorySP}, studying AEs for LDMs from the perspective of the adversarial  transferability and surrogate model may provide some hints for improving AEs' performance.

Treating the time step as a part of the surrogate model's weight, we view the time-step sampling in the MC-based adversarial attack (e.g., AdvDM)~\cite{AdvDM} as selecting surrogate models for manufacturing AEs. We find that the smoothness of surrogate models at different time steps varies, and such a discrepancy is significant enough to alter the performance of AEs. Empirical experiments demonstrate that limiting the sampling range of time steps to where surrogate models are smooth can substantially improve AdvDM's performance. Moreover, we conduct a theoretical study based on a theoretical framework~\cite{transferTheoryNIPS, transferTheorySP} for adversarial transferability in the image classification task to explain why smoother surrogate models can also boost AEs for LDMs.
Additionally, we find that AEs with good performance in corrupting image inpainting and variation may perform poorly in cracking image generation tasks requiring fine-tuning, e.g., textual inversion~\cite{textualInversion}.

Our contributions are as follows.

\begin{itemize}
	\item From the perspective of adversarial transferability and the smoothness of surrogate models, we conduct a theoretical study and empirical experiments to explain why limiting the sampling range of time steps in MC-based adversarial attacks works.
	
	\item Guided by our analyses and findings, we substantially improve MC-based AEs' performance in cracking image inpainting and variation by simply limiting the sampling range of time steps.
	
	\item We find that AEs possessing good capability in corrupting image inpainting and image variation may not work well in degrading generation tasks requiring fine-tuning. Based on a very recent work~\cite{simac}, we explain why such a phenomenon exists, which reveals the differences and commonalities between AEs for these two types of tasks.
\end{itemize}

\section{Preliminary}

\subsection{Diffusion Models and Latent Diffusion Models}
The core of DM is the diffusion process. Specifically, given a sample $x_{0} = x$, the diffusion process adds noise progressively to construct noisy samples $x_{1}, x_{2}, \dots, x_{T}$, where $x_{t} = \sqrt{\alpha_{t}} x_{t-1} + \sqrt{1 - \alpha_{t}} \epsilon$, and $\epsilon \sim \mathcal{N}(0, 1)$. This noise-adding process is commonly called the forward process. Such a progressive process can be transferred into a one-step process through reparameterization, where $x_{t} = \sqrt{\bar{\alpha}_{t}} x_{0} + \sqrt{1 - \bar{\alpha}_{t}} \epsilon$, and $\bar{\alpha}_{t} = \prod_{i=1}^{t} \alpha_{t}$. To generate a new sample, the DM reverses the forward process and denoises $x_{t}$ with the noise predicted by a denoising module $\epsilon_{\theta}(\cdot, t)$ at each step. The denoising module (usually an UNet) within the DM is trained by optimizing
\begin{equation}
	\min_{\theta} \mathbb{E}_{x_{0}, t\sim u(t), \epsilon\sim \mathcal{N}(0, 1)} \left\| \epsilon - \epsilon_{\theta}(x_{t}, t) \right\|_{2}^{2}\,.
\end{equation}

\citet{LDM} proposed to conduct these two processes in relatively low-dimensional latent space and put forward the latent diffusion model (LDM) to reduce the overhead in training and sampling, where an encoder $\mathcal{E}(\cdot)$ is used to map the image to the latent space, and a decoder $\mathcal{D}(\cdot)$ maps the latent to the pixel space. We denote $x_{t}$ as $\mathcal{A}(x, t) = \sqrt{\bar{\alpha}_{t}} \mathcal{E}(x) + \sqrt{1 - \bar{\alpha}_{t}} \epsilon$ to simplify the notation.

\subsection{Inference Task and Fine-tuning Task} \label{sec:modelTasks}
In this paper, we include two types of tasks, which are categorized into \emph{inference} task and \emph{fine-tuning} task.

The inference tasks we discuss include image variation and image inpainting~\cite{LDM}. Formally, these two tasks can be modeled as $f_{\text{infer}}(x, c') = \mathcal{D} \circ \epsilon_{t_{1}} \circ \dots \circ \epsilon_{t_{n}} \circ \mathcal{C}_{\text{infer}}(x, c')$, where $\mathcal{C}_{\text{infer}}$ extracts conditions from $x$ without optimizing, and $\epsilon_{t}$ is the algorithm conducting denoising at step $t$. $f_{\text{infer}}(x, c')$ can be seen as a end-to-end inference model because no optimizing process is included and because $f_{\text{infer}}(x, c')$ is actually used as an end-to-end inference model by some adversarial attacks~\cite{photoGuard, benchmark}.

The fine-tuning task we discuss includes textual inversion~\cite{textualInversion}. Formally, we model this task as $f_{\text{ft}}(x) = \mathcal{D} \circ \epsilon_{t_{1}} \circ \dots \circ \epsilon_{t_{n}} \circ \mathcal{C}_{\text{ft}}(x)$. Unlike the inference task, $f_{\text{ft}}(x)$ is not an end-to-end inference model because $\mathcal{C}_{\text{ft}}$ is an algorithm extracting conditions from the image $x$ by fine-tuning a pseudo word $S^{*}$.

\subsection{Adversarial Examples for LDMs} \label{sec:introAE}

AEs have been studied and utilized by many previous works~\cite{AdvDM, photoGuard, AntiDB} to prohibit the malicious use of LDMs. They, \emph{protectors}, optimize an adversarial perturbation $\delta$ within a space bounded by a budget $\varepsilon$ and add the adversarial perturbation to an image $x$ such that the \emph{adversary} can not use this image as a condition to conduct generative tasks with LDMs. Formally, the goal of the protectors can be interpreted as 
\begin{equation}
	\max_{\delta} \mathcal{L}(f(x + \delta), y) \quad \text{s.t.} \left\| \delta \right\|_{p} < \varepsilon\,,
\end{equation}
where $y$ is a supervised signal, $f(\cdot)$ is the LDM specialized for an arbitrary task, and $\mathcal{L}(\cdot, \cdot)$ is a loss function. Intuitively, the objective of AEs for LDMs is to make the image generated by LDMs with the protected image being a condition unreal, and such unreality is measured by a loss function $\mathcal{L}(\cdot, \cdot)$.

In practice, an adversarial attack is designed to optimize a surrogate loss function with a surrogate model to raise the loss $\mathcal{L}(\cdot, \cdot)$. For better illustration, we categorize existing adversarial attacks on LDMs into three types: Encoder-based, chain-based, and Monte-Carlo-based (MC-based).

The encoder-based adversarial attack, first proposed by \citet{photoGuard}, uses the encoder $\mathcal{E}(\cdot)$ within the LDM as its surrogate model. Formally, it crafts AEs by optimizing
\begin{equation} \label{eq:encoder}
	\left\| \mathcal{E}(x + \delta) - \mathcal{E}(x_{targ})\right\|_{2}^{2} \quad \text{s.t.} \quad \left\| \delta \right\|_{p} < \varepsilon\,,
\end{equation}
where $x_{targ}$ is a target image chosen by the protector. 
The key idea behind the encoder-based attack is to force the encoder to map the input image to some bad representation~\cite{photoGuard} and thus disrupt the image generation.

The chain-based adversarial attack (a.k.a diffusion attack~\cite{photoGuard}) includes the backward process of LDMs to simulate the inference of LDMs. Specifically, generating the chain-based targeted adversarial example~\cite{photoGuard} can be formulated as optimizing 
\begin{align}
	\left\| f_{\text{infer}}(x + \delta, c) - x_{targ} \right\|_{2}^{2}  \quad \text{s.t.} \quad \left\| \delta \right\|_{p} < \varepsilon\,,
\end{align}
where $c$ is the prompt and mask used during inference. In practice, the chain-based adversarial attack contains a long computation graph and thus induces a high computational cost.

The MC-based adversarial attack (a.k.a AdvDM)~\cite{AdvDM} constructs AEs on diffusion model by maximizing
\begin{align}\label{eq:advdm}
	\mathbb{E}_{t, \epsilon\sim\mathcal{N}(0,1)}\mathbb{E}_{x'_{t} \sim u(x'_{t})}\left[\left\|\epsilon - \epsilon_{\theta}(x'_{t}, t)\right\|_{2}^{2}\right]\,,
\end{align}
where $x' = x + \delta$, $\left\| \delta \right\|_{p} < \varepsilon$, and the expectation is estimated by Monte Carlo. The MC-based adversarial attack aims to minimize the likelihood $p_{\theta}(x')$, which is opposite to the objective of the training process of LDMs. In practice, this algorithm can be seen as the training algorithm of LDMs with reversed loss function, where $t$ is sampled from a uniform distribution $u(1, T)$, and $\delta$ is optimized by PGD~\cite{PGD} with gradient $\nabla_{x'}\left\|\epsilon - \epsilon_{\theta}(x'_{t}, t)\right\|_{2}^{2}$ at each step.

\section{The Perspective of Transferability} \label{sec:pers}

In this section, we reformulate the MC-based adversarial attack as a transfer-based attack possessing a set of surrogate models. We conduct a theoretical study and claim that the MC-based adversarial example generated on smooth surrogate models is more potent than treating surrogate models equally because smoother surrogate models can improve adversarial transferability. We also conduct empirical experiments. We find that the smoothness of $\epsilon_{\theta \cup t}(\cdot)$ within LDMs differs. Thus, we can improve AEs' performance by picking the subset $\mathcal{SM}'$ containing smoother $\epsilon_{\theta \cup t}(\cdot)$, which can support our findings and claims in the theoretical study.

\subsection{Reformulating the MC-based adversarial attack}

From \cref{sec:introAE}, we can infer that the MC-based adversarial attack utilizes the denoising models $\epsilon_{\theta}(\cdot, t)$ as surrogate models and manages to crack $f_{\text{infer}}(\cdot)$. To some extent, the MC-based adversarial attack can be seen as a transfer-based adversarial attack because $f_{\text{infer}}(\cdot)$, the target model, is not included when manufacturing AEs.

Moreover, we can unfold $\epsilon_{\theta}(x_{t}, t)$ to $\epsilon_{\theta}(\sqrt{\bar{\alpha}_{t}} \mathcal{E}(x) + \sqrt{1 - \bar{\alpha}_{t}} \epsilon, t)$ ($\epsilon_{\theta \cup t}(x)$ for short) and treat $t$ as a part of the denoising model's weight. In this sense, the protector owns a surrogate-model set $\mathcal{SM} = \left\{ \epsilon_{\theta \cup t}(\cdot) \mid 1 \le t \le T \right\}$ rather than one surrogate model $\epsilon_{\theta}(\cdot, t)$ for constructing AEs. Consequently, if the property, e.g., smoothness, of the surrogate models in $\mathcal{SM}$ varies, the protector can find a subset $\mathcal{SM}' \subseteq \mathcal{SM}$ such that one can generate better AEs with $\mathcal{SM}'$ and attack LDMs more effectively.

\subsection{Theoretical Study} \label{sec:theory}
The transferability of AEs has been studied theoretically in the image classification task~\cite{transferTheoryNIPS, transferTheorySP}. With modifications to definitions, we transfer a useful theorem on adversarial transferability in the image classification task to the image generation task. We only consider the untargeted attack.

\begin{definition}[Model-dependent loss function]\label{def:loss}
	For a specific model $\mathcal{F}$, the loss function is defined as $\mathcal{L}_{\mathcal{F}}(x, y) : \mathcal{X} \times \mathcal{Y} \mapsto \mathbb{R}$, where $x$ is the sample, and $y$ is the supervised signal.
\end{definition}

\begin{definition}[$\beta$-Smooth]\label{def:smooth}
	Given two arbitrary samples $x_{1}, x_{2} \in \mathcal{X}$ and an arbitrary ground truth $y \in \mathcal{Y}$, a model $\mathcal{F}$ is $\beta$-smooth if $\sup_{x_{1}, x_{2}, y} \frac{\left\| \nabla_{x_{1}}\mathcal{L}_{\mathcal{F}}(x_{1}, y) - \nabla_{x_{2}}\mathcal{L}_{\mathcal{F}}(x_{2}, y) \right\|_{2}}{\left\| x_{1} - x_{2} \right\|} \le \beta$.
\end{definition}

\begin{definition}[The risk and the empirical risk]\label{def:risk}
	Given a model $\mathcal{F}$ and a loss value $L \in \mathbb{R}$, the model's \emph{risk} is defined as $\gamma_{\mathcal{F}} = \text{Pr}(\mathcal{L}_{\mathcal{F}}(x, y)  > L)$. The model's \emph{empirical risk} is defined as $\mathbb{E}_{x}\left[\mathcal{L}_{\mathcal{F}}(x, y)  \right]$.
\end{definition}
\cref{def:risk} is one of the modifications to definitions compared to that of image classification task~\cite{transferTheoryNIPS, transferTheorySP}. Here, the risk is defined as the probability of the loss $\mathcal{L}_{\mathcal{F}}(x, y)$ being larger than a given value $L \in \mathbb{R}$.

\begin{definition}[($\alpha$, $\mathcal{F}$)-Effective adversarial example]\label{def:alpha}
	Given a model $\mathcal{F}$, an adversarial example $x'$ is ($\alpha$, $\mathcal{F}$)-effective if $\text{Pr}(\mathcal{L}_{\mathcal{F}}(x', y)  > L) \ge 1 - \alpha$.
\end{definition}
\begin{definition}[Loss gradient similarity]\label{def:gradSim}
	For two models $\mathcal{F}$ and $\mathcal{G}$ with loss functions $\mathcal{L}_{\mathcal{F}}$ and $\mathcal{L}_{\mathcal{G}}$, the loss gradient similarity is defined as $\mathcal{S}(\mathcal{L}_{\mathcal{F}}, \mathcal{L}_{\mathcal{G}}, x, y_{1}, y_{2}) = \frac{\nabla_{x}\mathcal{L}_{\mathcal{F}}(x, y_{1}) \cdot \nabla_{x}\mathcal{L}_{\mathcal{G}}(x, y_{2})}{\left\| \nabla_{x}\mathcal{L}_{\mathcal{F}}(x, y_{1}) \right\|_{2} \cdot \left\| \nabla_{x}\mathcal{L}_{\mathcal{G}}(x, y_{2}) \right\|_{2}}$.
\end{definition}

The loss gradient similarity utilizes Cosine Similarity to measure the difference in loss gradients between two models. Following \citet{transferTheorySP}, we further define the infimum of the loss gradient similarity as $\underline{\mathcal{S}}(\mathcal{L}_{\mathcal{F}}, \mathcal{L}_{\mathcal{G}}, x, y_{1}, y_{2}) = \inf_{x \in \mathcal{X}, y1, y2 \in \mathcal{Y}} \mathcal{S}(\mathcal{L}_{\mathcal{F}}, \mathcal{L}_{\mathcal{G}}, x, y_{1}, y_{2})$.

\begin{definition}[Adversarial transferability]\label{def:transfer}
	For two models $\mathcal{F}$ and $\mathcal{G}$ with loss values $L_{1}, L_{2} \in \mathbb{R}$, given a sample $x$, the transferability of the adversarial example $x' = x + \delta$ is defined as $T(\mathcal{F}, \mathcal{G}, x, y_{1}, y_{2}) = \mathbb{I}[\mathcal{L}_{\mathcal{F}}(x, y_{1}) \le L_{1} \wedge \mathcal{L}_{\mathcal{G}}(x, y_{2}) \le L_{2} \wedge $ $ \mathcal{L}_{\mathcal{F}}(x', y_{1}) > L_{1} \wedge \mathcal{L}_{\mathcal{G}}(x', y_{2}) > L_{2} ]$. $\mathbb{I}$ is the indicator function.
\end{definition}
\cref{def:transfer} is another key modification to definitions of previous works~\cite{transferTheoryNIPS, transferTheorySP}. Intuitively, high transferability means the adversarial example $x'$ can effectively degrade the performance of both $\mathcal{F}$ and $\mathcal{G}$.

Now, with modified definitions, we provide the lower bound of adversarial transferability.

\begin{theorem}[The lower bound of adversarial transferability]\label{the:lowBound}
	Assume the surrogate model $\mathcal{F}$ and the target model $\mathcal{G}$ are $\beta$-smooth. Given adversarial example $x' = x + \delta$ that is $(\alpha, \mathcal{F})$-Effective with $\left\| \delta \right\|_{2} < \varepsilon \in \mathbb{R}$, the transferability can be lower bounded by
	\begin{align}
		\nonumber &\text{Pr}(T(\mathcal{F}, \mathcal{G}, x, y_{1}, y_{2}) = 1) \ge (1 - \alpha) - \\ \nonumber &(\gamma_{\mathcal{F}} + \gamma_{\mathcal{G}}) - \frac{\varepsilon(1 + \alpha) - c_{\mathcal{F}}(1 - \alpha)}{\varepsilon - c_{\mathcal{G}}} \\ &- \frac{\varepsilon(1 - \alpha)}{\varepsilon - c_{\mathcal{G}}}\sqrt{2 - 2\underline{\mathcal{S}}(\mathcal{L}_{\mathcal{F}}, \mathcal{L}_{\mathcal{G}}, x, y_{1}, y_{2})}\,, \label{eq:keyTheorem}
	\end{align}
	where
	\begin{align}
		&c_{\mathcal{F}} = \min_{x \in D} \frac{L_{1} - \mathcal{L}_{\mathcal{F}}(x, y_{1}) - \beta\varepsilon^{2}/2}{\left\| \nabla_{x}\mathcal{L}_{\mathcal{F}}(x, y_{1}) \right\|_{2}}\,, \label{eq:tool1}\\
		&c_{\mathcal{G}} = \max_{x \in \mathcal{X}} \frac{L_{2} - \mathcal{L}_{\mathcal{G}}(x, y_{2}) + \beta\varepsilon^{2}/2}{\left\| \nabla_{x}\mathcal{L}_{\mathcal{G}}(x, y_{2}) \right\|_{2}}\,, \label{eq:tool2}\\
		&\gamma_{\mathcal{F}} = \text{Pr}(\mathcal{L}_{\mathcal{F}}(x, y_{1})  > L_{1})\,,\\
		&\gamma_{\mathcal{G}} = \text{Pr}(\mathcal{L}_{\mathcal{G}}(x, y_{2})  > L_{2})\,,
	\end{align}
	and $L_{1}, L_{2} \in \mathbb{R}$ and can be seen as constants.
\end{theorem}
The proof of \cref{the:lowBound} is provided in \cref{sec:proof}. Note that, though \cref{the:lowBound} is based on $l_{2}$-norm perturbation, one can trivially transfer \cref{the:lowBound} to $l_{\infty}$-norm perturbation-based because $\left\| \delta \right\|_{\infty} < \varepsilon \Rightarrow \left\| \delta \right\|_{2} < \sqrt{N} * \varepsilon$, where $N$ is the number of components in $\delta$.

\begin{remark}[on \cref{the:lowBound}]\label{remark:theroem}
	\cref{the:lowBound} can guide us to better manufacture the adversarial example from the perspective of the surrogate model's smoothness. Specifically, regarding generating MC-based AEs for image inpainting and variation, we have
	\begin{align}
		\mathcal{L}_{\mathcal{F}}(x, y) = \mathcal{L}_{\epsilon_{\theta \cup t}}(x, \epsilon) = \left\|\epsilon - \epsilon_{\theta \cup t}(x)\right\|_{2}\,,
	\end{align}
	and $\mathcal{L}_{\mathcal{G}}(x, y)$ can be
	\begin{align}
		\mathcal{L}_{f_{\text{infer}}}(x, x_{ori}) = \text{D}(f_{\text{infer}}(x), x_{ori})\,,
	\end{align}
	where $x_{ori}$ is the image to protect, and $\text{D}$ is an arbitrary differentiable metric to measure the quality of the generated image $f(x)$. We can infer from \cref{eq:keyTheorem} and \cref{eq:tool1} that we can raise the lower bound of the adversarial transferability by decreasing $\beta$ of the surrogate model, i.e., choosing a smooth surrogate model rather than a sharp one.
\end{remark}

\subsection{Empirical Study} \label{sec:empiricalStudy}

\cref{remark:theroem} demonstrates that we can improve the transferability by using smoother surrogate models. To confirm that this property can be applied to generating AEs for LDMs, we first conduct empirical experiments to measure the smoothness across different $\epsilon_{\theta \cup t}(\cdot)$ to see whether the smoothness differs according to $t$. Then, we partition $\mathcal{SM}$ into several $\mathcal{SM}'$ according to the smoothness of $\epsilon_{\theta \cup t}(\cdot)$ and construct AEs on each $\mathcal{SM}'$ to examine if AEs generated on the $\mathcal{SM}'$ containing smoother $\epsilon_{\theta \cup t}(\cdot)$ tend to have better performance. The setup of our empirical study is presented in \cref{sec:empiricalSetup}.

\subsubsection{Measuring Smoothness}\label{sec:smooth}

To measure the smoothness of the $\epsilon_{\theta \cup t}(\cdot)$, we choose $l_{2}$ magnitude of gradients\footnote{We acknowledge that there are other metrics to measure the smoothness like the dominant eigenvalue of the Hessian matrix, used by \citet{transferTheorySP}. However, the computation overhead of calculating the Hessian matrix of LDMs, where the input size is larger than that used by \citet{transferTheorySP}, is overly large. We can not afford it even on A100.} (i.e., $\left\|\nabla_{x}\mathcal{L}_{\epsilon_{\theta \cup t}}(x, \epsilon)\right\|_{2}$) because the smoothness defined by \cref{def:smooth} is bounded by the $l_{2}$ magnitude of gradients~\cite{transferTheoryNIPS}.

\begin{figure}[!t]
	\centering
	\includegraphics[width=\linewidth]{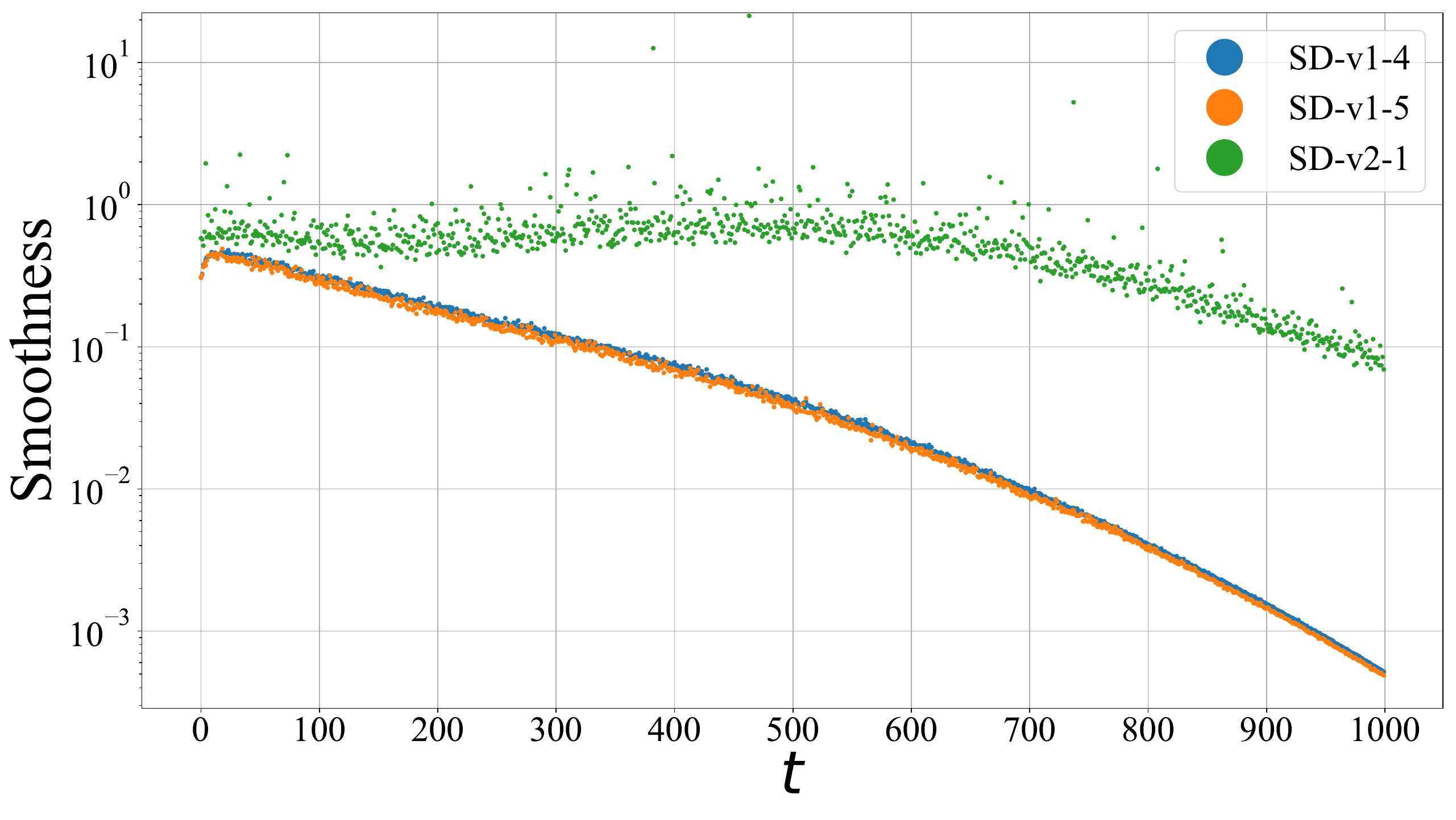}
	\caption{The smoothness of $\epsilon_{\theta \cup t}$ in different LDMs. Here, smoothness is measured by $\mathbb{E}_{x, \epsilon}[\left\|\nabla_{x}\mathcal{L}_{\epsilon_{\theta \cup t}}(x, \epsilon)\right\|_{2}]$. Lower the value of the y-axis, smoother the $\epsilon_{\theta \cup t}(\cdot)$ is.}\label{fig:smoothness}
\end{figure}

In \cref{fig:smoothness}, we can find that, on SD-v1-4 and SD-v1-5, the smoothness of $\epsilon_{\theta \cup t}(\cdot)$ decreases as $t$ increases. In this case, choosing the $\epsilon_{\theta \cup t}(\cdot)$ with larger $t$ should create AEs with better transferability and thus induce better protection to the image. On SD-v2-1, although the smoothness of $\epsilon_{\theta \cup t}(\cdot)$ also decreases and achieves its minimum when $t$ is large, it is still much larger than that of the SD-v1-4 and the SD-v1-5.

Overall, \cref{fig:smoothness} supports our argument that $t$ should be seen as a part of $\epsilon_{\theta}(\cdot, t)$'s weight as the property, e.g., smoothness, of $\epsilon_{\theta \cup t}(\cdot)$ within LDMs differs according to $t$. This finding motivates us to partition $\mathcal{SM}$ by $t$ and study the performance of AEs generated on each partition.

\begin{figure}[!t]
	\centering
	\includegraphics[width=\linewidth]{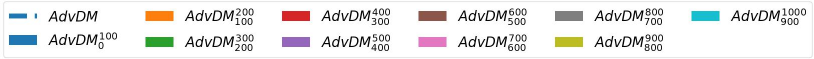}\\
	\begin{minipage}{\linewidth}
		\subfloat[SD-v1-4 to SD-v1-4]{\label{fig:difdIV4to4}
			\includegraphics[width=0.48\linewidth]{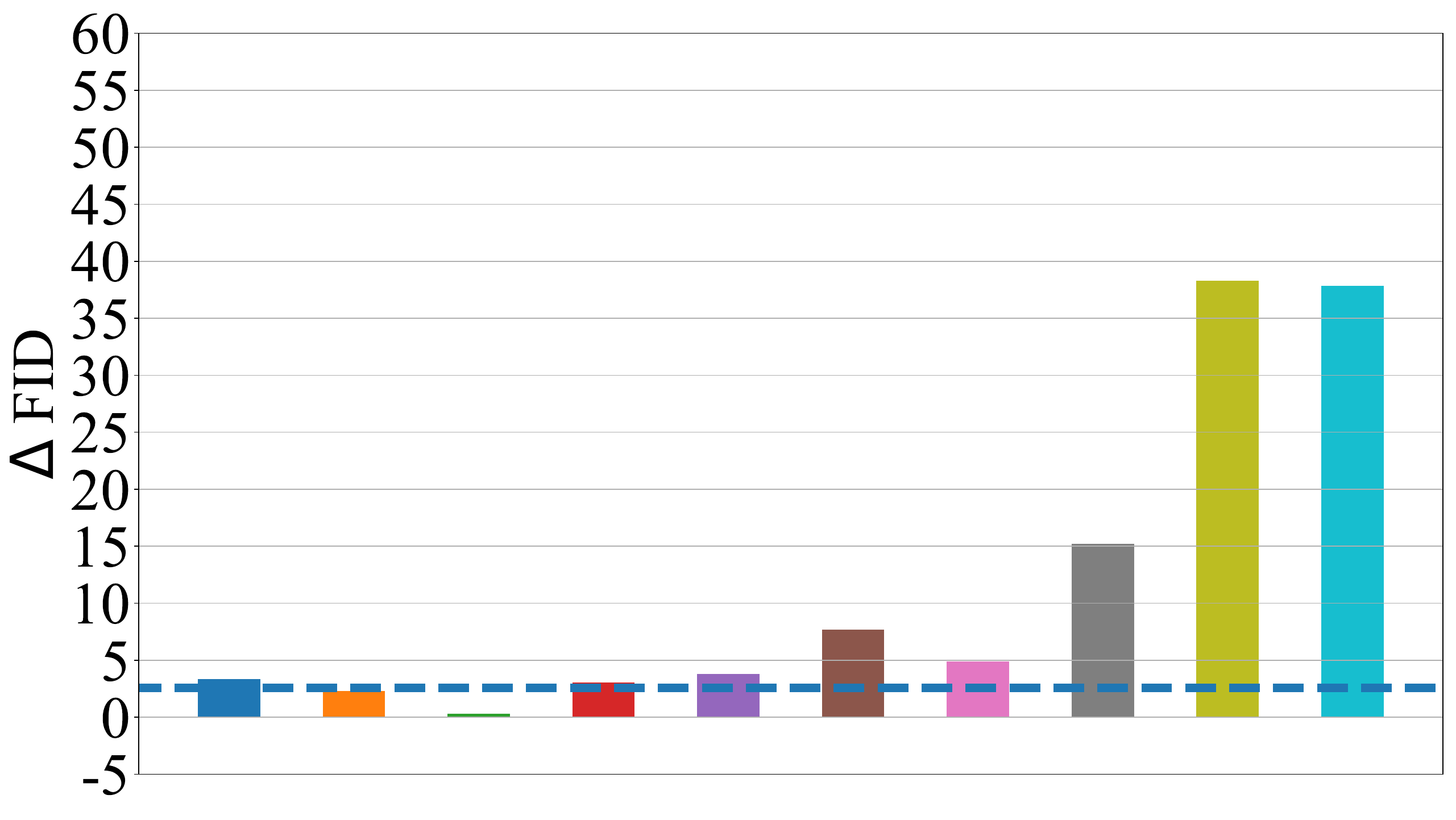}
		}
			%
		\subfloat[SD-v1-4 to SD-v2-1]{\label{fig:difdIV4to2}
			\includegraphics[width=0.48\linewidth]{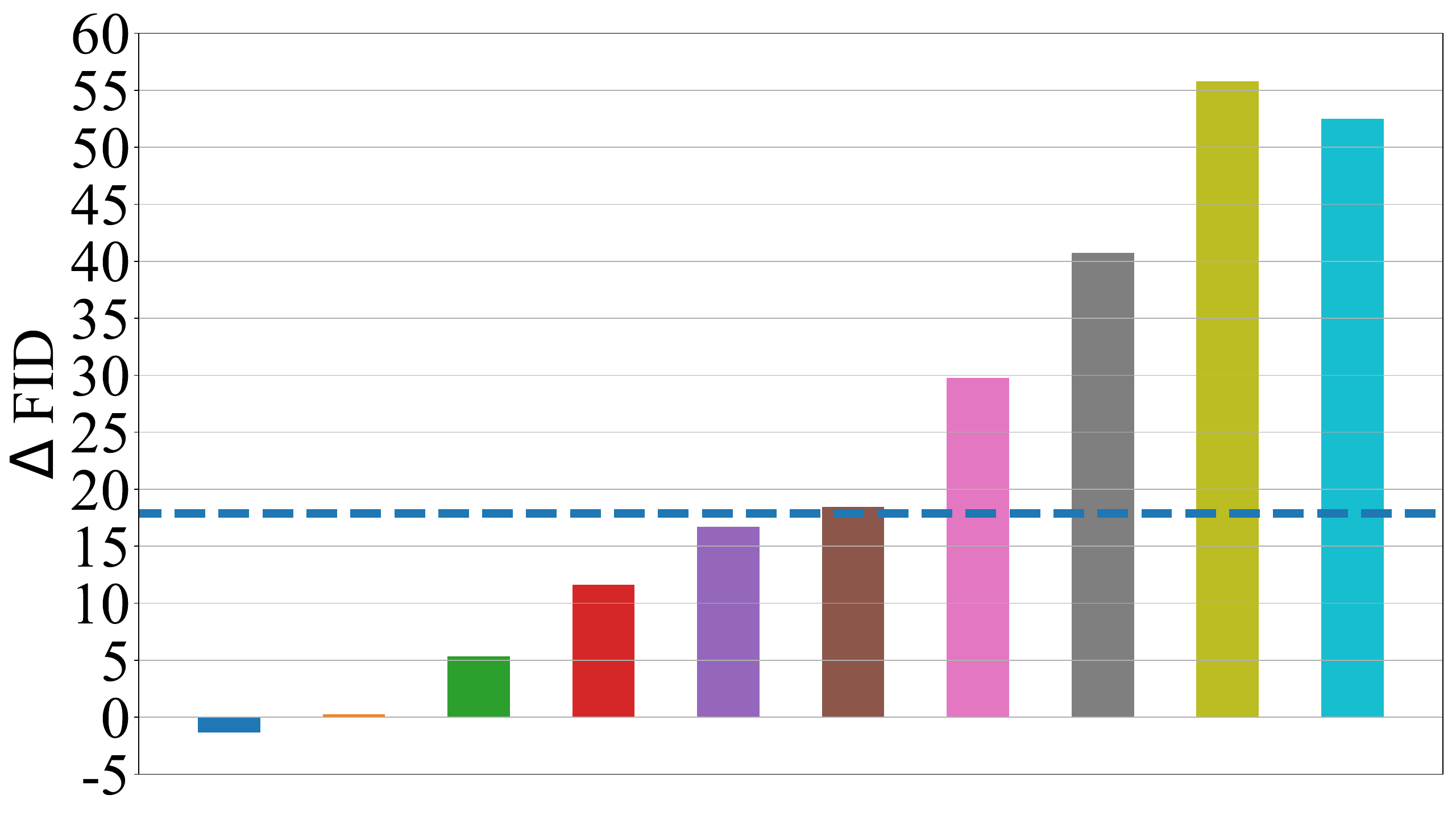}
		}
	\end{minipage}\\
	\begin{minipage}{\linewidth}
		\centering
		\subfloat[SD-v2-1 to SD-v1-4]{\label{fig:difdIV2to4}
			\includegraphics[width=0.48\linewidth]{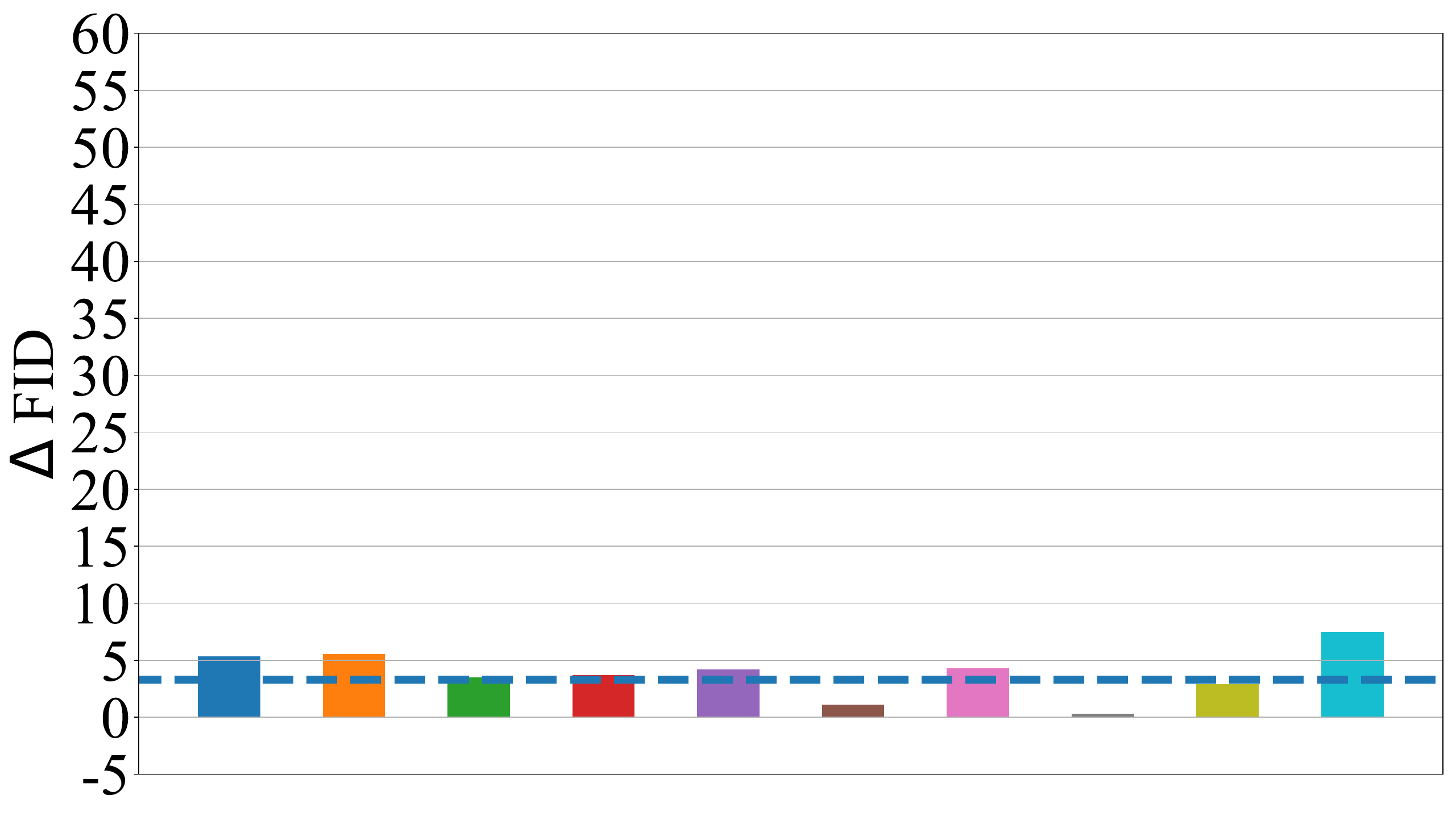}
		}
		\subfloat[SD-v2-1 to SD-v2-1]{\label{fig:difdIV2to2}
			\includegraphics[width=0.48\linewidth]{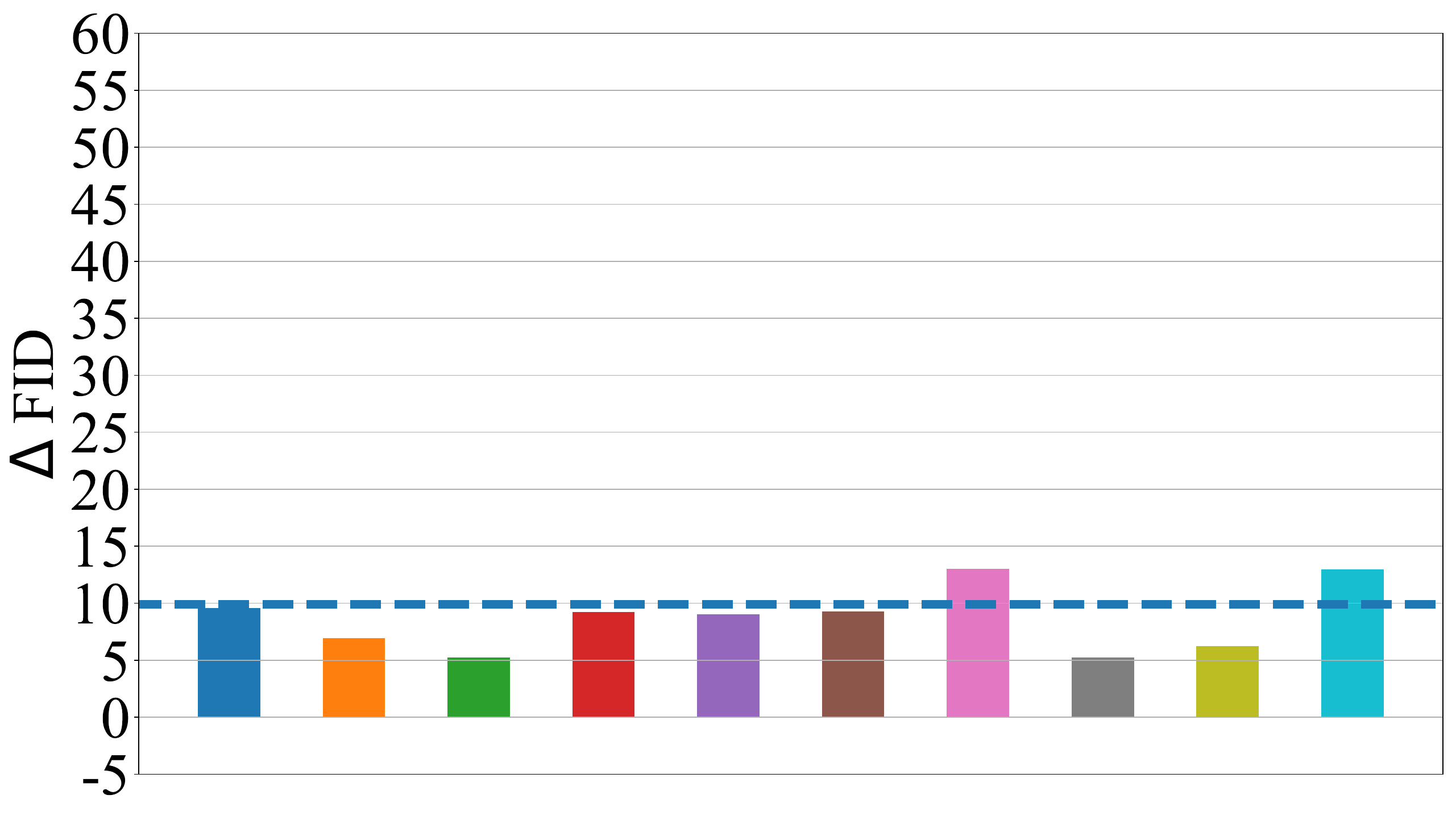}
		}
	\end{minipage}
	\caption{The increments $\uparrow$ of FID relative to the benign's. A larger increment means worse generated images and better AEs for breaking the image variation task. ``A to B'' means that AEs are generated by A and attack B.}\label{fig:dfidIV}
\end{figure}
\begin{figure*}[!t]
\centering
\subfloat[Original]{
	\begin{minipage}{0.115\linewidth}
		\includegraphics[width=\linewidth]{./ivVisual/ori/000000138115.pdf}\\
		\includegraphics[width=\linewidth]{./ivVisual/ori/000000211825.pdf}\\
		\includegraphics[width=\linewidth]{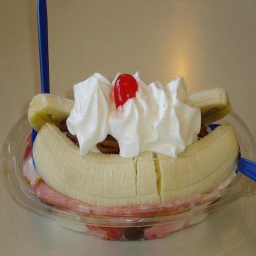}
	\end{minipage}
}
\subfloat[Benign]{
	\begin{minipage}{0.115\linewidth}
		\includegraphics[width=\linewidth]{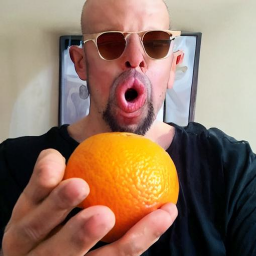}\\
		\includegraphics[width=\linewidth]{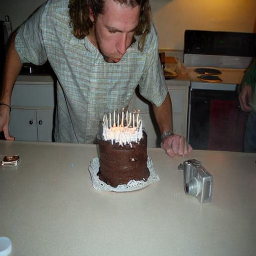}\\
		\includegraphics[width=\linewidth]{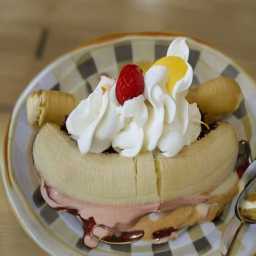}
	\end{minipage}
}
\subfloat[Chain]{
	\begin{minipage}{0.115\linewidth}
		\includegraphics[width=\linewidth]{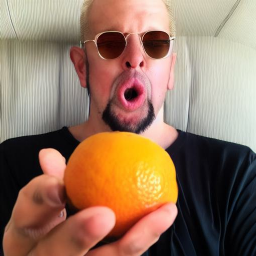}\\
		\includegraphics[width=\linewidth]{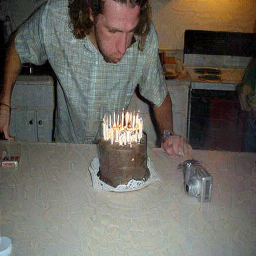}\\
		\includegraphics[width=\linewidth]{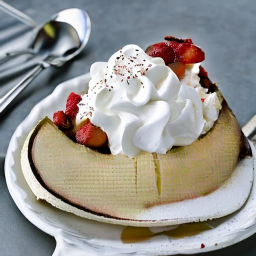}
	\end{minipage}
}
\subfloat[SDS]{
	\begin{minipage}{0.115\linewidth}
		\includegraphics[width=\linewidth]{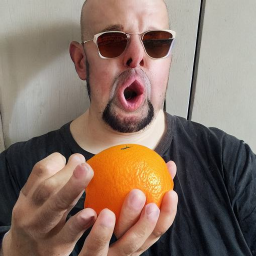}\\
		\includegraphics[width=\linewidth]{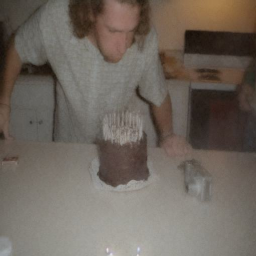}\\
		\includegraphics[width=\linewidth]{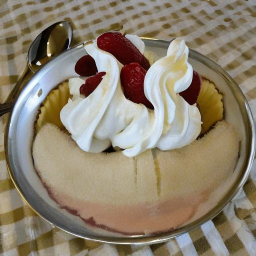}
	\end{minipage}
}
\subfloat[AdvDM]{
	\begin{minipage}{0.115\linewidth}
		\includegraphics[width=\linewidth]{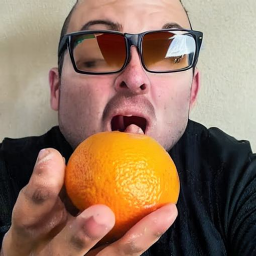}\\
		\includegraphics[width=\linewidth]{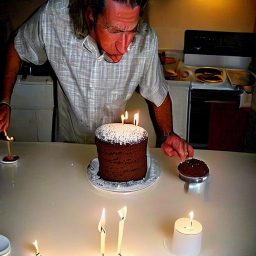}\\
		\includegraphics[width=\linewidth]{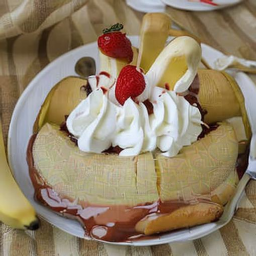}
	\end{minipage}
}
\subfloat[Mist (0)]{
	\begin{minipage}{0.115\linewidth}
		\includegraphics[width=\linewidth]{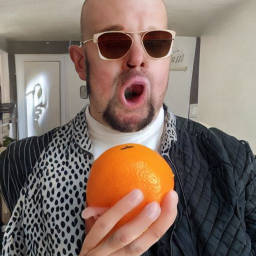}\\
		\includegraphics[width=\linewidth]{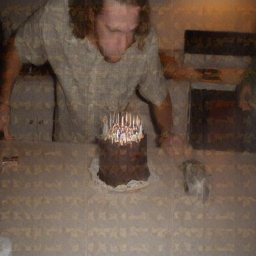}\\
		\includegraphics[width=\linewidth]{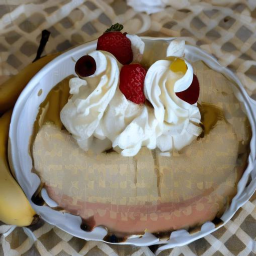}
	\end{minipage}
}
\subfloat[Mist (1)]{
	\begin{minipage}{0.115\linewidth}
		\includegraphics[width=\linewidth]{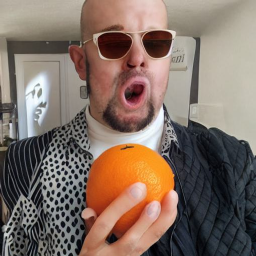}\\
		\includegraphics[width=\linewidth]{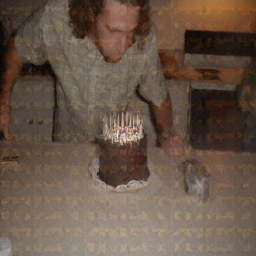}\\
		\includegraphics[width=\linewidth]{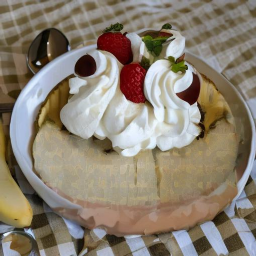}
	\end{minipage}
}
\subfloat[AdvDM$_{800}^{900}$]{
	\begin{minipage}{0.115\linewidth}
		\includegraphics[width=\linewidth]{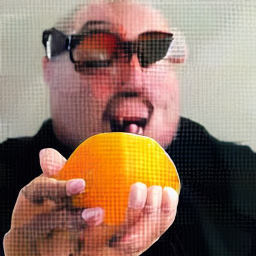}\\
		\includegraphics[width=\linewidth]{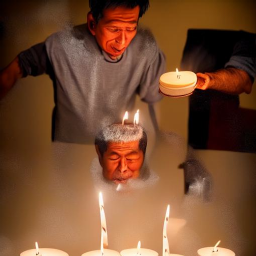}\\
		\includegraphics[width=\linewidth]{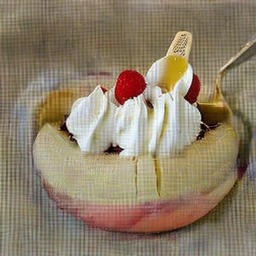}
	\end{minipage}
}
\caption{From top to bottom: Visualizations of image variation by SD-v1-4, image variation by InstructPix2Pix, and image inpainting by SD-2-Inpainting. Prompts from top to bottom: ``A man holds a large orange and puckers up for a silly selfie.'', ``On top of a brown cake, a man blows out white candles.'', and ``A depiction of spoons and whipped cream on a classic banana split served in a traditional way.'' View this figure in color. Use zoom-in to check the texture.} \label{fig:visual}
\end{figure*}
\subsubsection{Smoothness can Boost AEs for LDMs} \label{sec:smoothBoost}
The discussion and analysis in \cref{sec:smooth} show that the $\epsilon_{\theta \cup t}(\cdot)$ in SD-v1-4 and SD-v1-5 become smoother as $t$ grows and that $\epsilon_{\theta \cup t}(\cdot)$ in SD-v2-1 are all relatively sharp. Based on our analysis in \cref{sec:theory}, we conjecture that AEs generated on $\epsilon_{\theta \cup t}(\cdot)$ with larger $t$ in SD-v1-4 and SD-v1-5 are more effective in attacking LDMs. Also, due to the poor smoothness of $\epsilon_{\theta \cup t}(\cdot)$ in SD-v2-1, AEs generated on SD-v2-1 should be relatively ineffective.

To validate our conjectures above, we partition $\mathcal{SM}$ into 10 folds, where $\mathcal{SM}_{a}^{b} = \left\{ \epsilon_{\theta \cup t}(\cdot) \mid a < t \le b \right\}$, and test AEs generated on each fold by AdvDM~\cite{AdvDM} to examine whether the $\mathcal{SM}_{a}^{b}$ containing smoother $\epsilon_{\theta \cup t}(\cdot)$ generates better AEs. For conciseness, we denote AdvDM using $\mathcal{SM}_{a}^{b}$ as AdvDM$_{a}^{b}$ wherein AdvDM is AdvDM$_{0}^{1000}$. Following \citet{AdvDM} and \citet{benchmark}, we use FID~\cite{FID} to measure the effectiveness of AEs. We test AEs on image variation task and leave results on image inpainting in \cref{sec:moreSmoothness}. To save space, we only report the results of SD-v1-4 and SD-v2-1 and leave the results of other LDMs in \cref{sec:moreSmoothness}.

In \cref{fig:dfidIV}, we can find that AdvDM$_{800}^{900}$ and AdvDM$_{900}^{1000}$ can achieve larger increments of FID when AEs are generated on SD-v1-4, which means these AEs can effectively protect the images from being edited. Particularly in \cref{fig:dfidIV}\subref{fig:difdIV4to2}, the increment of FID grows monotonically as $a$ and $b$ of AdvDM$_{a}^{b}$ become larger. These results are consistent with our conjectures because $\epsilon_{\theta \cup t}(\cdot)$ in SD-v1-4 tends to be smooth as $t$ increases.

When the AE is generated on SD-v2-1, the differences in increments of FID between different methods are relatively minor, and AEs generated by $\epsilon_{\theta \cup t}(\cdot)$ with smaller $t$ even have better performance (see \cref{sec:moreSmoothness}), which suggests that \emph{the smoothness of $\epsilon_{\theta \cup t}(\cdot)$ may be overwhelmed by other factors when the discrepancy of the smoothness is insignificant}. This claim can also be supported by the results in \cref{fig:dfidIV}\subref{fig:difdIV4to4} and \cref{fig:smoothness}, where the increments of FID remain low when $t$ and the decline of $\mathbb{E}_{x, \epsilon}[\left\|\nabla_{x}\mathcal{L}_{\epsilon_{\theta \cup t}}(x, \epsilon)\right\|_{2}]$ is small.

One intriguing phenomenon in \cref{fig:dfidIV}\subref{fig:difdIV4to2} and \cref{fig:dfidIV}\subref{fig:difdIV2to2} is that, when $t$ is large enough, AEs generated on SD-v1-4 are even more effective in attacking SD-v2-1 than those generated on SD-v2-1. This phenomenon demonstrates that choosing a smoother surrogate model can be better than choosing a surrogate model with the same weights as the target model when we try to crack LDMs. This phenomenon also supports our motivation that we should re-examine the adversarial attack for LDMs from the perspective of transferability because choosing the surrogate model having the same weights as the target model's, which can be compared to white-box adversarial attack in image classification, does not ensure better performance.


\section{Experiments} \label{sec:experiment}

To demonstrate the effectiveness of our method, we compare the performance of AEs generated by our method to other SOTA adversarial attacks for LDMs. Considering that there exist other factors (e.g., iteration steps~\cite{moreIter} and gradient variance~\cite{ensemble2, ensemble3}) affecting the performance of AEs, we conduct ablation studies in \cref{sec:otherFactors} to show that the performance gained by our method indeed comes from the smoothness of surrogate models. Lastly, we empirically show that a good AE for degrading image inpainting and variation is not necessarily a good AE for breaking textual inversion~\cite{textualInversion} and explain why such a phenomenon exists. We leave the setups of our experiments in \cref{sec:experSetup} and the results of AEs facing preprocessing adversarial defenses~\cite{jpgDefense, tvm} in \cref{sec:defense}. We also leave the results with SD-v1-5 being the surrogate model in \cref{sec:v15MainReuslt} because its results are similar to SD-v1-4.

\begin{table*}[!t]
	\caption{The performance of AEs in corrupting image variation. The best result in each column is in \textbf{bold}.} \label{tab:ivMainResult}
	\Huge
	\resizebox{\linewidth}{!}{
		\begin{tabular}{@{}cccccccccccccccccccccc@{}}
			\toprule
			\multirow{2}{*}{Surrogate Model} & \multirow{2}{*}{Method}                    & \multicolumn{4}{c}{FID$\uparrow$}                                               & \multicolumn{4}{c}{IS$\downarrow$}                                           & \multicolumn{4}{c}{CLIP$\downarrow$}                                          & \multicolumn{4}{c}{CLIP-IQA (Natural)$\downarrow$}                        & \multicolumn{4}{c}{CLIP-IQA (Noisiness)$\downarrow$}                      \\ \cmidrule(l){3-22} 
			&                                            & SD-v1-4         & SD-v1-5         & SD-v2-1         & Instruct        & SD-v1-4        & SD-v1-5        & SD-v2-1       & Instruct       & SD-v1-4        & SD-v1-5        & SD-v2-1        & Instruct       & SD-v1-4       & SD-v1-5       & SD-v2-1       & Instruct      & SD-v1-4       & SD-v1-5       & SD-v2-1       & Instruct      \\ \midrule
			/                                & Benign                                     & 164.28          & 166.51          & 195.82          & 130.41          & 16.60          & 16.06          & 11.25         & 16.39          & 34.47          & 34.61          & 32.38          & 34.53          & 0.45          & 0.46          & 0.27          & 0.32          & 0.47          & 0.48          & 0.38          & 0.36          \\ \midrule
			\multirow{9}{*}{SD-v1-4}         & AdvDM                                      & 166.86          & 173.84          & 213.68          & 180.03          & 15.29          & 15.29          & 9.56          & 15.06          & 34.65          & 34.45          & 31.24          & 34.22          & 0.45          & 0.44          & 0.18          & 0.22          & 0.44          & 0.44          & 0.27          & 0.39          \\
			& SDS                                        & 175.53          & 178.20          & 209.84          & 158.95          & 16.53          & 16.66          & 11.31         & 15.52          & 34.48          & 34.36          & 31.90          & 34.08          & 0.42          & 0.43          & 0.22          & 0.22          & 0.46          & 0.46          & 0.34          & 0.31          \\
			& Chain                                      & 182.40          & 178.99          & 228.90          & 162.77          & 12.87          & 13.11          & 8.10          & 15.62          & 33.85          & 34.01          & 30.07          & 34.46          & 0.36          & 0.37          & 0.14          & 0.22          & 0.39          & 0.41          & 0.21          & 0.34          \\
			& Mist (0)                                   & 181.72          & 181.69          & 233.38          & 156.41          & 16.07          & 15.28          & 8.82          & 15.47          & 34.17          & 34.02          & 30.53          & 34.02          & 0.43          & 0.44          & 0.19          & \textbf{0.17} & 0.46          & 0.46          & 0.33          & \textbf{0.17} \\
			& Mist (1)                                   & 181.39          & 181.32          & 232.41          & 157.93          & 15.88          & 15.05          & 8.42          & \textbf{14.54} & 34.10          & 34.09          & 30.43          & \textbf{33.82} & 0.44          & 0.43          & 0.19          & \textbf{0.17} & 0.46          & 0.46          & 0.32          & \textbf{0.17} \\
			& Mist (5)                                   & 166.87          & 168.44          & 214.75          & 182.16          & 16.00          & 15.93          & 9.92          & 15.51          & 34.61          & 34.53          & 31.09          & 34.20          & 0.45          & 0.47          & 0.21          & 0.22          & 0.44          & 0.44          & 0.30          & 0.39          \\ \cmidrule(l){2-22} 
			& AdvDM$_{800}^{900}$    & 202.56          & \textbf{200.74}          & 251.58          & \textbf{188.60}          & \textbf{11.43} & \textbf{11.62} & 8.08          & 15.26          & \textbf{33.52} & \textbf{33.71} & \textbf{29.63}          & 33.93          & \textbf{0.30}          & \textbf{0.31} & \textbf{0.09} & 0.22          & \textbf{0.31} & \textbf{0.32} & \textbf{0.16} & 0.41          \\
			& Mist$_{800}^{900}$ (1) & 184.22          & 181.84          & 231.27          & 155.76          & 16.15          & 15.28          & 8.89          & 14.97          & 34.20          & 34.09          & 30.44          & 33.98          & 0.44          & 0.45          & 0.20          & 0.18          & 0.46          & 0.47          & 0.32          & \textbf{0.17} \\
			& Mist$_{800}^{900}$ (5) & \textbf{204.46} & 199.11          & \textbf{251.94} & 188.47          & 12.11          & 12.26          & \textbf{7.15} & 14.83          & 33.69          & 33.83          & 29.67          & 33.95          & \textbf{0.30}          & 0.32          & \textbf{0.09} & 0.21          & 0.32          & 0.34          & \textbf{0.16} & 0.41          \\ \midrule
			\multirow{9}{*}{SD-v2-1}         & AdvDM                                      & 167.58          & 167.57          & 205.71          & 158.43          & 16.49          & 16.26          & 10.81         & 16.66          & 34.45          & 34.50          & 31.63          & 34.46          & 0.44          & 0.43          & 0.24          & 0.22          & 0.46          & 0.46          & 0.33          & 0.31          \\
			& SDS                                        & 165.58          & 166.92          & 197.62          & 164.52          & 15.56          & 15.04          & 11.28         & 16.14          & 34.41          & 34.44          & 31.93          & 34.38          & 0.35          & 0.36          & 0.20          & 0.23          & 0.42          & 0.42          & 0.29          & 0.35          \\
			& Chain                                      & 169.05          & 169.11          & 209.25          & 142.08          & 16.41          & 15.98          & 10.19         & 16.33          & 34.66          & 34.39          & 30.84          & 34.61          & 0.43          & 0.48          & 0.23          & 0.24          & 0.43          & 0.45          & 0.25          & 0.28          \\
			& Mist (0)                                   & 182.05          & 181.88          & 231.28          & 158.82          & 15.83          & 15.12          & 8.75          & 15.02          & 34.20          & 34.06          & 30.54          & 33.90          & 0.43          & 0.44          & 0.20          & 0.18          & 0.46          & 0.45          & 0.33          & \textbf{0.17} \\
			& Mist (1)                                   & 179.14          & 179.71          & 227.45          & 139.39          & 16.07          & 15.26          & 8.94          & 15.81          & 34.24          & 34.31          & 30.85          & 34.18          & 0.44          & 0.45          & 0.22          & 0.21          & 0.45          & 0.46          & 0.33          & 0.20          \\
			& Mist (5)                                   & 165.70          & 167.17          & 205.23          & 159.58          & 16.95          & 16.84          & 10.89         & 16.52          & 34.35          & 34.47          & 31.75          & 34.53          & 0.45          & 0.45          & 0.23          & 0.22          & 0.45          & 0.46          & 0.32          & 0.33          \\ \cmidrule(l){2-22} 
			& AdvDM$_{800}^{900}$    & 167.19          & 169.71          & 202.05          & 160.66          & 15.13          & 16.12          & 11.77         & 16.18          & 34.33          & 34.34          & 31.99          & 34.38          & 0.39          & 0.39          & 0.24          & 0.22          & 0.44          & 0.46          & 0.30          & 0.35          \\
			& Mist$_{800}^{900}$ (1) & 179.03          & 181.39          & 236.64          & 148.00          & 15.81          & 15.13          & 8.58          & 15.21          & 34.05          & 34.05          & 30.13          & 34.17          & 0.44          & 0.44          & 0.19          & 0.19          & 0.47          & 0.48          & 0.29          & 0.19          \\
			& Mist$_{800}^{900}$ (5) & 168.08          & 167.66          & 201.62          & 161.90          & 15.73          & 15.82          & 11.89         & 15.87          & 34.19          & 34.32          & 31.91          & 34.33          & 0.39          & 0.39          & 0.23          & 0.23          & 0.45          & 0.46          & 0.31          & 0.34          \\ \bottomrule
	\end{tabular}}
\end{table*}

\subsection{Qualitative Results} \label{sec:quali}

To intuitively demonstrate how different adversarial attacks corrupt the image variation and inpainting, we visualize the generated images in \cref{fig:visual}. When attacking SD-v1-4 and SD-2-Inpainting, one significant pattern induced by AdvDM$_{800}^{900}$ is the latticed texture across the whole image. This pattern makes the generated images noisy, unreal, and lacking depth of field. Among the images generated by InstructPix2Pix, the image of AdvDM$_{800}^{900}$ fails to mimic the composition and the objects in the original image, where the cake becomes a head, and candles are burning above the head. Other methods, like Chain, SDS, and Mist, introduce noise, blurriness, and the pattern from the target image to the generated image, respectively. We present more visualizations in \cref{sec:moreVisual}.

\subsection{Quantitative Results} \label{sec:quanti}

\begin{table*}[!t]
	\begin{minipage}{\linewidth}
	\caption{The performance of AEs in corrupting image inpainting. The best result in each column is in \textbf{bold}.} \label{tab:iiMainResult}
	{\Huge
	\resizebox{\linewidth}{!}{
		\begin{tabular}{@{}cccccccccccc@{}}
			\toprule
			\multirow{2}{*}{Surrogate Model} & \multirow{2}{*}{Method}                    & \multicolumn{2}{c}{FID$\uparrow$}           & \multicolumn{2}{c}{IS$\downarrow$}           & \multicolumn{2}{c}{CLIP$\downarrow$}         & \multicolumn{2}{c}{CLIP-IQA (Natural)$\downarrow$} & \multicolumn{2}{c}{CLIP-IQA (Noisiness)$\downarrow$} \\ \cmidrule(l){3-12} 
			&                                            & SD-Inpainting   & SD-2-Inpainting & SD-Inpainting  & SD-2-Inpainting & SD-Inpainting  & SD-2-Inpainting & SD-Inpainting     & SD-2-Inpainting    & SD-Inpainting      & SD-2-Inpainting     \\ \midrule
			/                                & Benign                                     & 108.04          & 105.79          & 15.67          & 16.64           & 34.58          & 34.93           & 0.47              & 0.53               & 0.43               & 0.45                \\ \midrule
			\multirow{8}{*}{SD-v1-4}         & AdvDM                                      & 145.78          & 143.18          & 14.61          & 14.86           & 34.37          & 34.85           & 0.23              & 0.25               & 0.35               & 0.32                \\
			& SDS                                        & 142.93          & 147.80          & 15.19          & 16.09           & 34.56          & 35.00           & 0.35              & 0.34               & 0.36               & 0.27                \\
			& Mist (0)                                   & 156.25          & 163.67          & 13.90          & 14.75           & 34.02          & \textbf{34.17}           & 0.25              & 0.22               & 0.36               & 0.30                \\
			& Mist (1)                                   & 157.43          & 162.91          & 14.47          & 14.69           & 34.04          & 34.30           & 0.24              & \textbf{0.21}      & 0.37               & 0.29                \\
			& Mist (5)                                   & 145.68          & 142.33          & 14.18          & 15.13           & 34.44          & 35.04           & 0.25              & 0.27               & 0.35               & 0.32                \\ \cmidrule(l){2-12} 
			& AdvDM$_{800}^{900}$    & \textbf{161.83} & 168.86          & \textbf{13.18} & 12.79           & 34.18          & 34.66           & \textbf{0.21}     & \textbf{0.21}      & \textbf{0.26}      & \textbf{0.23}       \\
			& Mist$_{800}^{900}$ (1) & 158.27          & 163.56          & 14.13          & 14.89           & 33.97          & 34.23           & 0.24              & 0.22               & 0.38               & 0.30                \\
			& Mist$_{800}^{900}$ (5) & 161.11          & \textbf{169.32}          & 14.23          & \textbf{12.58}  & 34.21          & 34.54           & \textbf{0.21}     & 0.22               & 0.28               & \textbf{0.23}       \\ \midrule
			\multirow{8}{*}{SD-v2-1}         & AdvDM                                      & 126.19          & 128.06          & 15.32          & 17.29           & 34.68          & 34.85           & 0.35              & 0.36               & 0.39               & 0.38                \\
			& SDS                                        & 122.36          & 122.96          & 14.63          & 15.75           & 34.36          & 34.60           & 0.35              & 0.36               & 0.38               & 0.35                \\
			& Mist (0)                                   & 153.58          & 162.56          & 14.20          & 14.20           & 34.03          & 34.21           & 0.25              & \textbf{0.21}      & 0.37               & 0.28                \\
			& Mist (1)                                   & 146.19          & 149.69          & 15.61          & 16.38           & 34.48          & 34.54           & 0.28              & 0.26               & 0.38               & 0.32                \\
			& Mist (5)                                   & 124.32          & 130.38          & 15.66          & 16.48           & 34.47          & 34.77           & 0.35              & 0.38               & 0.39               & 0.38                \\ \cmidrule(l){2-12} 
			& AdvDM$_{800}^{900}$    & 123.76          & 123.33          & 16.51          & 15.62           & 34.42          & 34.76           & 0.36              & 0.39               & 0.38               & 0.36                \\
			& Mist$_{800}^{900}$ (1) & 150.37          & 152.98          & 14.84          & 14.86           & 34.09          & 34.33           & 0.26              & 0.25               & 0.39               & 0.32                \\
			& Mist$_{800}^{900}$ (5) & 120.26          & 120.34          & 16.78          & 15.57           & 34.33          & 34.66           & 0.37              & 0.38               & 0.39               & 0.36                \\ \midrule
			SD-Inpainting                    & Chain                                      & 151.68          & 147.03          & 14.83          & 14.49           & \textbf{33.82} & 34.30           & 0.25              & 0.26               & 0.33               & 0.33                \\ \midrule
			SD-2-Inpainting                  & Chain                                      & 154.01          & 149.47          & 14.94          & 14.55           & 34.10          & 34.49           & 0.23              & 0.26               & 0.34               & 0.37                \\ \bottomrule
			\end{tabular}}}
		\end{minipage}
\end{table*}

We quantitatively evaluate the performance of AEs from three perspectives, including preventing imitations, biasing the semantics provided by the prompt, and degrading the visual quality (see \cref{sec:experSetup}).

In \cref{tab:ivMainResult} and \cref{tab:iiMainResult}, we can find that the best results are achieved mainly by  AdvDM$_{800}^{900}$ or Mist$_{800}^{900} (5)$. Specifically, AEs generated by Mist$_{800}^{900}$ with SD-v1-4  being the surrogate model induce higher FID than other methods, suggesting that the generated images' distribution differs from the original images'. Thus, the adversary may fail to mimic the protected images. Regarding biasing the semantics provided by prompts, AdvDM$_{800}^{900}$ and Mist$_{800}^{900}$ tend to have low CLIP values, indicating that the generated images have different semantics from the prompts. In practice, this property can hinder the adversary from generating images fulfilling its purpose.

As for degrading the generated images' quality, AdvDM$_{800}^{900}$ and Mist$_{800}^{900} (5)$ also lead to low IS and CLIP-IQA values when attacking LDMs except for InstructPix2Pix. Significantly, the values of CLIP-IQA demonstrate that AdvDM$_{800}^{900}$ introduce more unnatural and noisy patterns into the generated images, which is consistent with our observation in \cref{sec:quali}. When attacking InstructPix2Pix, as we observe in \cref{sec:quali}, AdvDM$_{800}^{900}$ corrupts the image variation by biasing the semantics of the generated image rather than degrading the visual quality, which leads to relatively high FID and CLIP-IQA values.

Also, as we claimed in \cref{sec:smoothBoost}, SD-v2-1 is not a good surrogate model for manufacturing MC-based AEs. All MC-based AEs constructed on SD-v2-1 have poorer performance than those constructed on SD-v1-4. Only Mist $(0)$, an encoder-based adversarial attack, can achieve similar results across these two surrogate models.

\subsection{Ablation Study} \label{sec:otherFactors}

\subsubsection{Iteration Steps}
\begin{table}[!t]
	\caption{The performance of AEs generated by AdvDM~\cite{AdvDM} and AdvDM$_{800}^{900}$ with different iteration steps in cracking image variation. The surrogate model is SD-v1-4.} \label{tab:steps} {\Huge
		\resizebox{\linewidth}{!}{
			\begin{tabular}{@{}cccccccccc@{}}
				\toprule
				\multirow{2}{*}{Method}                 & \multirow{2}{*}{Iteration Steps} & \multicolumn{4}{c}{FID$\uparrow$}                & \multicolumn{4}{c}{IS$\downarrow$}                 \\ \cmidrule(l){3-10} 
				&                                  & SD-v1-4 & SD-v1-5 & SD-v2-1 & Instruct & SD-v1-4 & SD-v1-5 & SD-v2-1 & Instruct \\ \midrule
				\multirow{3}{*}{AdvDM}                  & 40                               & 166.86  & 173.84  & 213.68  & 180.03   & 15.29   & 15.29   & 9.56    & 15.06    \\
				& 500                              & 170.74  & 173.84  & 225.97  & 185.13   & 15.68   & 15.29   & 9.19    & 15.19    \\ 
				& 1,000                              & 173.77  & 171.11  & 230.73  & 184.93   & 15.30   & 15.17   & 9.22    & 15.29    \\\midrule
				\multirow{2}{*}{AdvDM$_{800}^{900}$} & 40 & 202.56          & 200.74          & 251.58          & 188.60          & 11.43 & 11.62 & 8.08          & 15.26     \\
				& 500                              & 243.93  & 242.59  & 279.78  & 192.93   & 7.08    & 7.71    & 6.17    & 14.02\\
				 \bottomrule
	\end{tabular}}}
\end{table}

\citet{moreIter} find that the adversarial attack with only a few iterations has limited the attack convergence to optimal targeted transferability. \citet{AdvDM} also improves AdvDM's performance by including more iteration steps. These findings inspire us to check whether the improvement gained by our method comes from a faster convergence rate rather than the smoothness. We set the iteration steps of AdvDM to at most 1,000 to ensure convergence. In \cref{tab:steps}, we can find that increasing iteration steps can boost the performance of AEs but still can not surpass AdvDM$_{800}^{900}$ with 40 iteration steps. This result demonstrates that the adversarial perturbation generated by AdvDM$_{800}^{900}$ converges to a more optimum point. Moreover, the performance of AdvDM$_{800}^{900}$ can also be improved by incorporating more iteration steps, indicating that the performance of AdvDM$_{800}^{900}$ is not gained by a faster convergence rate. We present the results of different adversarial attacks with 500 iteration steps and different inference tasks in \cref{sec:500Steps}.

\subsubsection{Gradient Variance}
\begin{figure}[!t]
	\centering
	\includegraphics[width=0.6\linewidth]{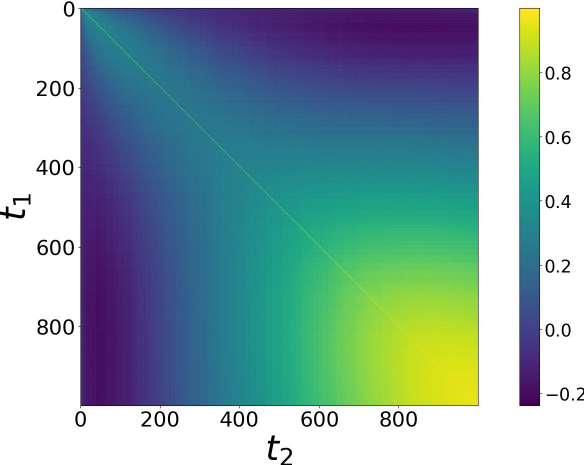}
	\caption{The loss gradient similarity matrix of SD-v1-4. The value of position $(t_1, t_2)$ is $\mathbb{E}_{x, \epsilon_{1}, \epsilon_{2}}\mathcal{S}(\mathcal{L}_{\epsilon_{\theta \cup t_{1}}}, \mathcal{L}_{\epsilon_{\theta \cup t_{2}}}, x, \epsilon_{1}, \epsilon_{2})$.} \label{fig:grad}
\end{figure}
\begin{table}[!t]
\caption{The performance of AEs generated by AdvDM$_{99}^{100}$, AdvDM$_{499}^{500}$, and AdvDM$_{999}^{1000}$ in cracking image variation. The surrogate model is SD-v1-4.} \label{tab:gv} {\huge
	\resizebox{\linewidth}{!}{
		\begin{tabular}{@{}ccccccccc@{}}
			\toprule
			\multirow{2}{*}{Method}                  & \multicolumn{4}{c}{FID$\uparrow$}                & \multicolumn{4}{c}{IS$\downarrow$}                 \\ \cmidrule(l){2-9} 
			& SD-v1-4 & SD-v1-5 & SD-v2-1 & Instruct & SD-v1-4 & SD-v1-5 & SD-v2-1 & Instruct \\ \midrule
			AdvDM$_{99}^{100}$   & 166.08  & 165.58  & 195.41  & 168.37   & 16.5    & 16.21   & 11.66   & 16.53    \\
			AdvDM$_{499}^{500}$  & 169.13  & 168.57  & 213.9   & 179.48   & 16.46   & 15.75   & 10.25   & 16.03    \\
			AdvDM$_{999}^{1000}$ & 195.62  & 193.11  & 243.57  & 188.71   & 12.64   & 12.18   & 7.66    & 15.34    \\ \bottomrule
\end{tabular}}}
\end{table}
We mentioned in \cref{sec:pers} that, to some extent, the MC-based adversarial attack owns a surrogate-model set rather than one surrogate model for constructing AEs, which makes the MC-based adversarial attack an ensemble-based attack~\cite{ensemble2, ensemble3}. \citet{ensemble3} claimed the low loss gradient similarity between surrogate models would degrade the performance of AEs. Thus, we need to confirm whether this factor, rather than smoothness, dominates the performance of AEs.

We measure the loss gradient similarity between $\epsilon_{\theta \cup t}(\cdot)$ with different $t$ (see \cref{fig:grad}) and find that the loss gradient similarity within $\mathcal{SM}_{t}^{t+l}$ increases as $t$ grows. To eliminate the discrepancy of the loss gradient similarity, we set $l$ to 1. We then construct MC-based AEs on $\mathcal{SM}_{t}^{t+1}$ and test these AEs. In \cref{tab:gv}, we find that AdvDM$_{999}^{1000}$, which uses smoother $\epsilon_{\theta \cup t}(\cdot)$, generates far better AEs than AdvDM$_{499}^{500}$ and AdvDM$_{99}^{100}$. This result demonstrates that higher loss gradient similarity is not the main factor improving the performance of AdvDM$_{t}^{t+100}$ with larger $t$.

\subsection{Fine-tuning Task} \label{sec:finetune}
\begin{figure}[!t]
	\centering
	\includegraphics[width=\linewidth]{./figure/loss_dfidLegend.pdf}\\
	\begin{minipage}{\linewidth}
		\centering
		\subfloat[Cat]{
			\includegraphics[width=0.48\linewidth]{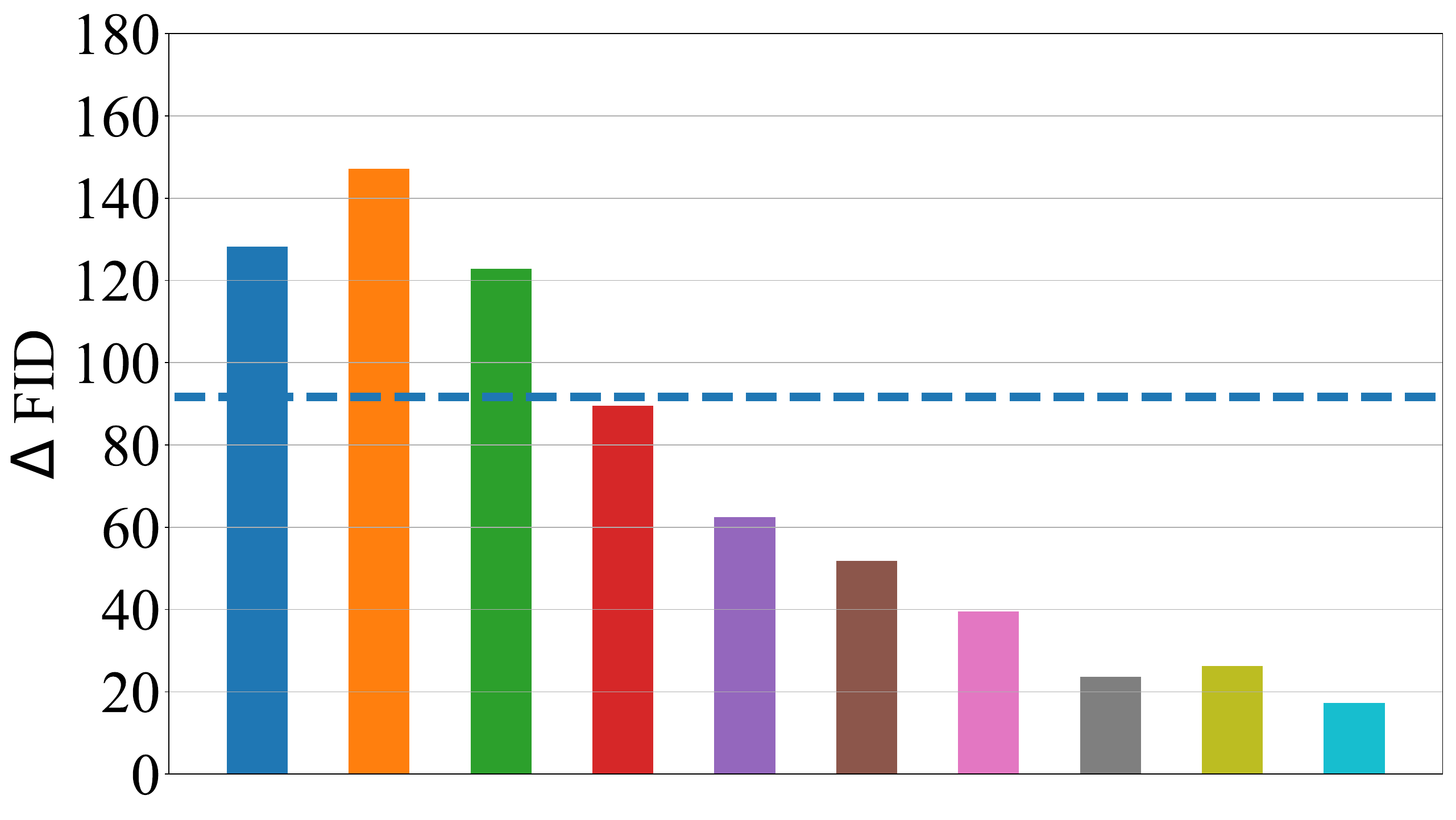}
		}
		\subfloat[Airplane]{
			\includegraphics[width=0.48\linewidth]{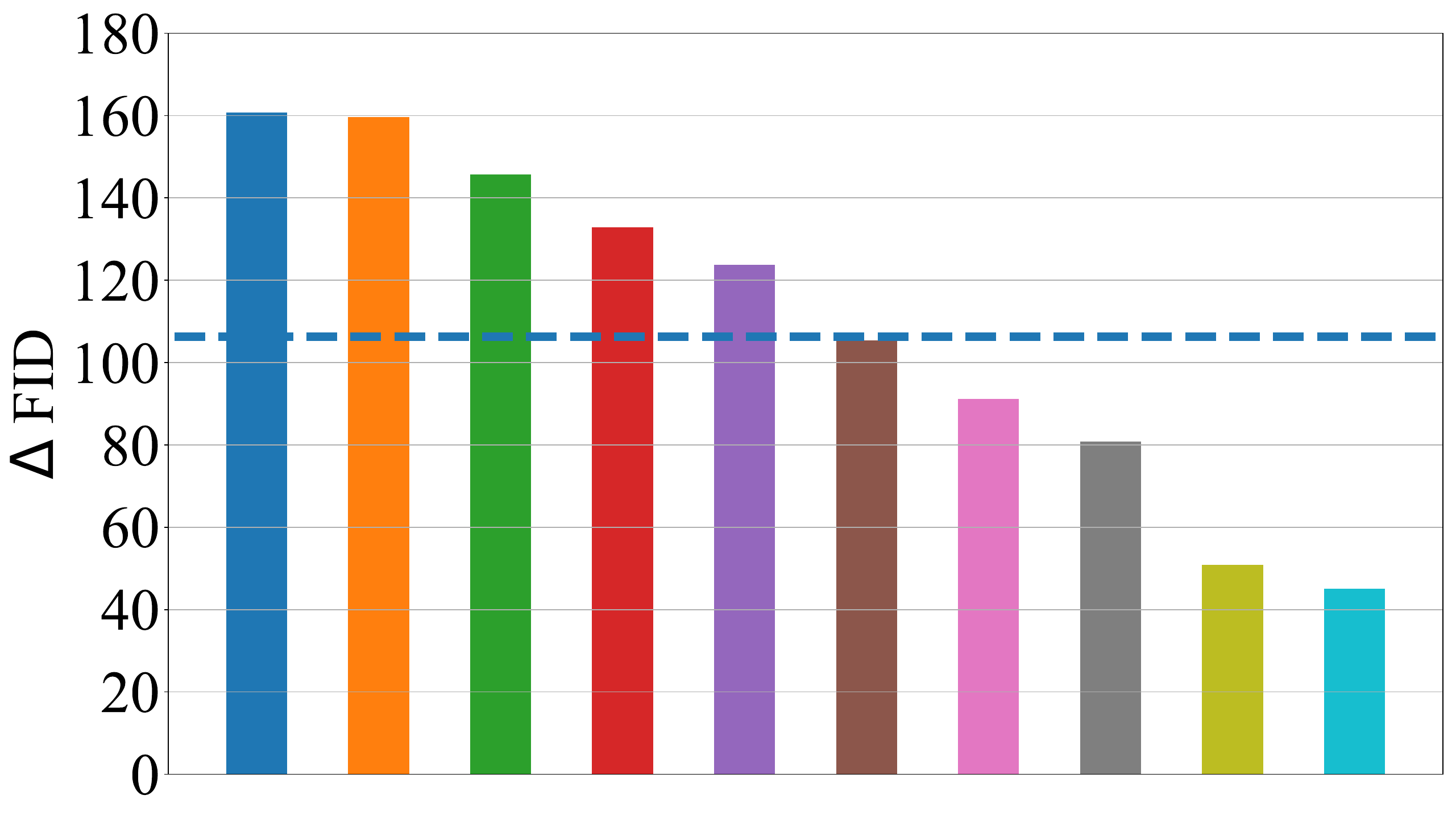}
		}
	\end{minipage}
	\caption{The increments $\uparrow$ of FID relative to the benign's. Larger increment means worse generated images and better AEs for textual inversion.} \label{fig:tiFID}
\end{figure}
We also test the performance of AEs generated on different $\mathcal{SM}_{a}^{b}$ in disrupting textual inversion. In \cref{fig:tiFID}, we can find that the performance of AEs generated on $\mathcal{SM}_{a}^{b}$ with larger $a$ and $b$ is worse, which is contrary to the results of image variation and inpainting. We mentioned in \cref{sec:modelTasks} that the fine-tuning task can not be seen as an end-to-end inference model because the fine-tuning task includes an optimizing process. In this sense, AEs for cracking the fine-tuning task act as poisoned data~\cite{Unlearnable, dataPoisonAdv} rather than carriers for adversarial attacks~\cite{goodfellowADV}. Consequently, our analysis based on end-to-end inference models can not explain the results of the fine-tuning task.

One reasonable explanation comes from a very recent work~\cite{simac}. \citet{simac} found that $\epsilon_{\theta \cup t}(\cdot)$ with smaller $t$ focuses on the higher frequency components of images while $\epsilon_{\theta \cup t}(\cdot)$ with larger $t$ focuses on the lower frequency. \citet{simac} claimed that introducing adversarial noises at larger time steps is ineffective because the subtle changes perturbed on images can hardly affect the low-frequency of the generated images. These findings may explain why sampling larger $t$ can not boost AEs for textual inversion. Moreover, these findings can also explain the results in \cref{fig:dfidInpaintMore}\subref{subfig:2toi1} and \cref{fig:dfidInpaintMore}\subref{subfig:2toi2}, where the performance of AEs generated on SD-v2-1 degrades as $t$ grows, because the smoothness of SD-v2-1 does not alter significantly over $t$. This property allows the factor found by \citet{simac} to dominate the results.

\section{Conclusion}

In this paper, we model the MC-based adversarial attack for the inference task as a transfer-based adversarial attack wherein the property of the surrogate model significantly impacts the adversarial transferability. We view the time-step sampling in AdvDM~\cite{AdvDM} as selecting different surrogate models and find that the smoothness of surrogate models at different time steps varies. In some LDMs, such variety is influential enough to alter the performance of AdvDM. We limit the range of time-step sampling to where surrogate models are smooth and boost the performance of AdvDM. Based on the theoretical framework~\cite{transferTheoryNIPS, transferTheorySP} for adversarial transferability in the image classification task, we conduct a theoretical study and claim that smooth surrogate models can also improve the performance of AEs for LDMs. Moreover, AEs with better performance in breaking the inference task may not work well in cracking the fine-tuning task, revealing the discrepancy and commonalities between AEs for the inference task and the fine-tuning task.


\bibliography{example_paper}

\begin{thebibliography}{29}
\providecommand{\natexlab}[1]{#1}
\providecommand{\url}[1]{\texttt{#1}}
\expandafter\ifx\csname urlstyle\endcsname\relax
  \providecommand{\doi}[1]{doi: #1}\else
  \providecommand{\doi}{doi: \begingroup \urlstyle{rm}\Url}\fi

\bibitem[Bianchi et~al.(2021)Bianchi, Attanasio, Pisoni, Terragni, Sarti, and
  Lakshmi]{CLIP}
Bianchi, F., Attanasio, G., Pisoni, R., Terragni, S., Sarti, G., and Lakshmi,
  S.
\newblock Contrastive language-image pre-training for the italian language.
\newblock \emph{CoRR}, abs/2108.08688, 2021.
\newblock URL \url{https://arxiv.org/abs/2108.08688}.

\bibitem[Chen et~al.(2023)Chen, Yin, Chen, Chen, and Liu]{ensemble2}
Chen, B., Yin, J., Chen, S., Chen, B., and Liu, X.
\newblock An adaptive model ensemble adversarial attack for boosting
  adversarial transferability.
\newblock \emph{CoRR}, abs/2308.02897, 2023.
\newblock \doi{10.48550/ARXIV.2308.02897}.
\newblock URL \url{https://doi.org/10.48550/arXiv.2308.02897}.

\bibitem[Fowl et~al.(2021)Fowl, Goldblum, Chiang, Geiping, Czaja, and
  Goldstein]{dataPoisonAdv}
Fowl, L., Goldblum, M., Chiang, P., Geiping, J., Czaja, W., and Goldstein, T.
\newblock Adversarial examples make strong poisons.
\newblock In Ranzato, M., Beygelzimer, A., Dauphin, Y.~N., Liang, P., and
  Vaughan, J.~W. (eds.), \emph{Advances in Neural Information Processing
  Systems 34: Annual Conference on Neural Information Processing Systems 2021,
  NeurIPS 2021, December 6-14, 2021, virtual}, pp.\  30339--30351, 2021.
\newblock URL
  \url{https://proceedings.neurips.cc/paper/2021/hash/fe87435d12ef7642af67d9bc82a8b3cd-Abstract.html}.

\bibitem[Gal et~al.(2023)Gal, Alaluf, Atzmon, Patashnik, Bermano, Chechik, and
  Cohen{-}Or]{textualInversion}
Gal, R., Alaluf, Y., Atzmon, Y., Patashnik, O., Bermano, A.~H., Chechik, G.,
  and Cohen{-}Or, D.
\newblock An image is worth one word: Personalizing text-to-image generation
  using textual inversion.
\newblock In \emph{The Eleventh International Conference on Learning
  Representations, {ICLR} 2023, Kigali, Rwanda, May 1-5, 2023}. OpenReview.net,
  2023.
\newblock URL \url{https://openreview.net/pdf?id=NAQvF08TcyG}.

\bibitem[Goodfellow et~al.(2015)Goodfellow, Shlens, and Szegedy]{goodfellowADV}
Goodfellow, I.~J., Shlens, J., and Szegedy, C.
\newblock Explaining and harnessing adversarial examples.
\newblock In Bengio, Y. and LeCun, Y. (eds.), \emph{3rd International
  Conference on Learning Representations, {ICLR} 2015, San Diego, CA, USA, May
  7-9, 2015, Conference Track Proceedings}, 2015.
\newblock URL \url{http://arxiv.org/abs/1412.6572}.

\bibitem[Gubri et~al.(2022)Gubri, Cordy, Papadakis, Traon, and Sen]{LGVtrans}
Gubri, M., Cordy, M., Papadakis, M., Traon, Y.~L., and Sen, K.
\newblock {LGV:} boosting adversarial example transferability from large
  geometric vicinity.
\newblock In Avidan, S., Brostow, G.~J., Ciss{\'{e}}, M., Farinella, G.~M., and
  Hassner, T. (eds.), \emph{Computer Vision - {ECCV} 2022 - 17th European
  Conference, Tel Aviv, Israel, October 23-27, 2022, Proceedings, Part {IV}},
  volume 13664 of \emph{Lecture Notes in Computer Science}, pp.\  603--618.
  Springer, 2022.
\newblock \doi{10.1007/978-3-031-19772-7\_35}.
\newblock URL \url{https://doi.org/10.1007/978-3-031-19772-7\_35}.

\bibitem[Guo et~al.(2018)Guo, Rana, Ciss{\'{e}}, and van~der Maaten]{tvm}
Guo, C., Rana, M., Ciss{\'{e}}, M., and van~der Maaten, L.
\newblock Countering adversarial images using input transformations.
\newblock In \emph{6th International Conference on Learning Representations,
  {ICLR} 2018, Vancouver, BC, Canada, April 30 - May 3, 2018, Conference Track
  Proceedings}. OpenReview.net, 2018.
\newblock URL \url{https://openreview.net/forum?id=SyJ7ClWCb}.

\bibitem[Heusel et~al.(2017)Heusel, Ramsauer, Unterthiner, Nessler, and
  Hochreiter]{FID}
Heusel, M., Ramsauer, H., Unterthiner, T., Nessler, B., and Hochreiter, S.
\newblock Gans trained by a two time-scale update rule converge to a local nash
  equilibrium.
\newblock In Guyon, I., von Luxburg, U., Bengio, S., Wallach, H.~M., Fergus,
  R., Vishwanathan, S. V.~N., and Garnett, R. (eds.), \emph{Advances in Neural
  Information Processing Systems 30: Annual Conference on Neural Information
  Processing Systems 2017, December 4-9, 2017, Long Beach, CA, {USA}}, pp.\
  6626--6637, 2017.
\newblock URL
  \url{https://proceedings.neurips.cc/paper/2017/hash/8a1d694707eb0fefe65871369074926d-Abstract.html}.

\bibitem[Huang et~al.(2021)Huang, Ma, Erfani, Bailey, and Wang]{Unlearnable}
Huang, H., Ma, X., Erfani, S.~M., Bailey, J., and Wang, Y.
\newblock Unlearnable examples: Making personal data unexploitable.
\newblock In \emph{9th International Conference on Learning Representations,
  {ICLR} 2021, Virtual Event, Austria, May 3-7, 2021}. OpenReview.net, 2021.
\newblock URL \url{https://openreview.net/forum?id=iAmZUo0DxC0}.

\bibitem[Le et~al.(2023)Le, Phung, Nguyen, Dao, Tran, and Tran]{AntiDB}
Le, T.~V., Phung, H., Nguyen, T.~H., Dao, Q., Tran, N., and Tran, A.
\newblock Anti-dreambooth: Protecting users from personalized text-to-image
  synthesis.
\newblock \emph{CoRR}, abs/2303.15433, 2023.
\newblock \doi{10.48550/ARXIV.2303.15433}.
\newblock URL \url{https://doi.org/10.48550/arXiv.2303.15433}.

\bibitem[Li et~al.(2020)Li, Bai, Zhou, Xie, Zhang, and Yuille]{ghostNet}
Li, Y., Bai, S., Zhou, Y., Xie, C., Zhang, Z., and Yuille, A.~L.
\newblock Learning transferable adversarial examples via ghost networks.
\newblock In \emph{The Thirty-Fourth {AAAI} Conference on Artificial
  Intelligence, {AAAI} 2020, The Thirty-Second Innovative Applications of
  Artificial Intelligence Conference, {IAAI} 2020, The Tenth {AAAI} Symposium
  on Educational Advances in Artificial Intelligence, {EAAI} 2020, New York,
  NY, USA, February 7-12, 2020}, pp.\  11458--11465. {AAAI} Press, 2020.
\newblock \doi{10.1609/AAAI.V34I07.6810}.
\newblock URL \url{https://doi.org/10.1609/aaai.v34i07.6810}.

\bibitem[Liang \& Wu(2023)Liang and Wu]{mist}
Liang, C. and Wu, X.
\newblock Mist: Towards improved adversarial examples for diffusion models.
\newblock \emph{CoRR}, abs/2305.12683, 2023.
\newblock \doi{10.48550/ARXIV.2305.12683}.
\newblock URL \url{https://doi.org/10.48550/arXiv.2305.12683}.

\bibitem[Liang et~al.(2023)Liang, Wu, Hua, Zhang, Xue, Song, Xue, Ma, and
  Guan]{AdvDM}
Liang, C., Wu, X., Hua, Y., Zhang, J., Xue, Y., Song, T., Xue, Z., Ma, R., and
  Guan, H.
\newblock Adversarial example does good: Preventing painting imitation from
  diffusion models via adversarial examples.
\newblock In Krause, A., Brunskill, E., Cho, K., Engelhardt, B., Sabato, S.,
  and Scarlett, J. (eds.), \emph{International Conference on Machine Learning,
  {ICML} 2023, 23-29 July 2023, Honolulu, Hawaii, {USA}}, volume 202 of
  \emph{Proceedings of Machine Learning Research}, pp.\  20763--20786. {PMLR},
  2023.
\newblock URL \url{https://proceedings.mlr.press/v202/liang23g.html}.

\bibitem[Madry et~al.(2018)Madry, Makelov, Schmidt, Tsipras, and Vladu]{PGD}
Madry, A., Makelov, A., Schmidt, L., Tsipras, D., and Vladu, A.
\newblock Towards deep learning models resistant to adversarial attacks.
\newblock In \emph{6th International Conference on Learning Representations,
  {ICLR} 2018, Vancouver, BC, Canada, April 30 - May 3, 2018, Conference Track
  Proceedings}. OpenReview.net, 2018.
\newblock URL \url{https://openreview.net/forum?id=rJzIBfZAb}.

\bibitem[Rombach et~al.(2022)Rombach, Blattmann, Lorenz, Esser, and Ommer]{LDM}
Rombach, R., Blattmann, A., Lorenz, D., Esser, P., and Ommer, B.
\newblock High-resolution image synthesis with latent diffusion models.
\newblock In \emph{{IEEE/CVF} Conference on Computer Vision and Pattern
  Recognition, {CVPR} 2022, New Orleans, LA, USA, June 18-24, 2022}, pp.\
  10674--10685. {IEEE}, 2022.
\newblock \doi{10.1109/CVPR52688.2022.01042}.
\newblock URL \url{https://doi.org/10.1109/CVPR52688.2022.01042}.

\bibitem[Salimans et~al.(2016)Salimans, Goodfellow, Zaremba, Cheung, Radford,
  and Chen]{IS}
Salimans, T., Goodfellow, I.~J., Zaremba, W., Cheung, V., Radford, A., and
  Chen, X.
\newblock Improved techniques for training gans.
\newblock In Lee, D.~D., Sugiyama, M., von Luxburg, U., Guyon, I., and Garnett,
  R. (eds.), \emph{Advances in Neural Information Processing Systems 29: Annual
  Conference on Neural Information Processing Systems 2016, December 5-10,
  2016, Barcelona, Spain}, pp.\  2226--2234, 2016.
\newblock URL
  \url{https://proceedings.neurips.cc/paper/2016/hash/8a3363abe792db2d8761d6403605aeb7-Abstract.html}.

\bibitem[Salman et~al.(2023)Salman, Khaddaj, Leclerc, Ilyas, and
  Madry]{photoGuard}
Salman, H., Khaddaj, A., Leclerc, G., Ilyas, A., and Madry, A.
\newblock Raising the cost of malicious ai-powered image editing.
\newblock In Krause, A., Brunskill, E., Cho, K., Engelhardt, B., Sabato, S.,
  and Scarlett, J. (eds.), \emph{International Conference on Machine Learning,
  {ICML} 2023, 23-29 July 2023, Honolulu, Hawaii, {USA}}, volume 202 of
  \emph{Proceedings of Machine Learning Research}, pp.\  29894--29918. {PMLR},
  2023.
\newblock URL \url{https://proceedings.mlr.press/v202/salman23a.html}.

\bibitem[Segura et~al.(2023)Segura, Geiping, and Goldstein]{jpgDefense}
Segura, P.~S., Geiping, J., and Goldstein, T.
\newblock {JPEG} compressed images can bypass protections against {AI} editing.
\newblock \emph{CoRR}, abs/2304.02234, 2023.
\newblock \doi{10.48550/ARXIV.2304.02234}.
\newblock URL \url{https://doi.org/10.48550/arXiv.2304.02234}.

\bibitem[Springer et~al.(2021)Springer, Mitchell, and Kenyon]{advtrainImprove}
Springer, J.~M., Mitchell, M., and Kenyon, G.~T.
\newblock A little robustness goes a long way: Leveraging robust features for
  targeted transfer attacks.
\newblock In Ranzato, M., Beygelzimer, A., Dauphin, Y.~N., Liang, P., and
  Vaughan, J.~W. (eds.), \emph{Advances in Neural Information Processing
  Systems 34: Annual Conference on Neural Information Processing Systems 2021,
  NeurIPS 2021, December 6-14, 2021, virtual}, pp.\  9759--9773, 2021.
\newblock URL
  \url{https://proceedings.neurips.cc/paper/2021/hash/50f3f8c42b998a48057e9d33f4144b8b-Abstract.html}.

\bibitem[von Platen et~al.(2022)von Platen, Patil, Lozhkov, Cuenca, Lambert,
  Rasul, Davaadorj, and Wolf]{diffusers}
von Platen, P., Patil, S., Lozhkov, A., Cuenca, P., Lambert, N., Rasul, K.,
  Davaadorj, M., and Wolf, T.
\newblock Diffusers: State-of-the-art diffusion models.
\newblock \url{https://github.com/huggingface/diffusers}, 2022.

\bibitem[{Wang} et~al.(2023){Wang}, {Tan}, {Wei}, {Wu}, and {Huang}]{simac}
{Wang}, F., {Tan}, Z., {Wei}, T., {Wu}, Y., and {Huang}, Q.
\newblock {SimAC: A Simple Anti-Customization Method against Text-to-Image
  Synthesis of Diffusion Models}.
\newblock \emph{arXiv e-prints}, art. arXiv:2312.07865, December 2023.
\newblock \doi{10.48550/arXiv.2312.07865}.

\bibitem[Wang et~al.(2023)Wang, Chan, and Loy]{CLIPIQA}
Wang, J., Chan, K. C.~K., and Loy, C.~C.
\newblock Exploring {CLIP} for assessing the look and feel of images.
\newblock In Williams, B., Chen, Y., and Neville, J. (eds.),
  \emph{Thirty-Seventh {AAAI} Conference on Artificial Intelligence, {AAAI}
  2023, Thirty-Fifth Conference on Innovative Applications of Artificial
  Intelligence, {IAAI} 2023, Thirteenth Symposium on Educational Advances in
  Artificial Intelligence, {EAAI} 2023, Washington, DC, USA, February 7-14,
  2023}, pp.\  2555--2563. {AAAI} Press, 2023.
\newblock \doi{10.1609/AAAI.V37I2.25353}.
\newblock URL \url{https://doi.org/10.1609/aaai.v37i2.25353}.

\bibitem[Xiong et~al.(2022)Xiong, Lin, Zhang, Hopcroft, and He]{ensemble3}
Xiong, Y., Lin, J., Zhang, M., Hopcroft, J.~E., and He, K.
\newblock Stochastic variance reduced ensemble adversarial attack for boosting
  the adversarial transferability.
\newblock In \emph{{IEEE/CVF} Conference on Computer Vision and Pattern
  Recognition, {CVPR} 2022, New Orleans, LA, USA, June 18-24, 2022}, pp.\
  14963--14972. {IEEE}, 2022.
\newblock \doi{10.1109/CVPR52688.2022.01456}.
\newblock URL \url{https://doi.org/10.1109/CVPR52688.2022.01456}.

\bibitem[Xue et~al.(2023)Xue, Liang, Wu, and Chen]{sds}
Xue, H., Liang, C., Wu, X., and Chen, Y.
\newblock Toward effective protection against diffusion based mimicry through
  score distillation.
\newblock \emph{CoRR}, abs/2311.12832, 2023.
\newblock \doi{10.48550/ARXIV.2311.12832}.
\newblock URL \url{https://doi.org/10.48550/arXiv.2311.12832}.

\bibitem[Yang et~al.(2021)Yang, Li, Xu, Zuo, Chen, Zhou, Rubinstein, Zhang, and
  Li]{transferTheoryNIPS}
Yang, Z., Li, L., Xu, X., Zuo, S., Chen, Q., Zhou, P., Rubinstein, B. I.~P.,
  Zhang, C., and Li, B.
\newblock {TRS:} transferability reduced ensemble via promoting gradient
  diversity and model smoothness.
\newblock In Ranzato, M., Beygelzimer, A., Dauphin, Y.~N., Liang, P., and
  Vaughan, J.~W. (eds.), \emph{Advances in Neural Information Processing
  Systems 34: Annual Conference on Neural Information Processing Systems 2021,
  NeurIPS 2021, December 6-14, 2021, virtual}, pp.\  17642--17655, 2021.
\newblock URL
  \url{https://proceedings.neurips.cc/paper/2021/hash/937936029af671cf479fa893db91cbdd-Abstract.html}.

\bibitem[Yu et~al.(2015)Yu, Zhang, Song, Seff, and Xiao]{lsun}
Yu, F., Zhang, Y., Song, S., Seff, A., and Xiao, J.
\newblock {LSUN:} construction of a large-scale image dataset using deep
  learning with humans in the loop.
\newblock \emph{CoRR}, abs/1506.03365, 2015.
\newblock URL \url{http://arxiv.org/abs/1506.03365}.

\bibitem[Zhang et~al.(2023{\natexlab{a}})Zhang, Xu, Cui, Meng, Wu, and
  Lyu]{benchmark}
Zhang, J., Xu, Z., Cui, S., Meng, C., Wu, W., and Lyu, M.~R.
\newblock On the robustness of latent diffusion models.
\newblock \emph{CoRR}, abs/2306.08257, 2023{\natexlab{a}}.
\newblock \doi{10.48550/ARXIV.2306.08257}.
\newblock URL \url{https://doi.org/10.48550/arXiv.2306.08257}.

\bibitem[Zhang et~al.(2023{\natexlab{b}})Zhang, Hu, Zhang, Shi, Li, Liu, Wan,
  and Jin]{transferTheorySP}
Zhang, Y., Hu, S., Zhang, L., Shi, J., Li, M., Liu, X., Wan, W., and Jin, H.
\newblock Why does little robustness help? a further step towards understanding
  adversarial transferability, 07 2023{\natexlab{b}}.

\bibitem[Zhao et~al.(2021)Zhao, Liu, and Larson]{moreIter}
Zhao, Z., Liu, Z., and Larson, M.~A.
\newblock On success and simplicity: {A} second look at transferable targeted
  attacks.
\newblock In Ranzato, M., Beygelzimer, A., Dauphin, Y.~N., Liang, P., and
  Vaughan, J.~W. (eds.), \emph{Advances in Neural Information Processing
  Systems 34: Annual Conference on Neural Information Processing Systems 2021,
  NeurIPS 2021, December 6-14, 2021, virtual}, pp.\  6115--6128, 2021.
\newblock URL
  \url{https://proceedings.neurips.cc/paper/2021/hash/30d454f09b771b9f65e3eaf6e00fa7bd-Abstract.html}.

\end{thebibliography}
\bibliographystyle{icml2024}

\newpage
\appendix
\onecolumn
\section{Experimental Setups}

\subsection{Setups for \cref{sec:empiricalStudy}} \label{sec:empiricalSetup}
The LDMs we choose are SD-v1-4 \footnote{https://huggingface.co/CompVis/stable-diffusion-v-1-4-original \label{fn:sdv14}}, SD-v1-5 \footnote{https://huggingface.co/runwayml/stable-diffusion-v1-5 \label{fn:sdv15}}, and SD-v2-1 \footnote{https://huggingface.co/stabilityai/stable-diffusion-2-1 \label{fn:sdv21}}. We choose
the dataset released by \citet{benchmark} to perform our experiments. AEs are all generated by $l_{\infty}$ PGD~\cite{PGD} with $\varepsilon=8/255$, $\text{steps} = 40$, and $\text{step size} = 1/255$, which is of the same setting as \citet{AdvDM}.
\subsection{Setups for \cref{sec:experiment}} \label{sec:experSetup}

\subsubsection{Datasets}
We use the dataset released by \citet{benchmark} to perform our experiments on image variation task and image inpainting task. This dataset contains 100 images for each task, and each image has 5 prompts for generation. The resolution of each image is $512\times512$.

Following \citet{AdvDM}, we choose LSUN-airplane and LSUN-cat~\cite{lsun} for textual inversion. Observing that LSUN-airplane and LSUN-cat are noisy, where airplane debris and texts are also included, we use CLIP~\cite{CLIP} to choose 200 high-quality images. For example, we calculate the similarity between the prompt ``cat'' and images in LSUN-cat and choose the top-200 images having higher similarity to the prompt. Lastly, we resize and center crop the image to $256 \times 256$.

\subsubsection{LDMs}
We use SD-v1-4 \footref{fn:sdv14}, SD-v1-5 \footref{fn:sdv15}, and SD-v2-1 \footref{fn:sdv21} as our surrogate models for generating MC-based~\cite{AdvDM, sds} and encoder-based AEs~\cite{mist}. Apart from the above three LDMs, we use SD-Inpainting \footnote{https://huggingface.co/runwayml/stable-diffusion-inpainting \label{fn:sdI}} and SD-2-Inpainting \footnote{https://huggingface.co/stabilityai/stable-diffusion-2-inpainting \label{fn:sd2I}} to generate chain-based AEs~\cite{benchmark} for image inpainting, which is of the same setting to \citet{benchmark}.

We use SD-v1-4, SD-v1-5, SD-v2-1, and InstructPix2Pix \footnote{https://huggingface.co/timbrooks/instruct-pix2pix} as target models for image variation. We use SD-Inpainting \footref{fn:sdI} and SD-2-Inpainting \footref{fn:sd2I} as target models for image inpainting.

We use SD-v1-5 to conduct textual inversion.

\subsubsection{Baselines}
We choose three types of adversarial attack as our baselines.

We choose AdvDM~\cite{AdvDM} and SDS with gradient descent~\cite{sds} to represent the MC-based adversarial attack. For the encoder-based adversarial attack, we choose Mist~\cite{mist} with the same target image provided by \citet{mist}. For the chain-based adversarial attack, we choose the algorithm proposed by \citet{benchmark}, which can be seen as an elaborately tuned photoguard~\cite{photoGuard}.

AdvDM$_{a}^{b}$ represents AdvDM that samples time steps $t$ from $u(a + 1, b)$, and $\text{AdvDM} = \text{AdvDM}_{0}^{1000}$. Mist's loss includes \cref{eq:encoder} and \cref{eq:advdm} and multiplies \cref{eq:advdm} by a fuse weight $w$, which can be seen as a combination of the encoder-based attack and AdvDM. For conciseness, we denote Mist $(w)$ as Mist with fuse weight $w$ wherein Mist $(0)$ is a pure encoder-based attack. We also try to use our modified AdvDM to replace the vanilla AdvDM in Mist and denote Mist$_{a}^{b} (w)$ as the combination of the encoder-based attack and AdvDM$_{a}^{b}$ with fuse weight $w$. 

Considering only one chain-based attack is included, we denote the algorithm proposed by \citet{benchmark} as \emph{Chain}.

\subsubsection{Hyper-parameters of Adversarial Attacks}

All adversarial attacks use $l_{\infty}$ PGD~\cite{PGD} to optimize the adversarial perturbation. We set perturbation budget $\varepsilon = 8/255$, iteration steps to $40$, and step size to $1/255$. Following the guidelines in the repository of \citet{mist}, we set the fuse weight of Mist to 1 and 5. As for Chain, \citet{benchmark} claimed that including more denoising steps can improve the performance of AEs. We gradually increase the denoising steps and find that we can, at most, include 5 denoising steps on a single A100 40GB. The rest of the settings of Chain is the same as the optimum setting provided by \citet{benchmark}. Note that \citet{benchmark} found attacking the encoder is the optimum for constructing AEs for image inpainting, which makes Chain an encoder-based adversarial attack.

\subsubsection{Hyper-parameters of Image Generation }
We use diffusers~\cite{diffusers} to conduct image variation and inpainting. Following \citet{benchmark} and \citet{photoGuard}, we set edit strength to 0.7, denoising steps to 100, and guidance scale to 7.5. InstructPix2Pix does not have the edit strength, and we set denoising steps to 100 and guidance scale to 7.5, which is the default setting provided by diffusers~\cite{diffusers}.

The script of textual inversion is also provided by diffusers~\cite{diffusers}. We partition 200 images into 40 5-image groups and obtain 40 pseudo words $S^*$. We set the batch size to 4 and left the rest of the parameters as default. After fine-tuning the pseudo word $S^*$, we use the prompt ``a good photo of a <sks>'' to generate 50 images per pseudo word $S^*$ and acquire 2000 images.

\subsubsection{Metrics}
Following \citet{benchmark}, we use FID~\cite{FID} to measure the distributional discrepancy between the generated images and the original images. High FID means that the distribution of the generated images is far from the distribution of the original images and that the adversary fails to conduct imitations.

Apart from preventing imitations, we can also corrupt the image variation and inpainting by degrading the generated images' quality, making the generated images noisy or unnatural. To measure the quality of the generated images, we use Inception Score (IS)~\cite{IS} and CLIP-IQA~\cite{CLIPIQA}. Specifically, the prompt pairs used in CLIP-IQA are ```Clean photo' vs`Noisy photo.''' and ```Natural photo.' vs `Synthetic photo.'''. Lower values of IS and CLIP-IQA mean worse quality of the generated images.

Lastly, since the adversary uses prompts to edit the image, we also use CLIP~\cite{CLIP} to measure the similarity between the prompt and the generated image. Lower CLIP values mean the semantics of the generated image fails to meet the adversary's need.

\section{Proof of \cref{the:lowBound}}\label{sec:proof}
First, we provide a lemma that we will use for the proof of \cref{the:lowBound}.

\begin{lemma}\label{lemma:1}
	For arbitrary vectors $\delta$, $x$, and $y$, suppose $\left\| \delta \right\|_{2} \le \varepsilon$, $\left\| x \right\|_{2} = \left\| y \right\|_{2} = 1$, and $c \in \mathbb{R}$. We have
	\begin{equation}
		\delta \cdot y \le c - \epsilon \sqrt{2 - 2\cos\left \langle x, y \right \rangle } \Rightarrow \delta \cdot x \le c\,,
	\end{equation}
	where $\cos\left \langle x, y \right \rangle  = \frac{x \cdot y}{\left\| x \right\|_{2} \cdot \left\| y \right\|_{2}}$.
	\begin{proof}
		$\delta \cdot x = \delta \cdot y + \delta \cdot (x - y) \le c - \epsilon \sqrt{2 - 2\cos\left \langle x, y \right \rangle } + \delta \dot (x - y)$. Applying the law of cosine, we have $\delta \dot (x - y) \le \left\| \delta \right\|_{2} \cdot \left\| x - y \right\|_{2} \le \varepsilon \sqrt{2 - 2\cos\left \langle x, y \right \rangle }$. Thus, $\delta \cdot x \le c$. 
	\end{proof}
\end{lemma}

Below, we provide the proof of \cref{the:lowBound}. The main differences between ours and \citet{transferTheoryNIPS, transferTheorySP} lie in \cref{eq:diff1} and \cref{eq:diff2}, which are derived from \cref{def:risk} and \cref{def:transfer}.

\begin{proof}
	First, we define two auxiliary functions $f$ and $g$ such that
	\begin{align}
		&f(x, y) = \frac{L_{1} - \mathcal{L}_{\mathcal{F}}(x, y) - \beta\varepsilon^{2}/2}{\left\| \nabla_{x}\mathcal{L}_{\mathcal{F}}(x, y) \right\|_{2}}\,,\\
		&g(x, y) = \frac{L_{2} - \mathcal{L}_{\mathcal{G}}(x, y) + \beta\varepsilon^{2}/2}{\left\| \nabla_{x}\mathcal{L}_{\mathcal{G}}(x, y) \right\|_{2}}\,.
	\end{align}
	Recall \cref{eq:tool1} and \cref{eq:tool2}. We have $c_{\mathcal{F}} = \min_{x \in \mathcal{X}} f(x, y)$ and $c_{\mathcal{G}} = \max_{x \in \mathcal{X}} g(x, y)$.
	
	From \cref{def:transfer}, we have
	\begin{align}
		&\text{Pr}(T(F, G, x, y_{1}, y_{2}) = 1)\\
		= &\text{Pr}(\mathcal{L}_{\mathcal{F}}(x, y_{1}) \le L_{1} \wedge \mathcal{L}_{\mathcal{G}}(x, y_{2}) \le L_{2} \wedge  \mathcal{L}_{\mathcal{F}}(x', y_{1}) > L_{1} \wedge \mathcal{L}_{\mathcal{G}}(x', y_{2}) > L_{2} )\\
		= &1 - \text{Pr}(\mathcal{L}_{\mathcal{F}}(x, y_{1}) > L_{1} \vee \mathcal{L}_{\mathcal{G}}(x, y_{2}) > L_{2} \vee  \mathcal{L}_{\mathcal{F}}(x', y_{1}) \le L_{1} \vee \mathcal{L}_{\mathcal{G}}(x', y_{2}) \le L_{2})\\
		\ge &1 - \text{Pr}(\mathcal{L}_{\mathcal{F}}(x, y_{1}) > L_{1}) - \text{Pr}(\mathcal{L}_{\mathcal{G}}(x, y_{2}) > L_{2}) - \text{Pr}(\mathcal{L}_{\mathcal{F}}(x', y_{1}) \le L_{1}) - \text{Pr}(\mathcal{L}_{\mathcal{G}}(x', y_{2}) \le L_{2})\label{eq:inex}\\
		\ge & (1 - \alpha) - (\gamma_{\mathcal{F}} + \gamma_{\mathcal{G}}) - \text{Pr}(\mathcal{L}_{\mathcal{G}}(x', y_{2}) \le L_{2})\label{eq:secondInequality}\,,
	\end{align}
	where $x' = x + \delta$, \cref{eq:inex} is derived from the principle of inclusion-exclusion, and \cref{eq:secondInequality} is derived from \cref{def:alpha} and \cref{def:risk}.
	
	Applying Taylor's Theorem with Lagrange remainder, we have
	\begin{align}
		\mathcal{L}_{\mathcal{F}}(x+\delta, y_{1}) = \mathcal{L}_{\mathcal{F}}(x, y_{1}) + \delta \nabla_{x}\mathcal{L}_{\mathcal{F}}(x, y_{1}) + \frac{1}{2} \xi^{\text{T}}\mathbf{H}_{\mathcal{F}}\xi\,,\\
		\mathcal{L}_{\mathcal{G}}(x+\delta, y_{2}) = \mathcal{L}_{\mathcal{G}}(x, y_{2}) + \delta \nabla_{x}\mathcal{L}_{\mathcal{G}}(x, y_{2}) + \frac{1}{2} \xi^{\text{T}}\mathbf{H}_{\mathcal{G}}\xi\,,
	\end{align}
	where $\xi = k\delta$ with $k \in [0, 1]$, and $\mathbf{H}_{\mathcal{F}}$ and $\mathbf{H}_{\mathcal{G}}$ are Hessian matrices of $\mathcal{L}_{\mathcal{F}}(x, y_{1})$ and $\mathcal{L}_{\mathcal{G}}(x, y_{2})$ respectively. Since both $F$ and $G$ are $\beta$-smooth, the maximum eigenvalues of $\mathbf{H}_{\mathcal{F}}$ and $\mathbf{H}_{\mathcal{G}}$ are bounded by $\beta$. Thus, we have $|\xi^{\text{T}}\mathbf{H}\xi| \le \beta\left\| \xi \right\|_{2}^{2} \le \beta \varepsilon^{2}$. With this inequality, we have
	\begin{align}
	 &\mathcal{L}_{\mathcal{F}}(x, y_{1}) + \delta \nabla_{x}\mathcal{L}_{\mathcal{F}}(x, y_{1}) - \frac{1}{2}\beta \varepsilon^{2} \le \mathcal{L}_{\mathcal{F}}(x+\delta, y_{1}) \le \mathcal{L}_{\mathcal{F}}(x, y_{1}) + \delta \nabla_{x}\mathcal{L}_{\mathcal{F}}(x, y_{1}) + \frac{1}{2}\beta \varepsilon^{2} \,,\label{eq:ftaylor}\\
	 &\mathcal{L}_{\mathcal{G}}(x, y_{2}) + \delta \nabla_{x}\mathcal{L}_{\mathcal{G}}(x, y_{2}) - \frac{1}{2}\beta \varepsilon^{2} \le \mathcal{L}_{\mathcal{G}}(x+\delta, y_{2}) \le \mathcal{L}_{\mathcal{G}}(x, y_{2}) + \delta \nabla_{x}\mathcal{L}_{\mathcal{G}}(x, y_{2}) + \frac{1}{2}\beta \varepsilon^{2} \,,\label{eq:gtaylor}
	\end{align}
	
	Applying the right side of \cref{eq:ftaylor}, we have
	\begin{align}
		&\text{Pr}(\mathcal{L}_{\mathcal{F}}(x + \delta, y_{1}) \le L_{1})\label{eq:diff1}\\
	\ge &\text{Pr}(\mathcal{L}_{\mathcal{F}}(x, y_{1}) + \delta \nabla_{x}\mathcal{L}_{\mathcal{F}}(x, y_{1}) + \frac{1}{2}\beta \varepsilon^{2} \le L_{1})\\
	= & \text{Pr}(\delta \cdot \frac{\nabla_{x}\mathcal{L}_{\mathcal{F}}(x, y_{1})}{\left\| \nabla_{x}\mathcal{L}_{\mathcal{F}}(x, y_{1}) \right\|_{2}} \le f(x, y_{1}))\,,
	\end{align}
	and thus
	\begin{align}
		\text{Pr}(\mathcal{L}_{\mathcal{F}}(x + \delta, y_{1}) \le L_{1}) \le \alpha \Rightarrow \text{Pr}(\delta \cdot \frac{\nabla_{x}\mathcal{L}_{\mathcal{F}}(x, y_{1})}{\left\| \nabla_{x}\mathcal{L}_{\mathcal{F}}(x, y_{1}) \right\|_{2}} \le f(x, y_{1})) \le \alpha\,.
	\end{align}
	
	Similarly, apply the left side of \cref{eq:gtaylor} to $\text{Pr}(\mathcal{L}_{\mathcal{G}}(x', y_{2}) \le L_{2})$:
	\begin{align}
		&\text{Pr}(\mathcal{L}_{\mathcal{G}}(x', y_{2}) \le L_{2})\label{eq:diff2}\\
	\le &\text{Pr}(\mathcal{L}_{\mathcal{G}}(x, y_{2}) + \delta \nabla_{x}\mathcal{L}_{\mathcal{G}}(x, y_{2}) - \frac{1}{2}\beta \varepsilon^{2} \le L_{2})\\
	= & \text{Pr}(\delta \cdot\frac{\nabla_{x}\mathcal{L}_{\mathcal{G}}(x, y_{2})}{\left\| \nabla_{x}\mathcal{L}_{\mathcal{G}}(x, y_{2}) \right\|_{2}} \le g(x, y_{2}))\\
	\le & \text{Pr}(\delta \cdot \frac{\nabla_{x}\mathcal{L}_{\mathcal{G}}(x, y_{2})}{\left\| \nabla_{x}\mathcal{L}_{\mathcal{G}}(x, y_{2}) \right\|_{2}} \le c_{\mathcal{G}})\,.\label{eq:keyInequal}
	\end{align}
	
	Note that $\left\| \delta \right\| \le \varepsilon$ and that both $\frac{\nabla_{x}\mathcal{L}_{\mathcal{F}}(x, y_{1})}{\left\| \nabla_{x}\mathcal{L}_{\mathcal{F}}(x, y_{1}) \right\|_{2}}$ and $\frac{\nabla_{x}\mathcal{L}_{\mathcal{G}}(x, y_{2})}{\left\| \nabla_{x}\mathcal{L}_{\mathcal{G}}(x, y_{2}) \right\|_{2}}$ are unit vectors. Consequently, we apply \cref{lemma:1} and deduce
	\begin{align}
		&\delta \cdot \frac{\nabla_{x}\mathcal{L}_{\mathcal{G}}(x, y_{2})}{\left\| \nabla_{x}\mathcal{L}_{\mathcal{G}}(x, y_{2}) \right\|_{2}} \le f(x, y_{1}) - \varepsilon \sqrt{2 - 2\underline{\mathcal{S}}(\mathcal{L}_{\mathcal{F}}, \mathcal{L}_{\mathcal{G}}, x, y_{1}, y_{2})}\\
	\Rightarrow &\delta \cdot \frac{\nabla_{x}\mathcal{L}_{\mathcal{G}}(x, y_{2})}{\left\| \nabla_{x}\mathcal{L}_{\mathcal{G}}(x, y_{2}) \right\|_{2}} \le f(x, y_{1}) - \varepsilon \sqrt{2 - 2\cos\left\langle\nabla_{x}\mathcal{L}_{\mathcal{F}}(x, y_{1}),  \nabla_{x}\mathcal{L}_{\mathcal{G}}(x, y_{2})\right\rangle}\\
	\Rightarrow &\delta \cdot \frac{\nabla_{x}\mathcal{L}_{\mathcal{F}}(x, y_{1})}{\left\| \nabla_{x}\mathcal{L}_{\mathcal{F}}(x, y_{1}) \right\|_{2}} \le f(x, y_{1})\\
	\Rightarrow &\text{Pr}(\delta \cdot \frac{\nabla_{x}\mathcal{L}_{\mathcal{G}}(x, y_{2})}{\left\| \nabla_{x}\mathcal{L}_{\mathcal{G}}(x, y_{2}) \right\|_{2}} \le f(x, y_{1}) - \varepsilon \sqrt{2 - 2\underline{\mathcal{S}}(\mathcal{L}_{\mathcal{F}}, \mathcal{L}_{\mathcal{G}}, x, y_{1}, y_{2})}) \le \text{Pr}(\delta \cdot \frac{\nabla_{x}\mathcal{L}_{\mathcal{F}}(x, y_{1})}{\left\| \nabla_{x}\mathcal{L}_{\mathcal{F}}(x, y_{1}) \right\|_{2}} \le f(x, y_{1})) \le \alpha.
	\end{align}
	
	Since $c_{\mathcal{F}} = \min_{x \in \mathcal{X}} f(x, y_{1}) \le f(x, y_{1})$, we have
	\begin{align}
		\text{Pr}(\delta \cdot \frac{\nabla_{x}\mathcal{L}_{\mathcal{G}}(x, y_{2})}{\left\| \nabla_{x}\mathcal{L}_{\mathcal{G}}(x, y_{2}) \right\|_{2}} \le c_{\mathcal{F}} - \varepsilon \sqrt{2 - 2\underline{\mathcal{S}}(\mathcal{L}_{\mathcal{F}}, \mathcal{L}_{\mathcal{G}}, x, y_{1}, y_{2})}) \le \alpha\,.
	\end{align}
	Note that $\delta \cdot \frac{\nabla_{x}\mathcal{L}_{\mathcal{G}}(x, y_{2})}{\left\| \nabla_{x}\mathcal{L}_{\mathcal{G}}(x, y_{2}) \right\|_{2}} \ge -\varepsilon$. Thus, its expectation is bounded:
	\begin{align}
		\mathbb{E}[\delta \cdot \frac{\nabla_{x}\mathcal{L}_{\mathcal{G}}(x, y_{2})}{\left\| \nabla_{x}\mathcal{L}_{\mathcal{G}}(x, y_{2}) \right\|_{2}}] \ge -\varepsilon\alpha + (c_{\mathcal{F}} - \varepsilon \sqrt{2 - 2\underline{\mathcal{S}}(\mathcal{L}_{\mathcal{F}}, \mathcal{L}_{\mathcal{G}}, x, y_{1}, y_{2})})(1 - \alpha)\,.
	\end{align}
	Applying Markov's inequality and combining \cref{eq:keyInequal}, we have
	\begin{align}
		\text{Pr}(\mathcal{L}_{\mathcal{G}}(x', y_{2}) \le L_{2}) \le \text{Pr}(\delta \cdot \frac{\nabla_{x}\mathcal{L}_{\mathcal{G}}(x, y_{2})}{\left\| \nabla_{x}\mathcal{L}_{\mathcal{G}}(x, y_{2}) \right\|_{2}} \le c_{\mathcal{G}})
		 \le \frac{\varepsilon(1+\alpha) - (c_{\mathcal{F}} - \varepsilon \sqrt{2 - 2\underline{\mathcal{S}}(\mathcal{L}_{\mathcal{F}}, \mathcal{L}_{\mathcal{G}}, x, y_{1}, y_{2})})(1 - \alpha)}{\varepsilon - c_{\mathcal{G}}}\,,
	\end{align}
	and thus
	\begin{align}
		\text{Pr}(T(\mathcal{F}, \mathcal{G}, x, y_{1}, y_{2}) = 1) \ge (1 - \alpha) - (\gamma_{\mathcal{F}} + \gamma_{\mathcal{G}}) - \frac{\varepsilon(1 + \alpha) - c_{\mathcal{F}}(1 - \alpha)}{\varepsilon - c_{\mathcal{G}}} - \frac{\varepsilon(1 - \alpha)}{\varepsilon - c_{\mathcal{G}}}\sqrt{2 - 2\underline{\mathcal{S}}(\mathcal{L}_{\mathcal{F}}, \mathcal{L}_{\mathcal{G}}, x, y_{1}, y_{2})}
	\end{align}
\end{proof}

\newpage

\section{Additional Results Demonstrating Smoothness can Boost AEs for LDMs}\label{sec:moreSmoothness}
\begin{figure}[!h]
	\centering
	\includegraphics[width=0.9\linewidth]{./figure/loss_dfidLegend.pdf}\\
	\begin{minipage}{0.9\linewidth}
		\centering
		\subfloat[SD-v1-4 to SD-v1-4]{
			\includegraphics[width=0.24\linewidth]{./figure/dfid4to4.pdf}
		}
		\subfloat[SD-v1-4 to SD-v1-5]{\label{fig:difdIV4to5}
				\includegraphics[width=0.24\linewidth]{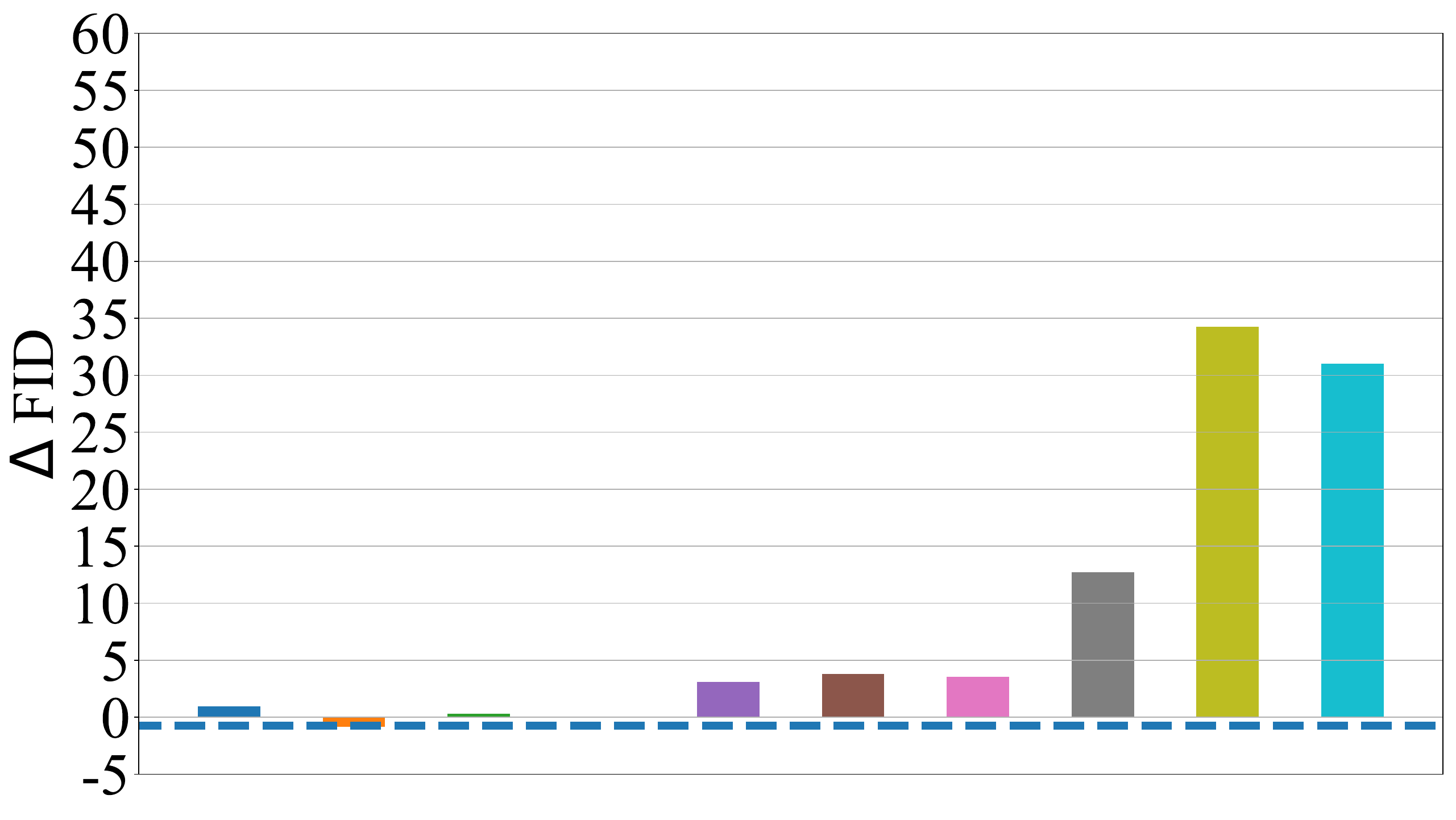}
				
		}
		\subfloat[SD-v1-4 to SD-v2-1]{
			\includegraphics[width=0.24\linewidth]{./figure/dfid4to2.pdf}
		}
		\subfloat[SD-v1-4 to Instruct]{
			\includegraphics[width=0.24\linewidth]{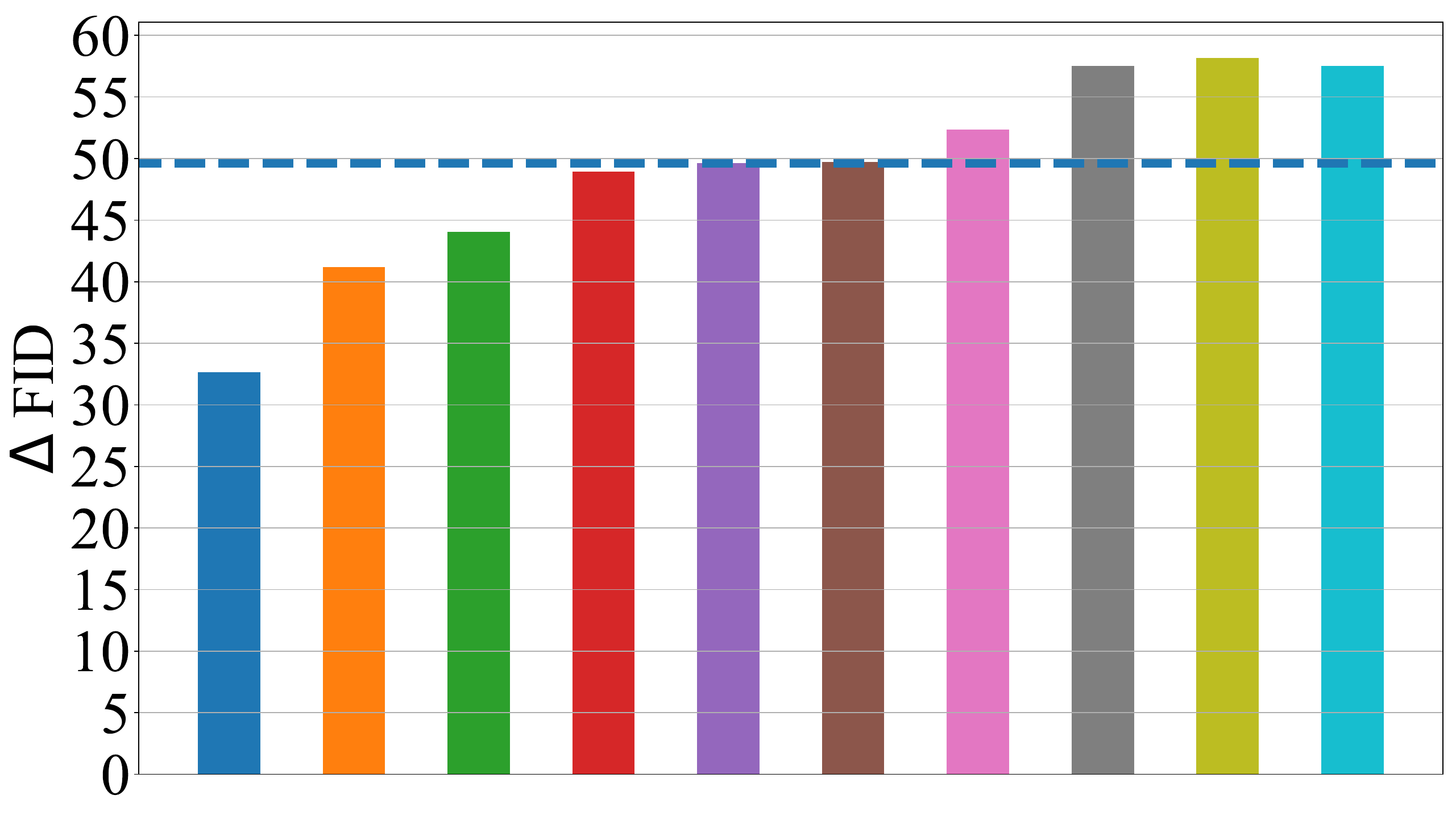}
		}
	\end{minipage}\\
	\begin{minipage}{0.9\linewidth}
			\centering
			\subfloat[SD-v1-5 to SD-v1-4]{\label{fig:difdIV5to4}
					\includegraphics[width=0.24\linewidth]{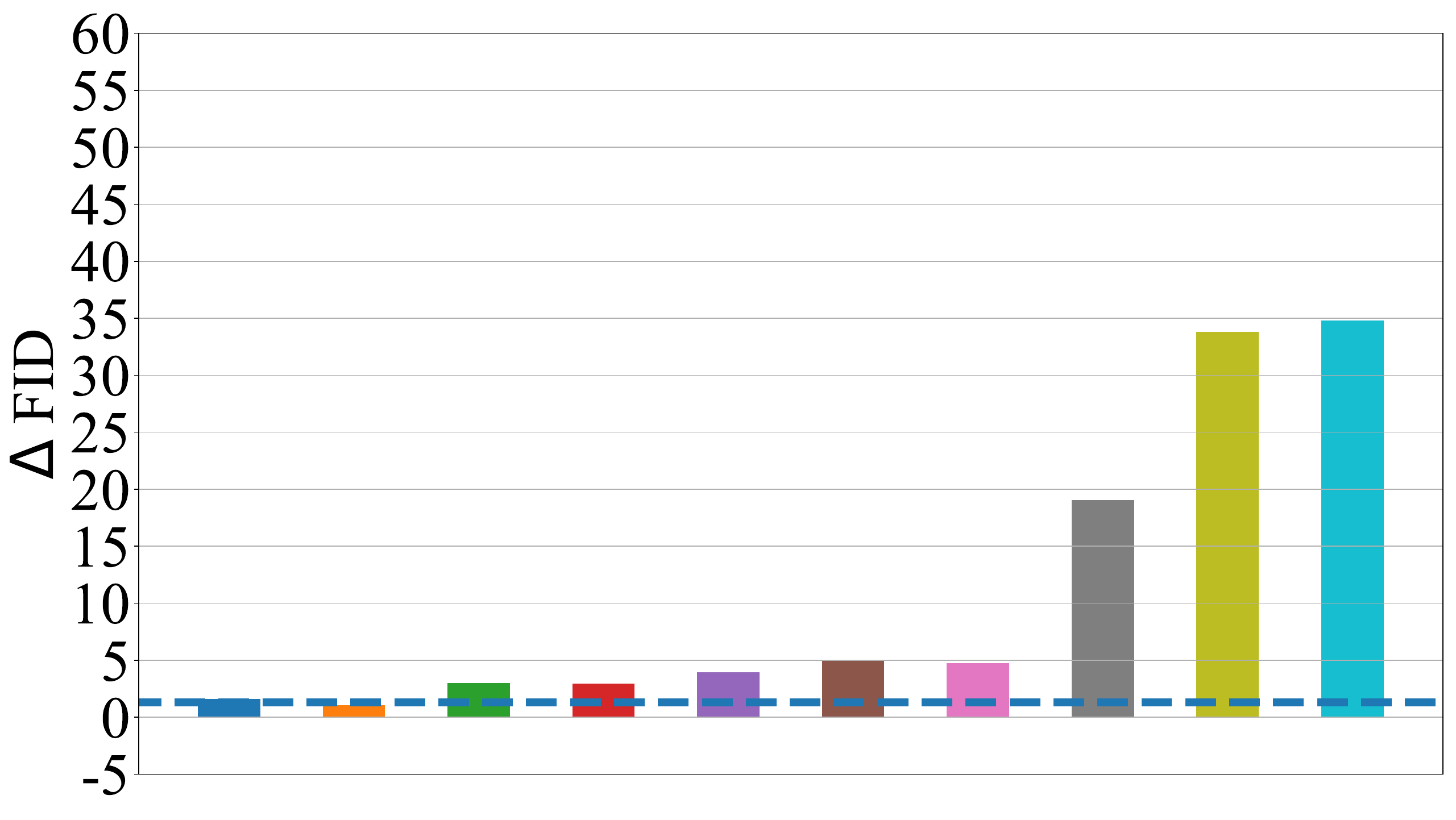}
				}
			\subfloat[SD-v1-5 to SD-v1-5]{\label{fig:difdIV5to5}
					\includegraphics[width=0.24\linewidth]{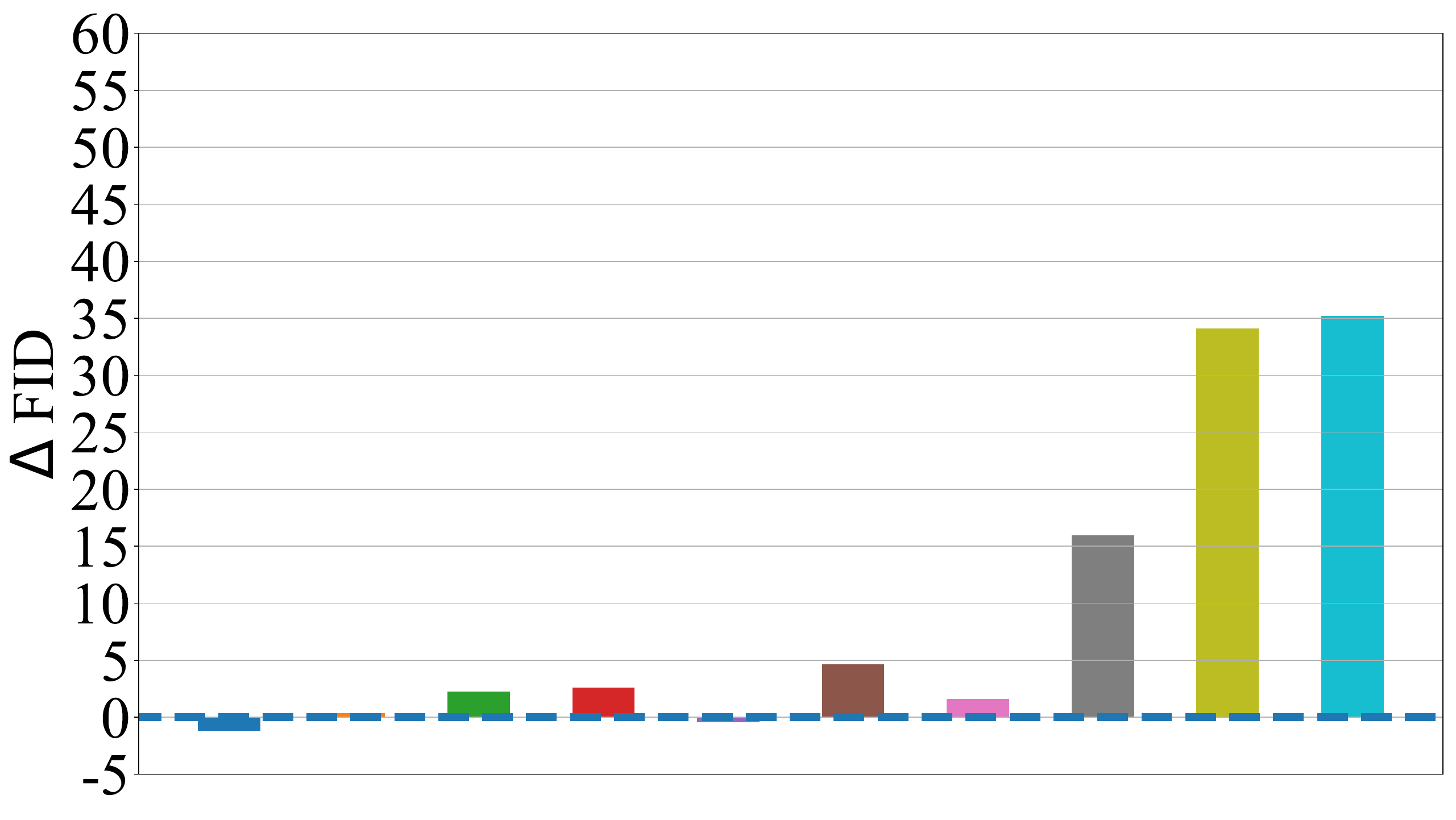}
				}
			\subfloat[SD-v1-5 to SD-v2-1]{\label{fig:difdIV5to2}
					\includegraphics[width=0.24\linewidth]{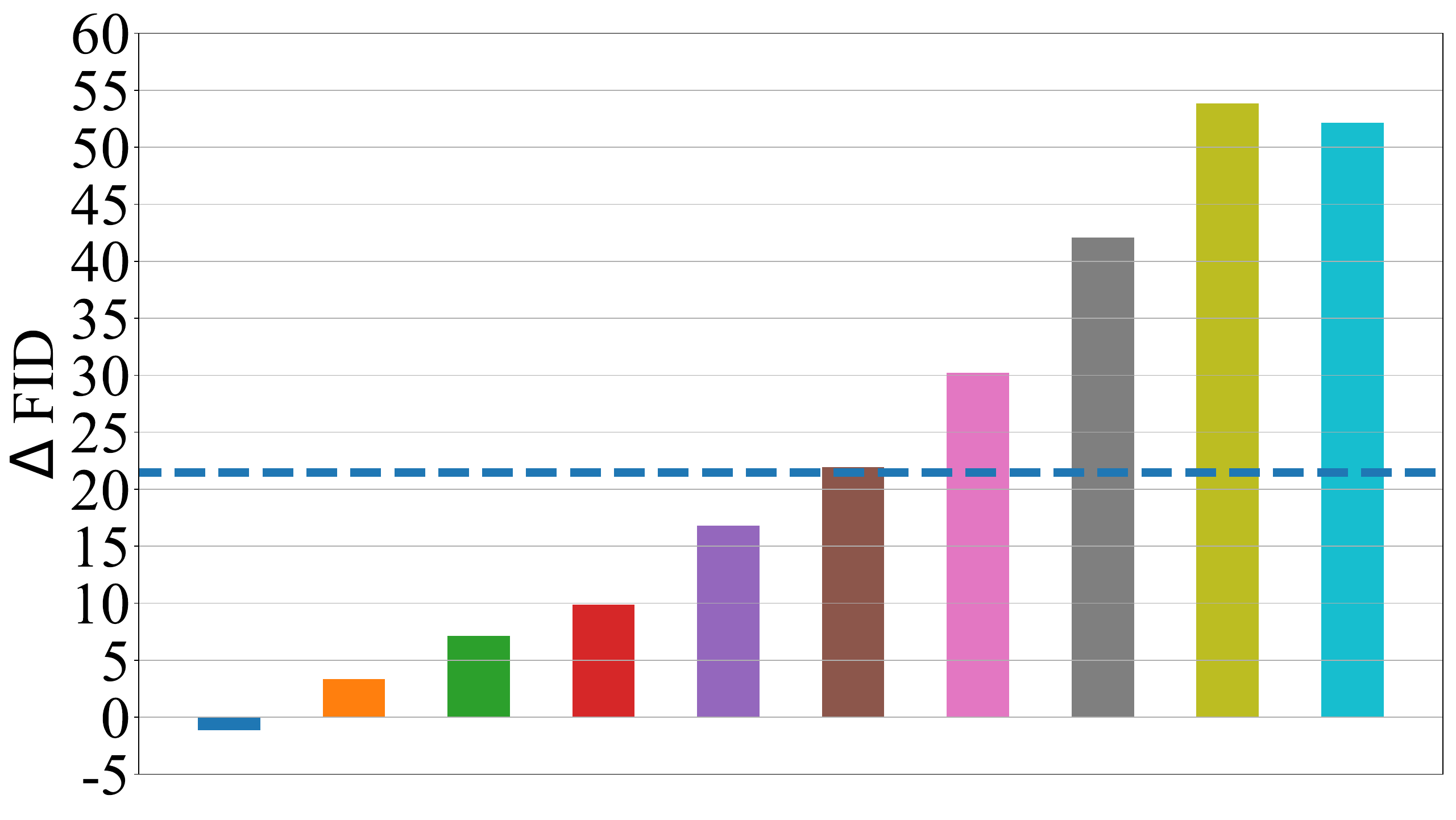}
				}
			\subfloat[SD-v1-5 to Instruct]{
				\includegraphics[width=0.24\linewidth]{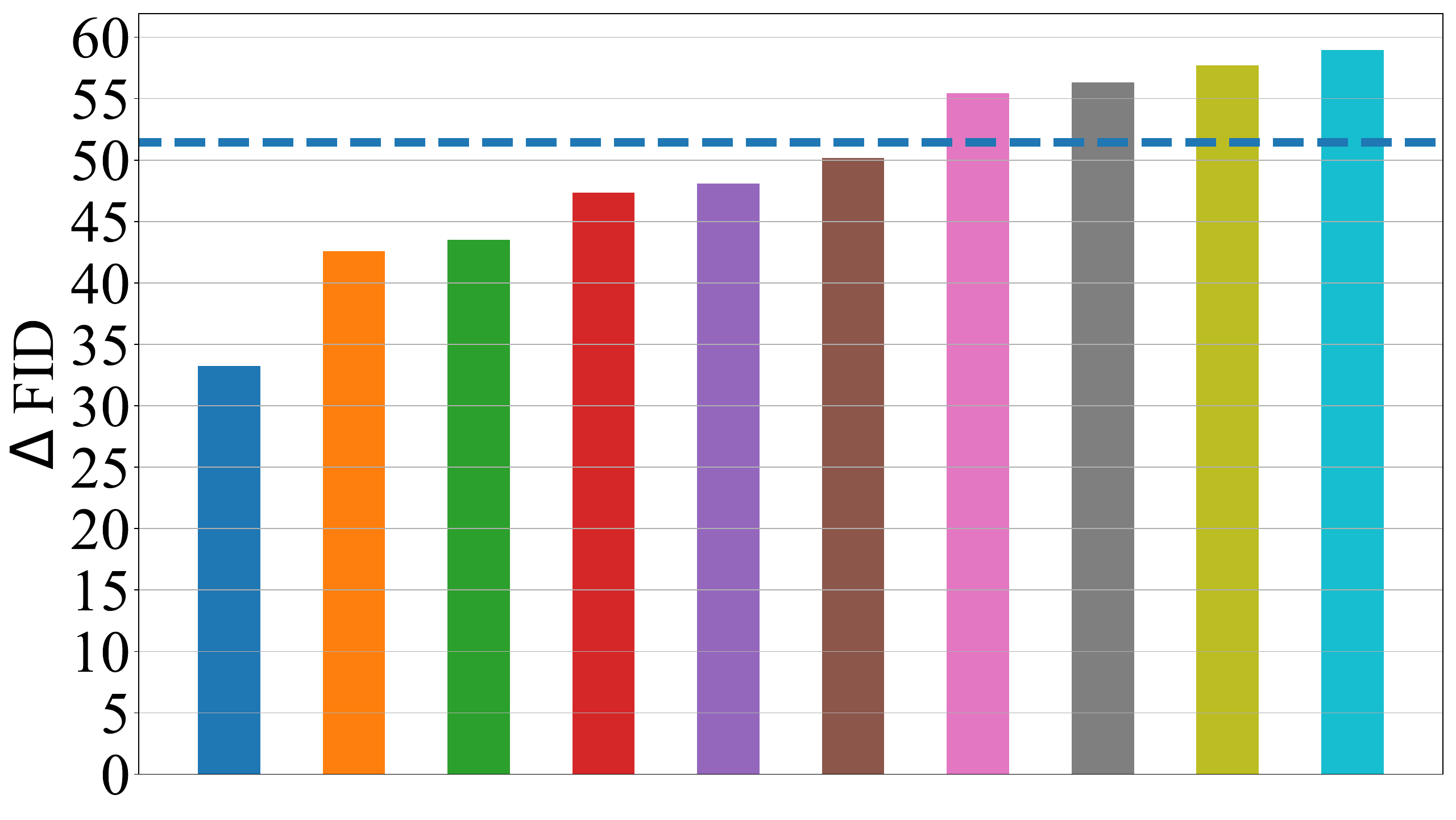}
			}
		\end{minipage}\\
	\begin{minipage}{0.9\linewidth}
		\centering
		\subfloat[SD-v2-1 to SD-v1-4]{
			\includegraphics[width=0.24\linewidth]{./figure/dfid2to4.pdf}
		}
		\subfloat[SD-v2-1 to SD-v1-5]{\label{fig:difdIV2to5}
				\includegraphics[width=0.24\linewidth]{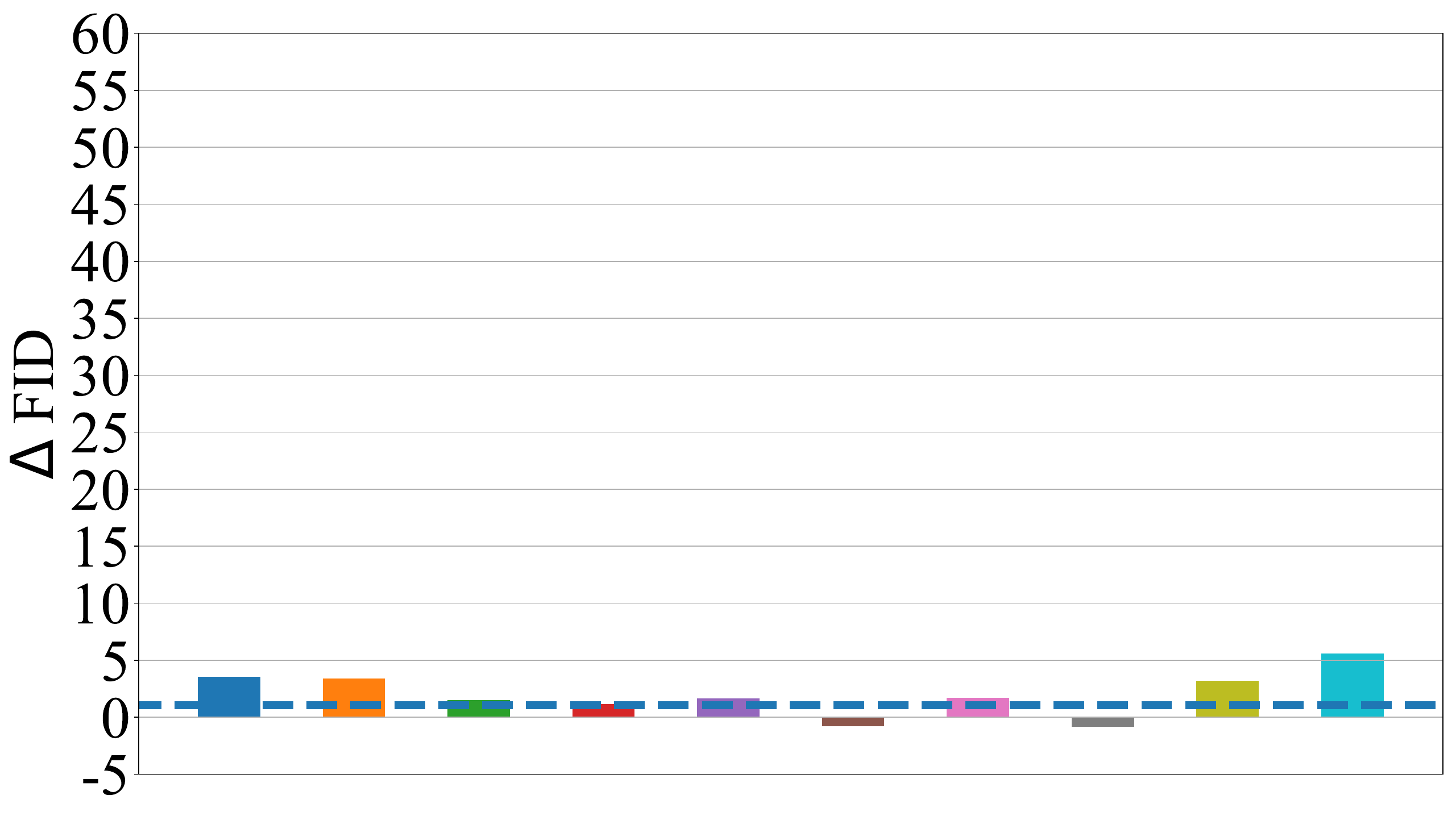}
			}
		\subfloat[SD-v2-1 to SD-v2-1]{
			\includegraphics[width=0.24\linewidth]{./figure/dfid2to2.pdf}
		}
		\subfloat[SD-v2-1 to Instruct]{
			\includegraphics[width=0.24\linewidth]{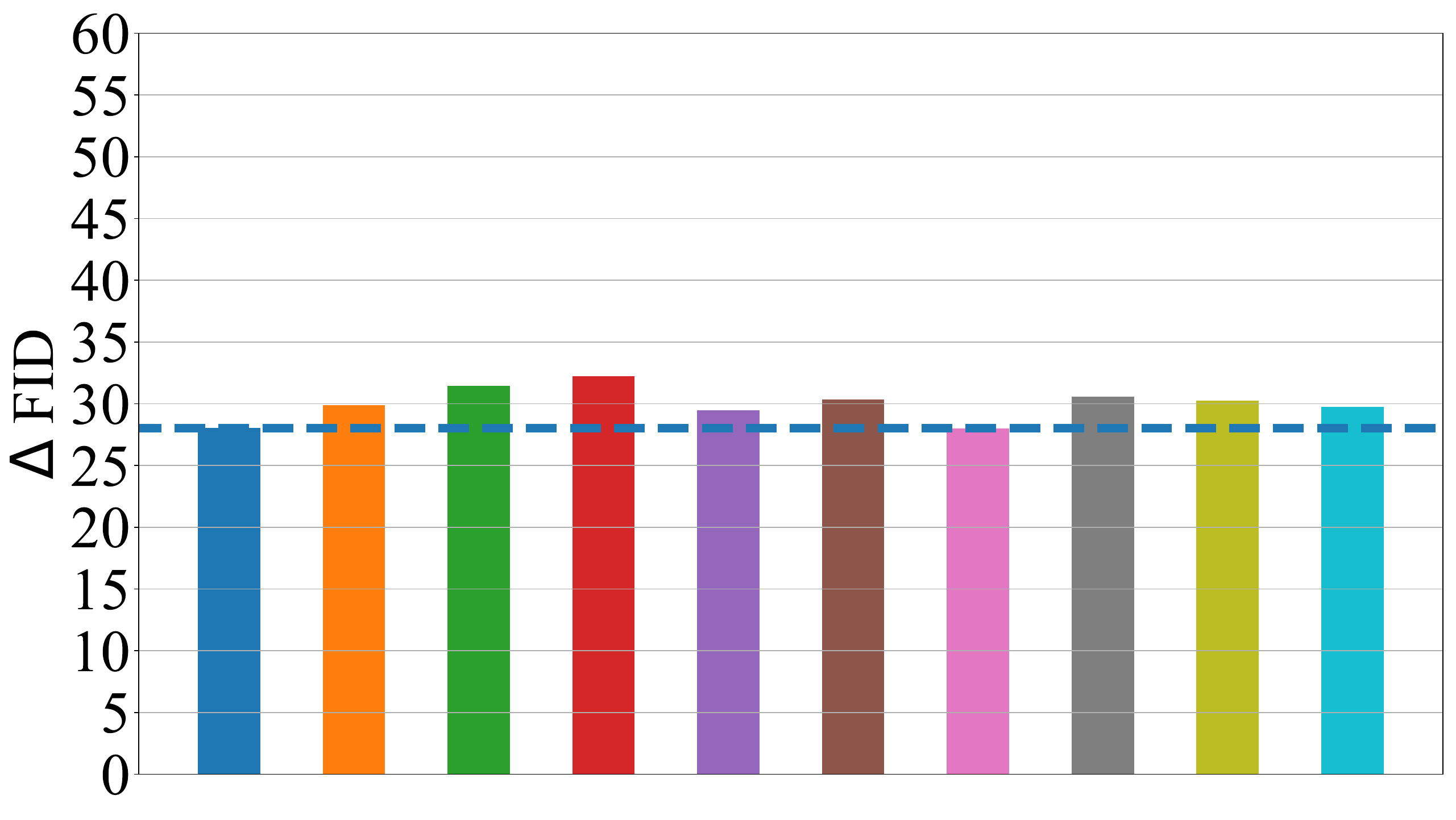}
		}
	\end{minipage}
	\caption{The increments $\uparrow$ of FID relative to the benign's. Larger increment means worse generated images and better AEs for image variation task. ``A to B'' means that AEs are generated by A and attack B.}\label{fig:dfidIVMore}
\end{figure}

\begin{figure}[!h]
	\centering
	\includegraphics[width=0.9\linewidth]{./figure/loss_dfidLegend.pdf}\\
	\begin{minipage}{0.9\linewidth}
		\centering
		\subfloat[{\scriptsize SD-v1-4 to SD-Inpainting}]{
			\includegraphics[width=0.33\linewidth]{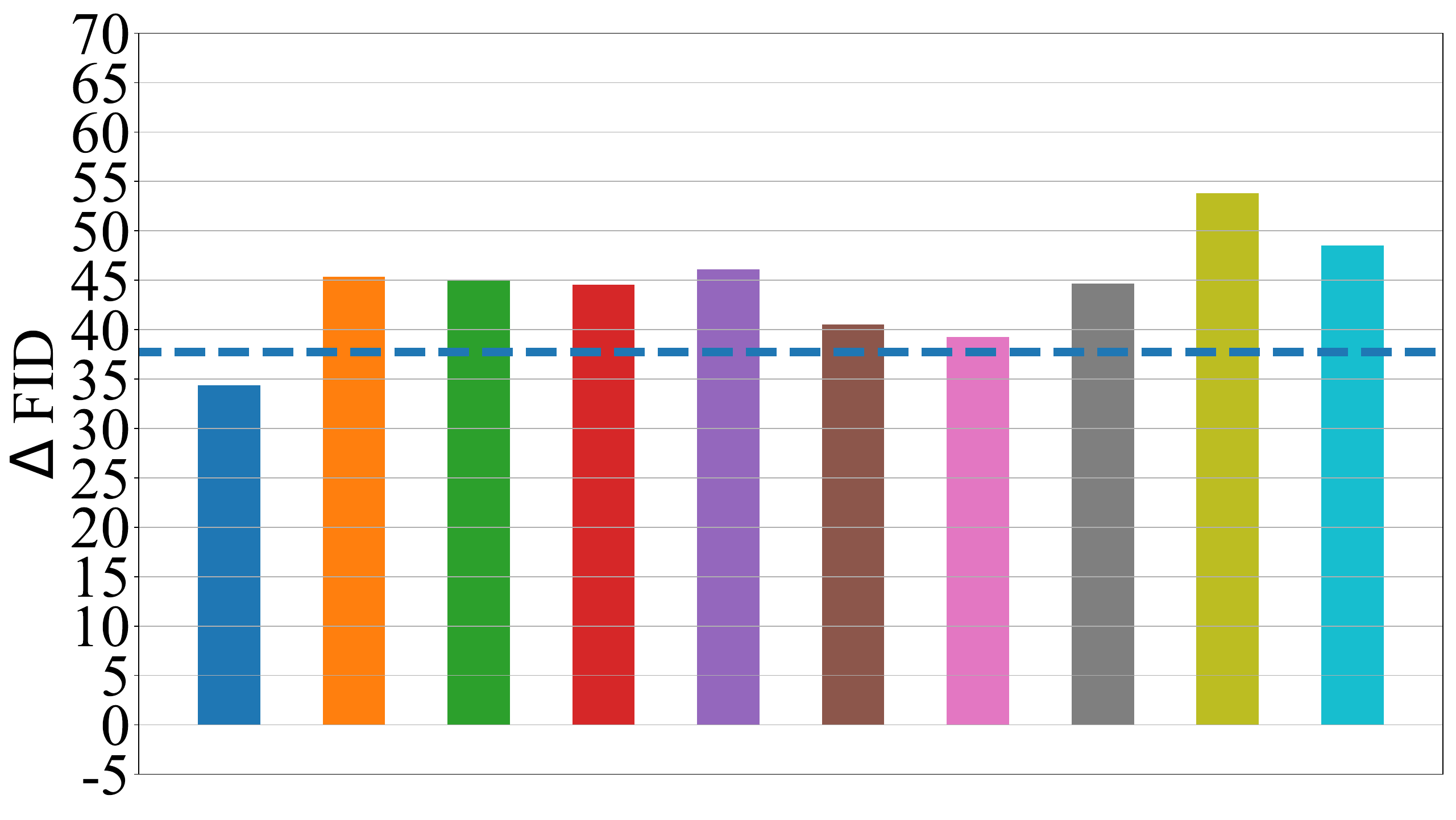}
		}
		\subfloat[{\scriptsize SD-v1-5 to SD-Inpainting}]{
				\includegraphics[width=0.33\linewidth]{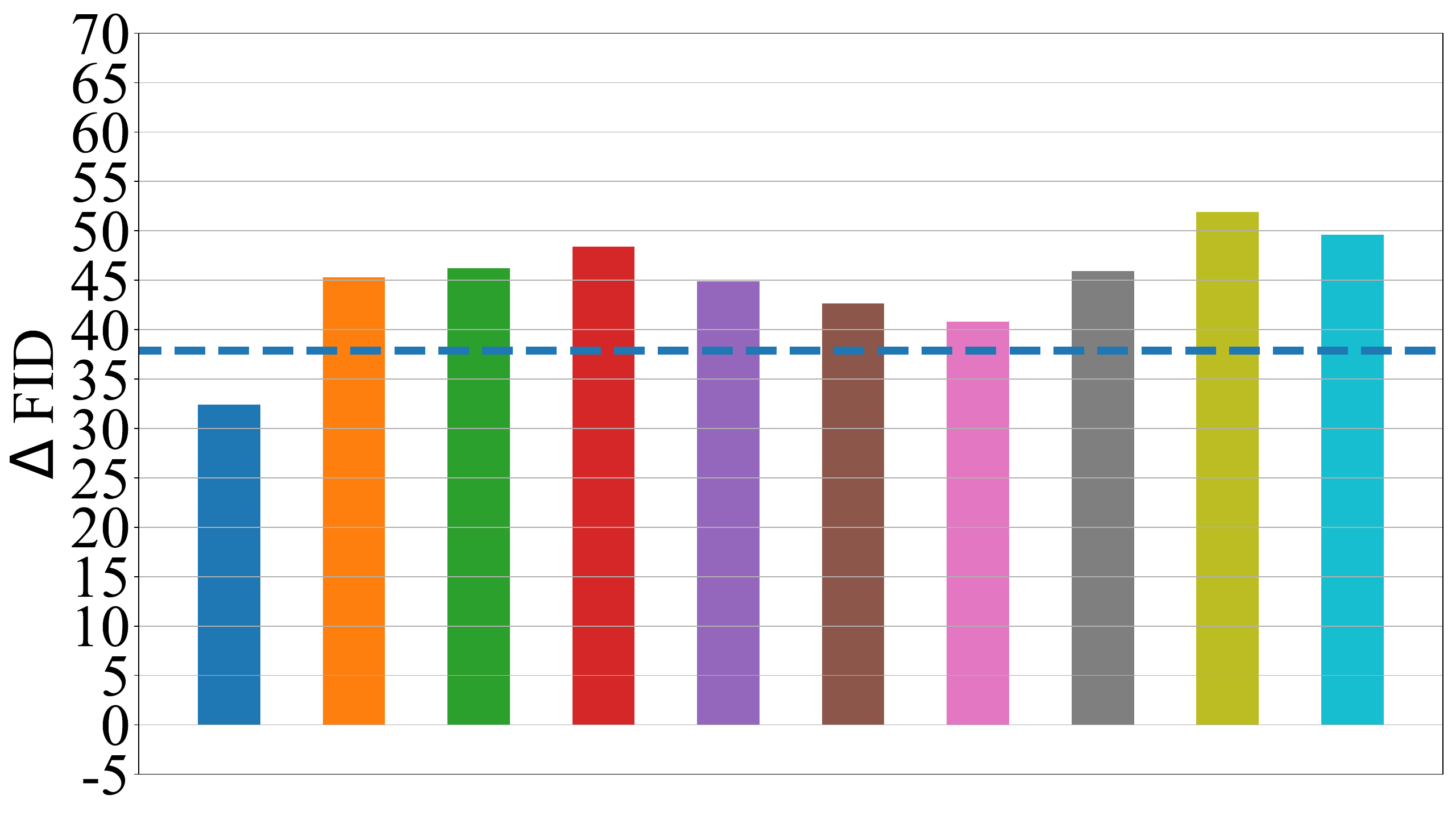}
				
		}
		\subfloat[{\scriptsize SD-v2-1 to SD-Inpainting}]{\label{subfig:2toi1}
		\includegraphics[width=0.33\linewidth]{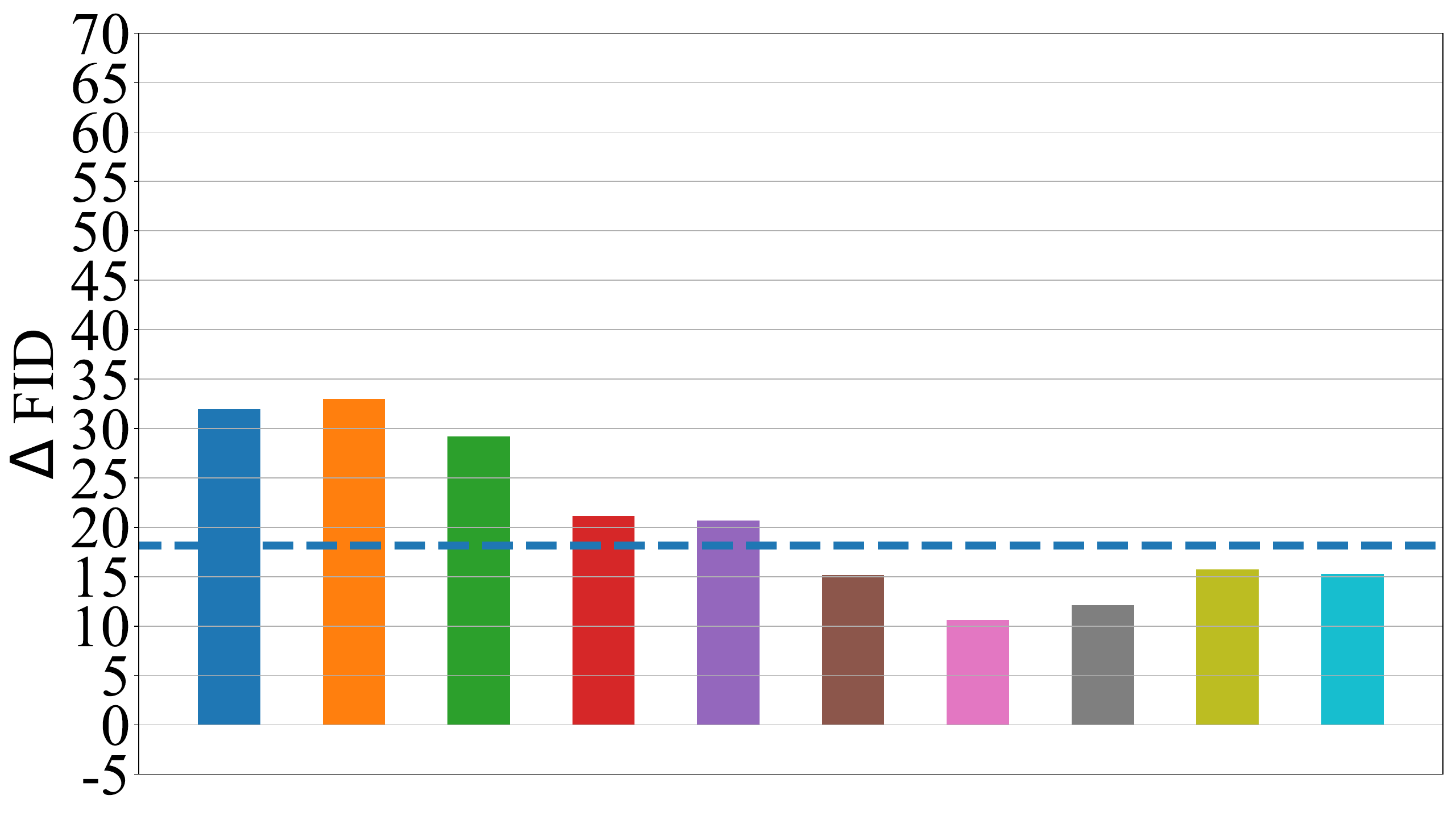}
		}
	\end{minipage}\\
	\begin{minipage}{0.9\linewidth}
		\centering
		\subfloat[{\scriptsize SD-v1-4 to SD-2-Inpainting}]{
			\includegraphics[width=0.33\linewidth]{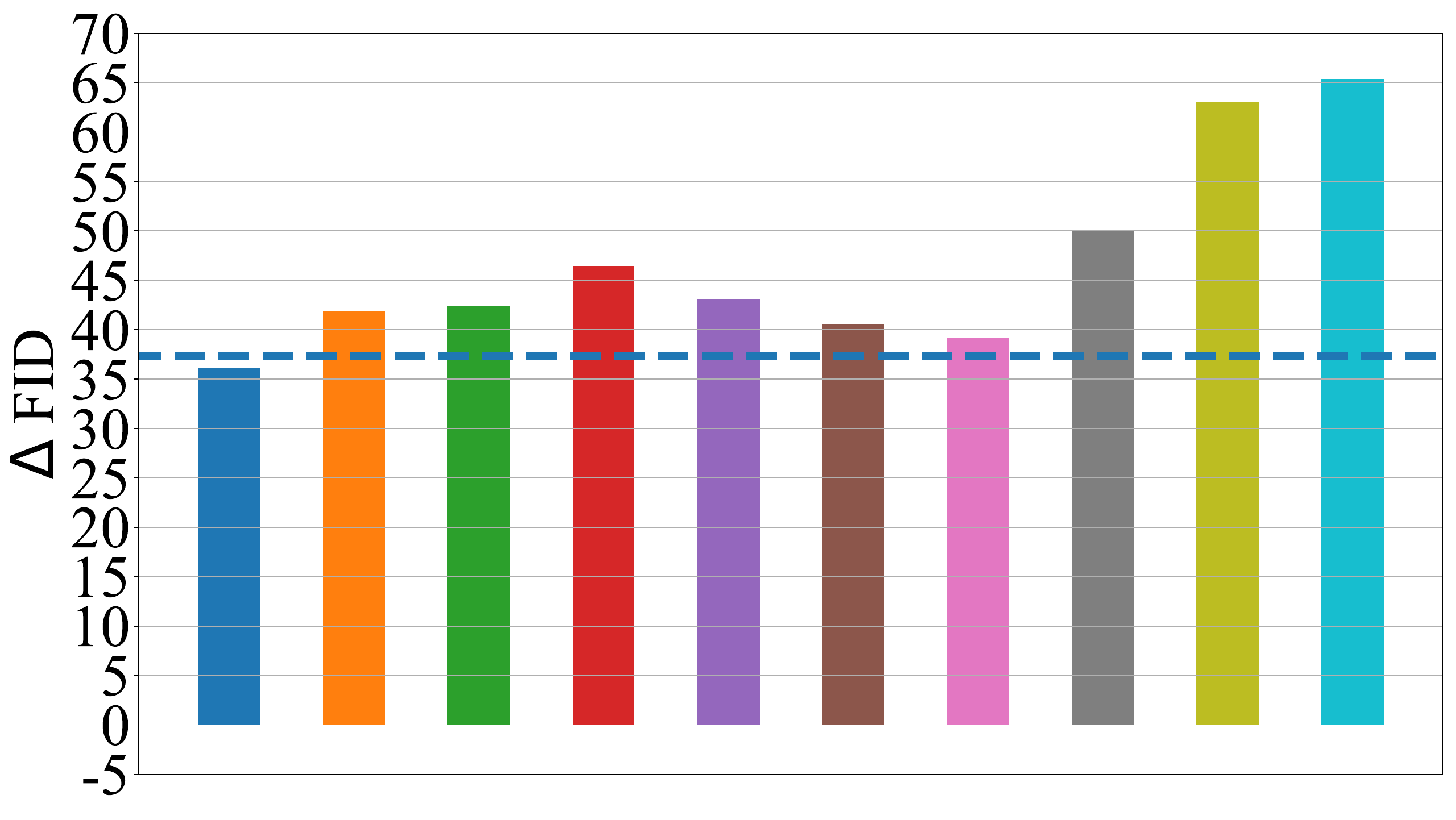}
		}
		\subfloat[{\scriptsize SD-v1-5 to SD-2-Inpainting}]{
				\includegraphics[width=0.33\linewidth]{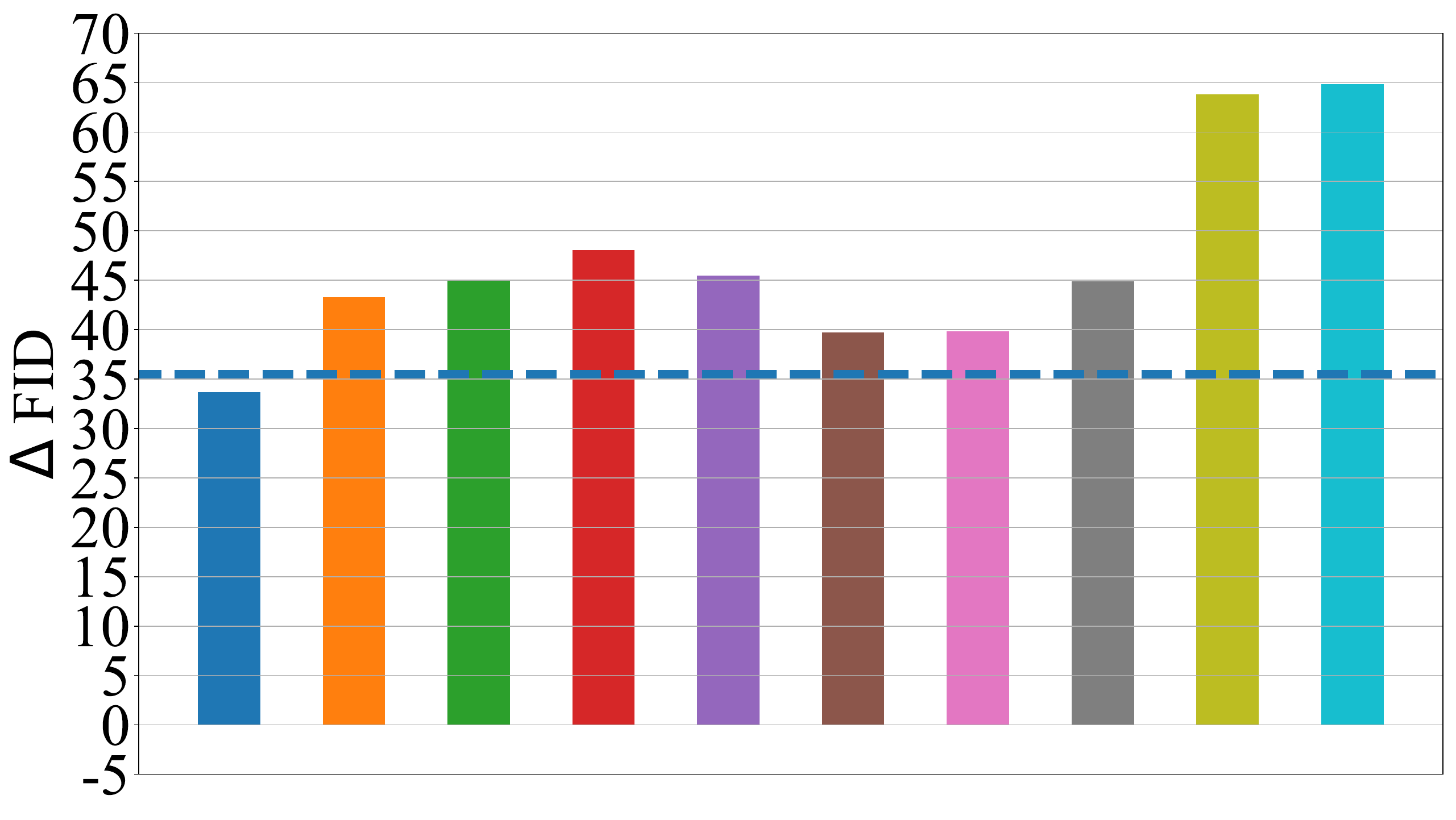}
			}
		\subfloat[{\scriptsize SD-v2-1 to SD-2-Inpainting}]{\label{subfig:2toi2}
			\includegraphics[width=0.33\linewidth]{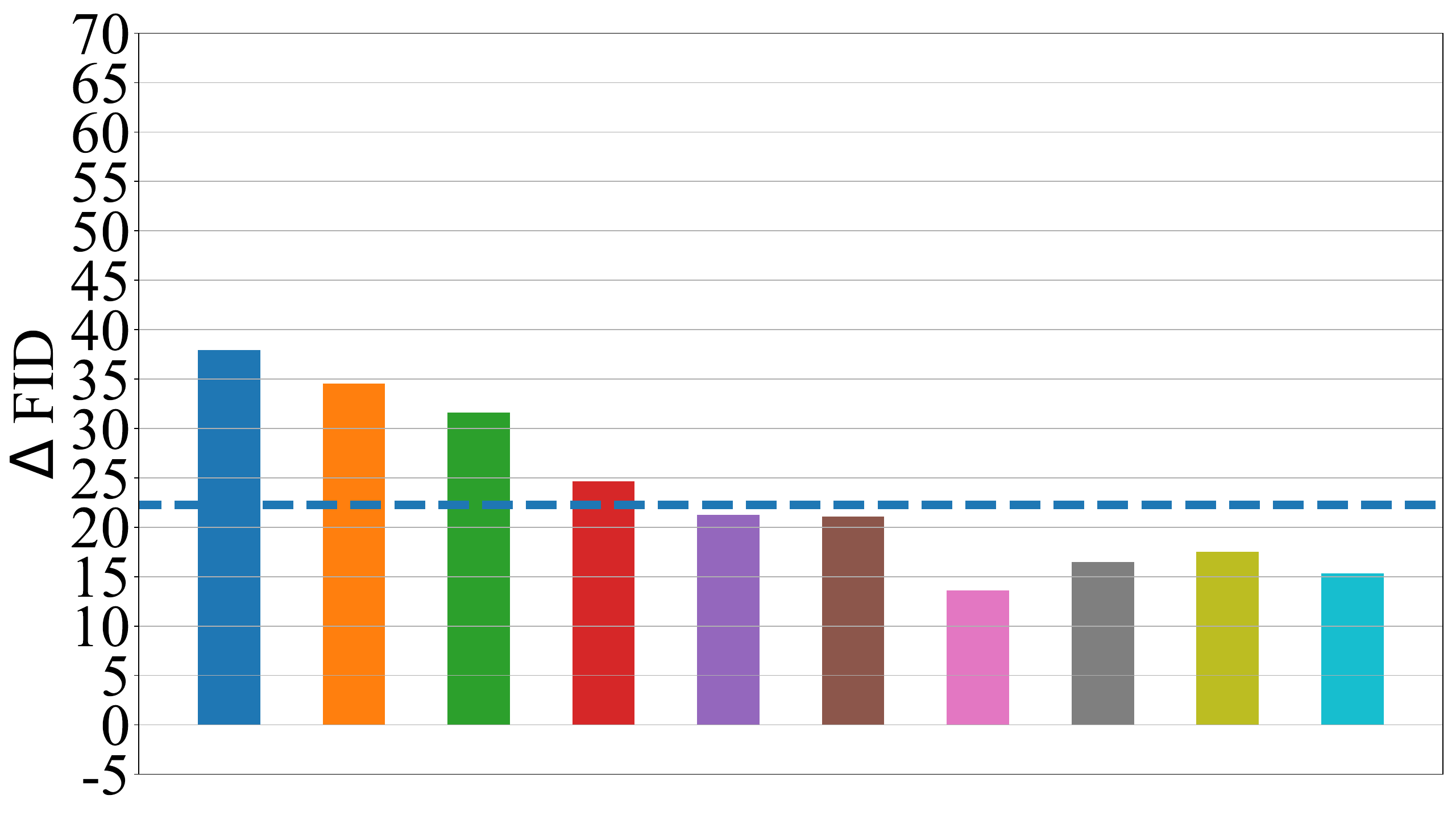}
		}
	\end{minipage}
	\caption{The increments $\uparrow$ of FID relative to the benign's. Larger increment means worse generated images and better AEs for image inpainting task. ``A to B'' means that AEs are generated by A and attack B.}\label{fig:dfidInpaintMore}
\end{figure}
\newpage
\newpage
\section{Results with SD-v1-5 being the Surrogate Model} \label{sec:v15MainReuslt}

\begin{table*}[!h]
	\caption{The performance of AEs in corrupting image variation. The best result in each column is in \textbf{bold}.} \label{tab:v15VarResult}
	\Huge
	\resizebox{\linewidth}{!}{
		\begin{tabular}{@{}cccccccccccccccccccccc@{}}
			\toprule
			\multirow{2}{*}{Surrogate Model} & \multirow{2}{*}{Method}                    & \multicolumn{4}{c}{FID$\uparrow$}                                               & \multicolumn{4}{c}{IS$\downarrow$}                                           & \multicolumn{4}{c}{CLIP$\downarrow$}                                          & \multicolumn{4}{c}{CLIP-IQA (Natural)$\downarrow$}                        & \multicolumn{4}{c}{CLIP-IQA (Noisiness)$\downarrow$}                      \\ \cmidrule(l){3-22} 
			&                                            & SD-v1-4         & SD-v1-5         & SD-v2-1         & Instruct        & SD-v1-4        & SD-v1-5        & SD-v2-1       & Instruct       & SD-v1-4        & SD-v1-5        & SD-v2-1        & Instruct       & SD-v1-4       & SD-v1-5       & SD-v2-1       & Instruct      & SD-v1-4       & SD-v1-5       & SD-v2-1       & Instruct      \\ \midrule
			\multirow{9}{*}{SD-v1-5}         & AdvDM                                      & 165.60          & 166.50          & 217.27          & 181.85          & 15.85          & 15.64          & 9.74          & 15.34          & 34.67          & 34.57          & 30.96          & 34.29          & 0.45          & 0.45          & 0.19          & 0.22          & 0.44          & 0.44          & 0.28          & 0.39          \\
			& SDS                                        & 175.04          & 176.68          & 210.96          & 158.57          & 16.91          & 16.17          & 11.24         & 14.94          & 34.42          & 34.31          & 31.83          & 34.15          & 0.43          & 0.42          & 0.23          & 0.23          & 0.46          & 0.46          & 0.33          & 0.31          \\
			& Chain                                      & 174.58          & 172.95          & 211.11          & 157.83          & 15.02          & 14.41          & 9.88          & 16.62          & 34.28          & 34.37          & 31.43          & 34.55          & 0.39          & 0.39          & 0.18          & 0.23          & 0.44          & 0.45          & 0.31          & 0.32          \\
			& Mist (0)                                   & 181.24          & 181.81          & 231.18          & 158.81          & 16.62          & 15.04          & 9.01          & 15.23          & 34.12          & 34.09          & 30.54          & 34.01          & 0.44          & 0.44          & 0.21          & 0.18          & 0.45          & 0.46          & 0.32          & \textbf{0.17} \\
			& Mist (1)                                   & 180.79          & 183.68          & 235.27          & 157.01          & 15.93          & 15.00          & 9.36          & \textbf{14.71}          & 34.12          & 34.08          & 30.44          & 33.90          & 0.44          & 0.44          & 0.21          & \textbf{0.17} & 0.46          & 0.46          & 0.33          & \textbf{0.17} \\
			& Mist (5)                                   & 167.58          & 167.50          & 216.56          & 181.35          & 16.58          & 15.04          & 10.02         & 14.94          & 34.60          & 34.56          & 31.12          & 34.20          & 0.45          & 0.45          & 0.19          & 0.22          & 0.44          & 0.46          & 0.28          & 0.39          \\ \cmidrule(l){2-22} 
			& AdvDM$_{800}^{900}$    & 198.10          & 200.62          & \textbf{249.66}          & 188.11          & 11.79          & 12.69          & 7.75          & 15.56          & 33.70          & 33.73          & 29.75          & 33.95          & 0.30          & 0.32          & \textbf{0.09} & 0.21          & \textbf{0.32}          & \textbf{0.33}          & \textbf{0.16} & 0.41          \\
			& Mist$_{800}^{900}$ (1) & 181.74          & 180.81          & 234.47          & 157.97          & 16.05          & 15.26          & 8.60          & 15.41          & 34.13          & 34.03          & 30.41          & 33.97          & 0.44          & 0.44          & 0.19          & 0.18          & 0.45          & 0.47          & 0.31          & \textbf{0.17} \\
			& Mist$_{800}^{900}$ (5) & \textbf{198.51}          & \textbf{202.07} & 249.01          & \textbf{189.57} & \textbf{11.49}          & \textbf{11.96}          & \textbf{7.38}          & 14.90          & \textbf{33.67}          & \textbf{33.71}          & \textbf{29.52} & \textbf{33.88}          & \textbf{0.29} & \textbf{0.32}          & \textbf{0.09} & 0.21          & 0.32          & \textbf{0.33}          & \textbf{0.16} & 0.41          \\ \bottomrule
	\end{tabular}}
\end{table*}

\begin{table*}[!h]
	\begin{minipage}{\linewidth}
		\caption{The performance of AEs in corrupting image inpainting. The best result in each column is in \textbf{bold}.} \label{tab:v15InpaintResult}
		{\Huge
			\resizebox{\linewidth}{!}{
				\begin{tabular}{@{}cccccccccccc@{}}
					\toprule
					\multirow{2}{*}{Surrogate Model} & \multirow{2}{*}{Method}                    & \multicolumn{2}{c}{FID$\uparrow$}           & \multicolumn{2}{c}{IS$\downarrow$}           & \multicolumn{2}{c}{CLIP$\downarrow$}         & \multicolumn{2}{c}{CLIP-IQA (Natural)$\downarrow$} & \multicolumn{2}{c}{CLIP-IQA (Noisiness)$\downarrow$} \\ \cmidrule(l){3-12} 
					&                                            & SD-Inpainting   & SD-2-Inpainting & SD-Inpainting  & SD-2-Inpainting & SD-Inpainting  & SD-2-Inpainting & SD-Inpainting     & SD-2-Inpainting    & SD-Inpainting      & SD-2-Inpainting     \\ \midrule
					\multirow{8}{*}{SD-v1-5}         & AdvDM                                      & 145.94          & 141.29          & 14.59          & 14.36           & 34.32          & 34.97           & 0.26              & 0.28               & 0.36               & 0.33                \\
					& SDS                                        & 145.06          & 148.10          & 15.14          & 15.27           & 34.59          & 34.86           & 0.35              & 0.34               & 0.36               & 0.27                \\
					& Mist (0)                                   & 155.18          & 163.12          & 13.90          & 14.73           & \textbf{33.99}          & 34.16           & 0.25              & \textbf{0.21}      & 0.37               & 0.30                \\
					& Mist (1)                                   & 155.24          & 161.80          & 14.96          & 14.44           & 34.04          & 34.31           & 0.25              & 0.22               & 0.37               & 0.29                \\
					& Mist (5)                                   & 146.10          & 144.39          & 14.52          & 14.93           & 34.28          & 34.99           & 0.26              & 0.27               & 0.37               & 0.32                \\ \cmidrule(l){2-12} 
					& AdvDM$_{800}^{900}$    & \textbf{159.95}          & \textbf{169.59} & \textbf{13.57}          & 12.78           & 34.34          & 34.57           & \textbf{0.21}     & \textbf{0.21}      & \textbf{0.27}               & \textbf{0.23}       \\
					& Mist$_{800}^{900}$ (1) & 156.85          & 161.62          & 14.61          & 14.70           & 34.13          & \textbf{34.14}  & 0.25              & 0.22               & 0.37               & 0.30                \\
					& Mist$_{800}^{900}$ (5) & 158.79          & 168.68          & 13.69          & \textbf{12.69}           & 34.29          & 34.65           & \textbf{0.21}     & \textbf{0.21}      & \textbf{0.27}               & \textbf{0.23}        \\ \bottomrule
		\end{tabular}}}
	\end{minipage}
\end{table*}

\section{Results of adversarial attacks with 500 iteration steps} \label{sec:500Steps}

We present the results of adversarial attacks with 500 iteration steps in \cref{tab:iv500Steps} and \cref{tab:ii500Steps}. Due to the high computation overhead of Chain \cite{benchmark}, it is impossible to conduct Chain with 500 iteration steps.

\begin{table}[!h]
	\caption{The performance of AEs generated by different methods with 500 iteration steps in cracking image variation. The surrogate model is SD-v1-4.} \label{tab:iv500Steps}
	\resizebox{\linewidth}{!}{{\Huge
			\begin{tabular}{@{}ccccccccccccccccccccc@{}}
				\toprule
				\multirow{2}{*}{Method}                 & \multicolumn{4}{c}{FID$\uparrow$}                                               & \multicolumn{4}{c}{IS$\downarrow$}                                         & \multicolumn{4}{c}{CLIP$\downarrow$}                                          & \multicolumn{4}{c}{CLIP-IQA (Natural)$\downarrow$}                        & \multicolumn{4}{c}{CLIP-IQA (Noisiness)$\downarrow$}                      \\ \cmidrule(l){2-21} 
				& SD-v1-4         & SD-v1-5         & SD-v2-1         & Instruct        & SD-v1-4       & SD-v1-5       & SD-v2-1       & Instruct       & SD-v1-4        & SD-v1-5        & SD-v2-1        & Instruct       & SD-v1-4       & SD-v1-5       & SD-v2-1       & Instruct      & SD-v1-4       & SD-v1-5       & SD-v2-1       & Instruct      \\ \midrule
				AdvDM                                   & 173.77          & 171.11          & 230.73          & 184.93          & 15.30         & 15.17         & 9.22          & 15.29          & 34.47          & 34.54          & 30.52          & 33.95          & 0.44          & 0.43          & 0.18          & 0.22          & 0.44          & 0.45          & 0.27          & 0.39          \\
				SDS                                     & 181.94          & 179.55          & 214.01          & 178.81          & 16.85         & 15.61         & 10.81         & 15.48          & 34.17          & 34.15          & 31.54          & 33.65          & 0.42          & 0.43          & 0.22          & 0.18          & 0.46          & 0.47          & 0.35          & 0.32          \\
				Mist (0)                                & 184.27          & 183.17          & 245.44          & 181.69          & 15.88         & 14.83         & 8.37          & 14.27          & 34.07          & 33.87          & 29.74          & 33.32          & 0.44          & 0.43          & 0.18          & 0.13          & 0.46          & 0.47          & 0.31          & 0.17          \\
				Mist (1)                                & 182.31          & 186.22          & 243.68          & 184.48          & 16.71         & 15.68         & 8.47          & \textbf{13.70} & 33.97          & 33.91          & 29.74          & \textbf{33.27} & 0.43          & 0.44          & 0.18          & \textbf{0.12} & 0.47          & 0.48          & 0.30          & \textbf{0.15} \\
				Mist (5)                                & 171.63          & 172.66          & 229.18          & 186.08          & 15.82         & 15.74         & 9.76          & 15.72          & 34.41          & 34.29          & 30.53          & 34.00          & 0.43          & 0.43          & 0.16          & 0.22          & 0.44          & 0.45          & 0.26          & 0.40          \\ \midrule
				AdvDM$_{800}^{900}$ & \textbf{243.93} & \textbf{242.59} & \textbf{279.78} & \textbf{192.93} & \textbf{7.08} & \textbf{7.71} & \textbf{6.17} & 14.02          & \textbf{32.39} & \textbf{32.54} & \textbf{28.05} & 33.82          & \textbf{0.17} & \textbf{0.19} & \textbf{0.05} & 0.20          & \textbf{0.19} & \textbf{0.19} & \textbf{0.11} & 0.42          \\ \bottomrule
	\end{tabular}}}
\end{table}

\begin{table}[!h]
	\caption{The performance of AEs generated by different methods with 500 iteration steps in cracking image inpainting. The surrogate model is SD-v1-4.} \label{tab:ii500Steps}
	\resizebox{\linewidth}{!}{{\Huge
	\begin{tabular}{@{}ccccccccccc@{}}
		\toprule
		\multirow{2}{*}{Method}                 & \multicolumn{2}{c}{FID$\uparrow$}         & \multicolumn{2}{c}{IS$\downarrow$}          & \multicolumn{2}{c}{CLIP$\downarrow$}        & \multicolumn{2}{c}{CLIP-IQA (Natural)$\downarrow$} & \multicolumn{2}{c}{CLIP-IQA (Noisiness)$\downarrow$} \\ \cmidrule(l){2-11} 
		& SD-Inpainting & SD-2-Inpainting & SD-Inpainting & SD-2-Inpainting & SD-Inpainting & SD-2-Inpainting & SD-Inpainting     & SD-2-Inpainting    & SD-Inpainting      & SD-2-Inpainting     \\ \midrule
		AdvDM                                   & 155.47          & 152.80          & 14.42          & 14.67           & 34.24          & 34.92           & 0.23              & 0.25               & 0.35               & 0.32                \\
		SDS                                     & 154.22          & 156.96          & 15.89          & 15.26           & 34.43          & 34.61           & 0.31              & 0.30               & 0.31               & 0.23                \\
		Mist (0)                                & 172.52          & 176.01          & 13.18          & 13.84           & \textbf{33.63} & 33.86           & 0.22              & 0.19               & 0.36               & 0.29                \\
		Mist (1)                                & 171.89          & 178.58          & 13.79          & 14.29           & 33.73          & \textbf{33.85}  & 0.22              & 0.18               & 0.36               & 0.28                \\
		Mist (5)                                & 155.40          & 150.25          & 14.86          & 14.66           & 34.17          & 34.96           & 0.24              & 0.24               & 0.35               & 0.31                \\ \midrule
		AdvDM$_{800}^{900}$ & \textbf{188.28} & \textbf{213.68} & \textbf{10.92} & \textbf{8.51}   & 33.71          & 33.99           & \textbf{0.17}     & \textbf{0.14}      & \textbf{0.21}      & \textbf{0.15}       \\ \bottomrule
	\end{tabular}}}
\end{table}

\newpage
\section{Pre-processing Adversarial Defense} \label{sec:defense}
We use JPEG compression~\cite{jpgDefense} and TVM~\cite{tvm} to test the robustness of AEs generated by different methods (see \cref{tab:defense}). Generally, JPEG can degrade the performance of all methods to the same level since none have considered the robustness of AEs. One intriguing phenomenon in \cref{tab:defense} is that TVM, being an adversarial defense, even degrades the quality of the generated image, which suggests that disturbance brought by the pre-processing adversarial defense may even boost the performance of AEs for LDMs.
\begin{table}[!h]
	\caption{The performance of AEs facing different pre-processing adversarial defenses. The surrogate model is SD-v1-4. JPEG ($a$) represents JPEG compression with quality $a$.} \label{tab:defense}
	\resizebox{\linewidth}{!}{{ \Huge
	\begin{tabular}{@{}cccccccccccccccccccccc@{}}
		\toprule
		\multirow{2}{*}{Defense}  & \multirow{2}{*}{Method}                    & \multicolumn{4}{c}{FID$\uparrow$}                        & \multicolumn{4}{c}{IS$\downarrow$}                 & \multicolumn{4}{c}{CLIP$\downarrow$}               & \multicolumn{4}{c}{CLIP-IQA (Natural)$\downarrow$} & \multicolumn{4}{c}{CLIP-IQA (Noisiness)$\downarrow$} \\ \cmidrule(l){3-22} 
		&                                            & SD-v1-4         & SD-v1-5 & SD-v2-1 & Instruct & SD-v1-4 & SD-v1-5 & SD-v2-1 & Instruct & SD-v1-4 & SD-v1-5 & SD-v2-1 & Instruct & SD-v1-4 & SD-v1-5 & SD-v2-1 & Instruct & SD-v1-4  & SD-v1-5  & SD-v2-1 & Instruct \\ \midrule
		\multirow{9}{*}{JPEG (85)} & AdvDM                                      & 166.09          & 164.73  & 206.84  & 171.89   & 16.25   & 15.74   & 10.56   & 15.57    & 34.52   & 34.59   & 31.60   & 34.31    & 0.46    & 0.46    & 0.24    & 0.22     & 0.45     & 0.45     & 0.31    & 0.37     \\
		& SDS                                        & 170.51          & 170.02  & 203.24  & 133.20   & 16.45   & 15.92   & 12.12   & 16.55    & 34.57   & 34.61   & 32.34   & 34.55    & 0.43    & 0.44    & 0.26    & 0.30     & 0.46     & 0.46     & 0.36    & 0.29     \\
		& Chain                                      & 177.41          & 176.04  & 220.79  & 153.53   & 13.66   & 14.12   & 9.04    & 16.53    & 34.05   & 34.20   & 30.65   & 34.58    & 0.36    & 0.36    & 0.15    & 0.24     & 0.42     & 0.43     & 0.27    & 0.33     \\
		& Mist (0)                                   & 174.84          & 174.81  & 211.24  & 133.64   & 16.27   & 16.03   & 10.39   & 16.14    & 34.40   & 34.31   & 31.51   & 34.48    & 0.44    & 0.45    & 0.22    & 0.26     & 0.46     & 0.47     & 0.33    & 0.24     \\
		& Mist (1)                                   & 173.76          & 173.57  & 217.31  & 135.33   & 16.14   & 15.61   & 10.15   & 16.37    & 34.51   & 34.57   & 31.59   & 34.67    & 0.45    & 0.45    & 0.23    & 0.26     & 0.45     & 0.47     & 0.33    & 0.24     \\
		& Mist (5)                                   & 165.02          & 166.14  & 206.60  & 171.81   & 16.23   & 15.79   & 10.75   & 15.76    & 34.67   & 34.55   & 31.60   & 34.32    & 0.45    & 0.47    & 0.24    & 0.23     & 0.45     & 0.46     & 0.33    & 0.38     \\ \cmidrule(l){2-22} 
		& AdvDM$_{800}^{900}$    & 177.50          & 178.91  & 231.19  & 181.72   & 14.96   & 15.06   & 8.73    & 15.67    & 34.41   & 34.28   & 30.40   & 34.03    & 0.38    & 0.38    & 0.14    & 0.22     & 0.41     & 0.41     & 0.22    & 0.40     \\
		& Mist$_{800}^{900}$ (1) & 173.11          & 174.27  & 212.44  & 134.38   & 15.94   & 15.60   & 10.38   & 15.82    & 34.46   & 34.42   & 31.70   & 34.51    & 0.45    & 0.46    & 0.24    & 0.25     & 0.45     & 0.47     & 0.34    & 0.24     \\
		& Mist$_{800}^{900}$ (5) & 178.65 & 179.96  & 228.83  & 180.12   & 14.65   & 15.12   & 9.03    & 15.20    & 34.48   & 34.24   & 30.32   & 34.04    & 0.38    & 0.39    & 0.13    & 0.23     & 0.41     & 0.41     & 0.21    & 0.40     \\ \midrule
		\multirow{9}{*}{JPEG (75)} & AdvDM                                      & 165.40          & 165.09  & 206.95  & 166.66   & 16.45   & 15.46   & 11.31   & 16.75    & 34.56   & 34.52   & 31.78   & 34.38    & 0.46    & 0.47    & 0.27    & 0.23     & 0.45     & 0.46     & 0.33    & 0.35     \\
		& SDS                                        & 167.17          & 168.36  & 200.83  & 129.91   & 16.32   & 16.40   & 12.21   & 16.77    & 34.51   & 34.62   & 32.24   & 34.62    & 0.45    & 0.45    & 0.25    & 0.31     & 0.46     & 0.48     & 0.36    & 0.30     \\
		& Chain                                      & 171.84          & 171.06  & 215.63  & 148.18   & 14.50   & 14.73   & 9.65    & 16.49    & 34.30   & 34.41   & 30.96   & 34.71    & 0.39    & 0.40    & 0.18    & 0.26     & 0.45     & 0.45     & 0.30    & 0.32     \\
		& Mist (0)                                   & 172.01          & 172.69  & 206.14  & 131.94   & 16.68   & 15.73   & 11.23   & 16.74    & 34.33   & 34.50   & 31.99   & 34.69    & 0.45    & 0.45    & 0.26    & 0.28     & 0.47     & 0.47     & 0.35    & 0.27     \\
		& Mist (1)                                   & 170.70          & 169.86  & 206.38  & 135.30   & 16.50   & 15.43   & 11.08   & 16.99    & 34.32   & 34.50   & 31.97   & 34.67    & 0.46    & 0.45    & 0.25    & 0.27     & 0.46     & 0.47     & 0.34    & 0.27     \\
		& Mist (5)                                   & 166.51          & 166.65  & 201.29  & 165.70   & 16.53   & 15.79   & 11.31   & 16.19    & 34.62   & 34.62   & 31.76   & 34.42    & 0.46    & 0.47    & 0.25    & 0.23     & 0.45     & 0.46     & 0.33    & 0.36     \\ \cmidrule(l){2-22} 
		& AdvDM$_{800}^{900}$    & 171.64          & 171.97  & 216.75  & 171.96   & 14.99   & 15.60   & 9.56    & 15.50    & 34.57   & 34.48   & 31.08   & 34.22    & 0.41    & 0.41    & 0.17    & 0.23     & 0.43     & 0.44     & 0.27    & 0.39     \\
		& Mist$_{800}^{900}$ (1) & 172.39          & 171.03  & 208.86  & 133.40   & 16.54   & 15.13   & 10.51   & 16.66    & 34.40   & 34.52   & 32.00   & 34.64    & 0.46    & 0.46    & 0.25    & 0.28     & 0.47     & 0.47     & 0.33    & 0.27     \\
		& Mist$_{800}^{900}$ (5) & 171.68          & 171.91  & 217.01  & 174.63   & 15.34   & 15.43   & 9.93    & 15.68    & 34.56   & 34.57   & 31.12   & 34.29    & 0.41    & 0.42    & 0.16    & 0.23     & 0.44     & 0.44     & 0.27    & 0.39     \\ \midrule
		\multirow{9}{*}{JPEG (65)} & AdvDM                                      & 166.78          & 165.42  & 204.28  & 165.22   & 16.37   & 15.98   & 11.44   & 16.78    & 34.51   & 34.60   & 31.79   & 34.45    & 0.46    & 0.47    & 0.26    & 0.23     & 0.45     & 0.46     & 0.34    & 0.34     \\
		& SDS                                        & 166.18          & 168.42  & 203.62  & 133.36   & 16.22   & 15.79   & 11.68   & 16.90    & 34.50   & 34.59   & 32.02   & 34.66    & 0.45    & 0.45    & 0.26    & 0.30     & 0.47     & 0.48     & 0.38    & 0.30     \\
		& Chain                                      & 169.69          & 170.03  & 211.33  & 147.95   & 15.70   & 15.53   & 9.84    & 16.36    & 34.37   & 34.48   & 31.28   & 34.76    & 0.41    & 0.42    & 0.19    & 0.26     & 0.46     & 0.48     & 0.32    & 0.33     \\
		& Mist (0)                                   & 170.60          & 171.17  & 204.06  & 135.86   & 17.47   & 16.11   & 11.43   & 16.67    & 34.46   & 34.50   & 32.00   & 34.74    & 0.45    & 0.46    & 0.26    & 0.29     & 0.47     & 0.49     & 0.36    & 0.29     \\
		& Mist (1)                                   & 170.09          & 169.28  & 206.35  & 136.44   & 16.61   & 15.87   & 11.51   & 16.63    & 34.40   & 34.52   & 31.98   & 34.70    & 0.45    & 0.45    & 0.25    & 0.28     & 0.48     & 0.49     & 0.35    & 0.28     \\
		& Mist (5)                                   & 165.74          & 164.92  & 199.34  & 165.08   & 16.26   & 15.15   & 11.28   & 16.24    & 34.56   & 34.52   & 31.97   & 34.51    & 0.46    & 0.47    & 0.25    & 0.23     & 0.46     & 0.47     & 0.34    & 0.35     \\ \cmidrule(l){2-22} 
		& AdvDM$_{800}^{900}$    & 168.03          & 166.58  & 215.81  & 171.55   & 15.38   & 15.91   & 10.05   & 15.76    & 34.57   & 34.57   & 31.36   & 34.20    & 0.43    & 0.43    & 0.19    & 0.23     & 0.46     & 0.45     & 0.29    & 0.38     \\
		& Mist$_{800}^{900}$ (1) & 169.88          & 169.93  & 202.05  & 136.78   & 16.89   & 15.50   & 11.36   & 16.61    & 34.50   & 34.48   & 32.06   & 34.60    & 0.45    & 0.46    & 0.25    & 0.28     & 0.47     & 0.49     & 0.33    & 0.28     \\
		& Mist$_{800}^{900}$ (5) & 168.80          & 167.00  & 213.81  & 171.64   & 15.71   & 15.26   & 9.81    & 15.79    & 34.57   & 34.60   & 31.29   & 34.24    & 0.43    & 0.43    & 0.19    & 0.23     & 0.46     & 0.47     & 0.31    & 0.38     \\ \midrule
		\multirow{9}{*}{TVM}      & AdvDM                                      & 173.80          & 173.91  & 213.00  & 150.90   & 15.11   & 14.95   & 9.71    & 15.24    & 33.99   & 33.96   & 31.10   & 34.17    & 0.43    & 0.43    & 0.27    & 0.21     & 0.43     & 0.44     & 0.31    & 0.28     \\
		& SDS                                        & 175.08          & 175.35  & 215.86  & 149.69   & 15.50   & 15.73   & 9.76    & 15.88    & 33.99   & 34.09   & 30.80   & 34.14    & 0.43    & 0.43    & 0.25    & 0.25     & 0.47     & 0.47     & 0.31    & 0.32     \\
		& Chain                                      & 176.11          & 175.96  & 222.36  & 148.43   & 14.17   & 14.78   & 8.77    & 15.72    & 33.99   & 33.90   & 30.66   & 34.19    & 0.41    & 0.41    & 0.23    & 0.24     & 0.44     & 0.44     & 0.27    & 0.32     \\
		& Mist (0)                                   & 177.01          & 176.85  & 225.51  & 147.70   & 15.12   & 15.82   & 8.31    & 15.89    & 33.88   & 34.01   & 30.47   & 34.14    & 0.43    & 0.42    & 0.24    & 0.23     & 0.46     & 0.46     & 0.29    & 0.30     \\
		& Mist (1)                                   & 176.23          & 177.80  & 226.19  & 149.39   & 14.95   & 15.55   & 9.12    & 15.30    & 33.81   & 33.97   & 30.60   & 34.15    & 0.43    & 0.42    & 0.24    & 0.23     & 0.46     & 0.46     & 0.29    & 0.30     \\
		& Mist (5)                                   & 172.80          & 174.49  & 213.84  & 151.58   & 14.80   & 14.88   & 9.87    & 15.15    & 33.95   & 33.96   & 31.14   & 34.17    & 0.43    & 0.42    & 0.27    & 0.21     & 0.44     & 0.43     & 0.31    & 0.28     \\ \cmidrule(l){2-22} 
		& AdvDM$_{800}^{900}$    & 171.16          & 173.60  & 215.45  & 154.71   & 15.20   & 15.82   & 9.92    & 14.90    & 34.07   & 34.03   & 31.25   & 34.23    & 0.43    & 0.44    & 0.27    & 0.23     & 0.44     & 0.44     & 0.31    & 0.30     \\
		& Mist$_{800}^{900}$ (1) & 177.29          & 177.80  & 222.25  & 149.47   & 15.11   & 15.03   & 9.04    & 15.42    & 33.85   & 33.97   & 30.64   & 34.12    & 0.42    & 0.42    & 0.23    & 0.24     & 0.45     & 0.46     & 0.29    & 0.30     \\
		& Mist$_{800}^{900}$ (5) & 170.84          & 173.36  & 213.76  & 154.74   & 14.71   & 15.60   & 9.44    & 14.53    & 34.03   & 34.05   & 31.23   & 34.19    & 0.43    & 0.43    & 0.27    & 0.23     & 0.43     & 0.44     & 0.31    & 0.30     \\ \bottomrule
	\end{tabular}}}
\end{table}

\section{More Visualizations} \label{sec:moreVisual}

We post more visualizations of images generated with benign images and AEs. The surrogate model used to manufacture AEs for image variation is SD-v1-4. When attacking SD-Inpainting and SD-2-Inpainting, Chain~\cite{benchmark} uses SD-Inpainting and SD-2-Inpainting as surrogate models, respectively. Other adversarial attacks use SD-v1-4 as their surrogate model.
\newpage
\subsection{Attacking SD-v1-4}
\begin{figure*}[!h]
	\centering
	\begin{minipage}{\linewidth}
		\centering
		\includegraphics[width=0.12\linewidth]{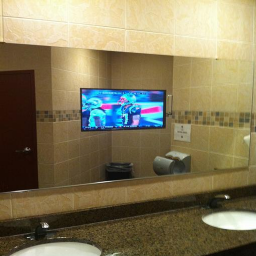}
		\includegraphics[width=0.12\linewidth]{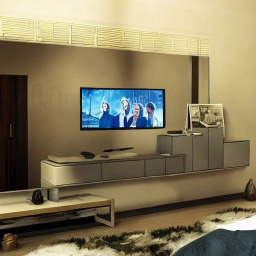}
		\includegraphics[width=0.12\linewidth]{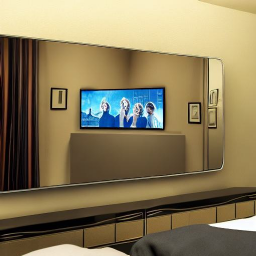}
		\includegraphics[width=0.12\linewidth]{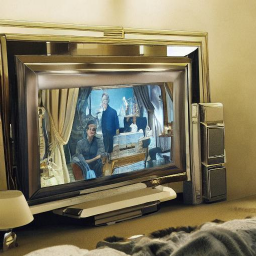}
		\includegraphics[width=0.12\linewidth]{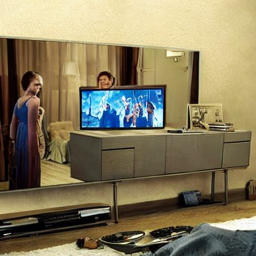}
		\includegraphics[width=0.12\linewidth]{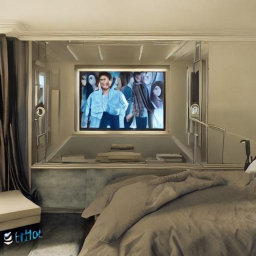}
		\includegraphics[width=0.12\linewidth]{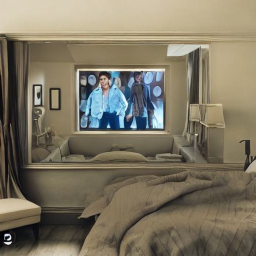}
		\includegraphics[width=0.12\linewidth]{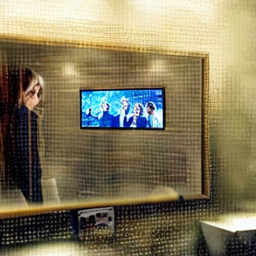}
	\end{minipage}\\
	``A movie on TV reflects in a bedroom mirror.''
	\begin{minipage}{\linewidth}
		\centering
		\includegraphics[width=0.12\linewidth]{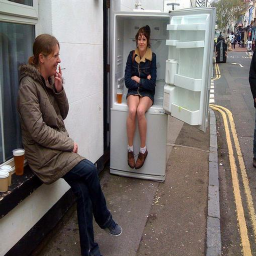}
		\includegraphics[width=0.12\linewidth]{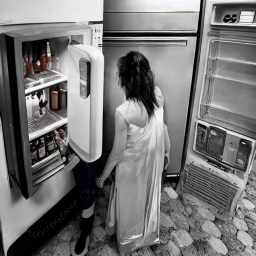}
		\includegraphics[width=0.12\linewidth]{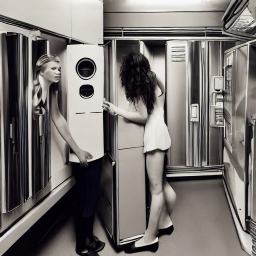}
		\includegraphics[width=0.12\linewidth]{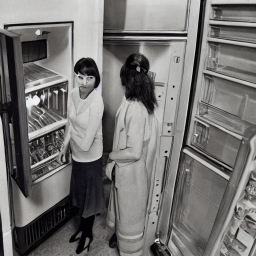}
		\includegraphics[width=0.12\linewidth]{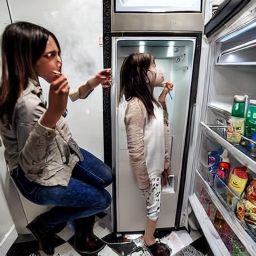}
		\includegraphics[width=0.12\linewidth]{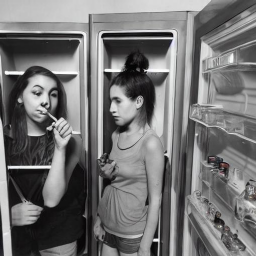}
		\includegraphics[width=0.12\linewidth]{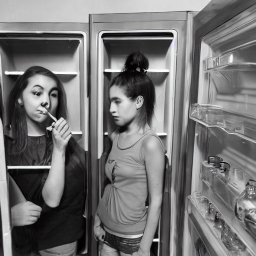}
		\includegraphics[width=0.12\linewidth]{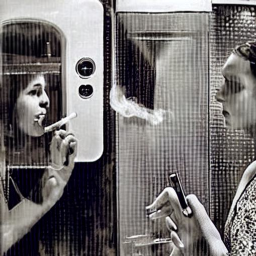}
	\end{minipage}\\
	``A woman smoking a cigarette stands next to a girl inside a refrigerator.''
	\begin{minipage}{\linewidth}
		\centering
		\includegraphics[width=0.12\linewidth]{./ivVisual/ori/000000052565.pdf}
		\includegraphics[width=0.12\linewidth]{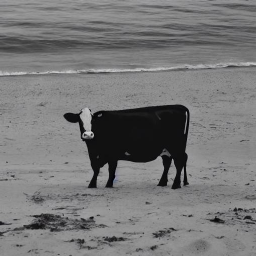}
		\includegraphics[width=0.12\linewidth]{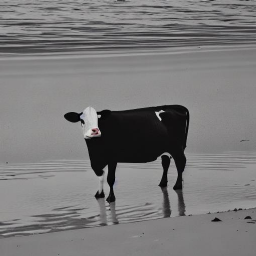}
		\includegraphics[width=0.12\linewidth]{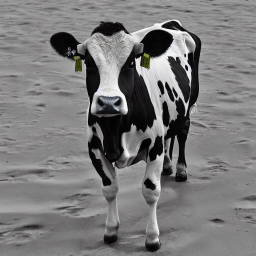}
		\includegraphics[width=0.12\linewidth]{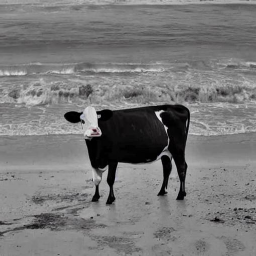}
		\includegraphics[width=0.12\linewidth]{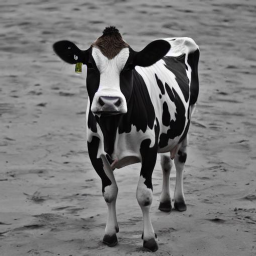}
		\includegraphics[width=0.12\linewidth]{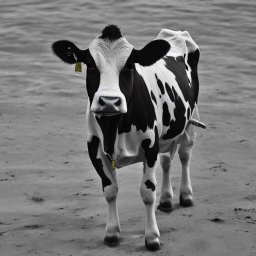}
		\includegraphics[width=0.12\linewidth]{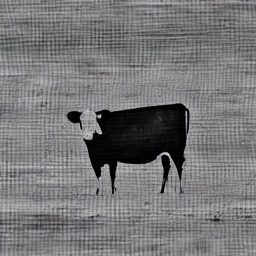}
	\end{minipage}\\
	``A cow positioned on a beach by a body of water.''
	\begin{minipage}{\linewidth}
		\centering
		\includegraphics[width=0.12\linewidth]{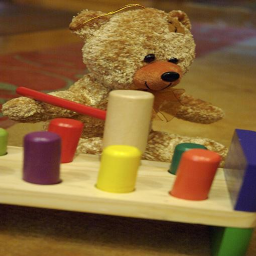}
		\includegraphics[width=0.12\linewidth]{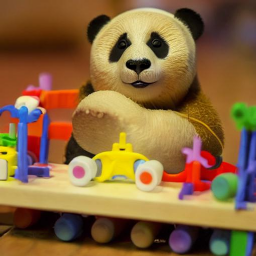}
		\includegraphics[width=0.12\linewidth]{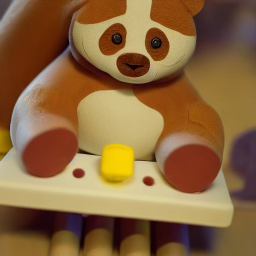}
		\includegraphics[width=0.12\linewidth]{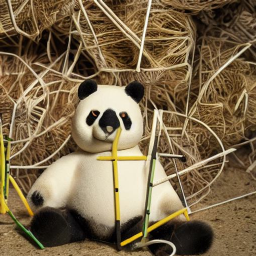}
		\includegraphics[width=0.12\linewidth]{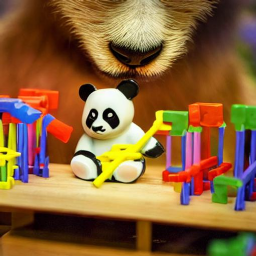}
		\includegraphics[width=0.12\linewidth]{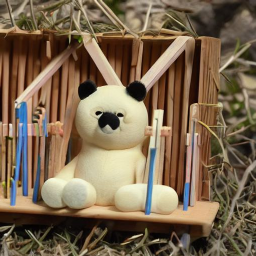}
		\includegraphics[width=0.12\linewidth]{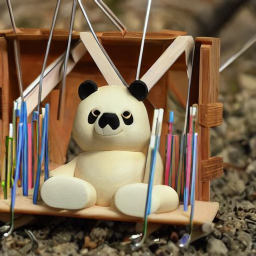}
		\includegraphics[width=0.12\linewidth]{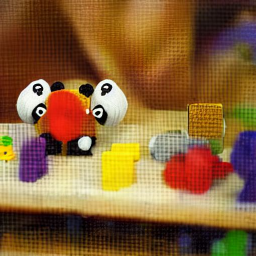}
	\end{minipage}\\
	``A panda bear is sitting near to a toy with pegs.''
	\begin{minipage}{\linewidth}
		\centering
		\includegraphics[width=0.12\linewidth]{./ivVisual/ori/000000105264.pdf}
		\includegraphics[width=0.12\linewidth]{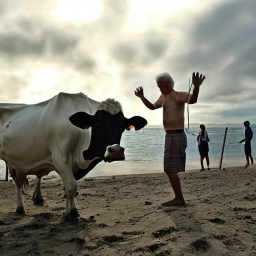}
		\includegraphics[width=0.12\linewidth]{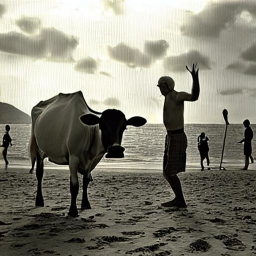}
		\includegraphics[width=0.12\linewidth]{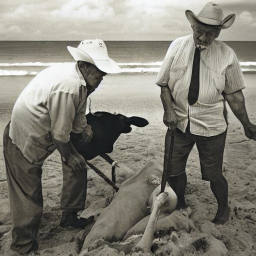}
		\includegraphics[width=0.12\linewidth]{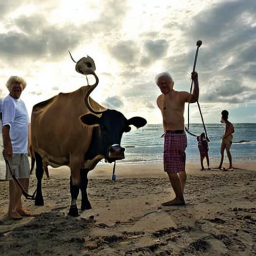}
		\includegraphics[width=0.12\linewidth]{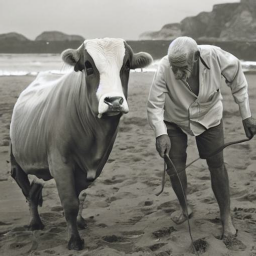}
		\includegraphics[width=0.12\linewidth]{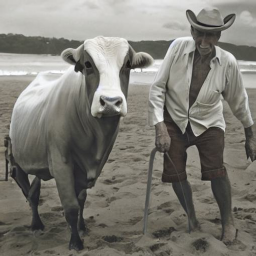}
		\includegraphics[width=0.12\linewidth]{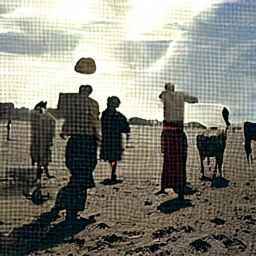}
	\end{minipage}\\
	``Elderly men playing on the beach with a cow in the foreground.''
	\begin{minipage}{\linewidth}
		\centering
		\includegraphics[width=0.12\linewidth]{./ivVisual/ori/000000107087.pdf}
		\includegraphics[width=0.12\linewidth]{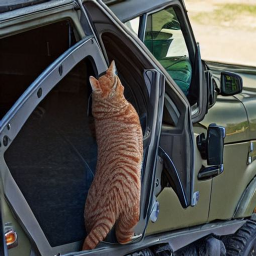}
		\includegraphics[width=0.12\linewidth]{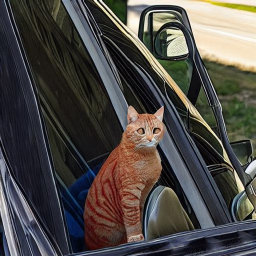}
		\includegraphics[width=0.12\linewidth]{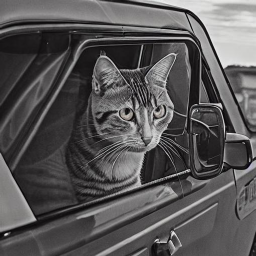}
		\includegraphics[width=0.12\linewidth]{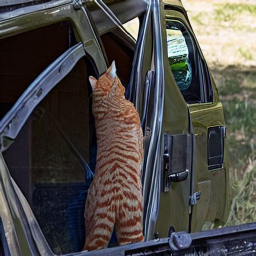}
		\includegraphics[width=0.12\linewidth]{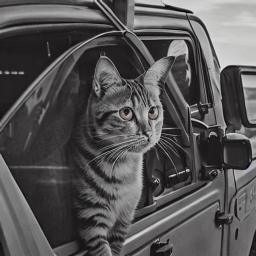}
		\includegraphics[width=0.12\linewidth]{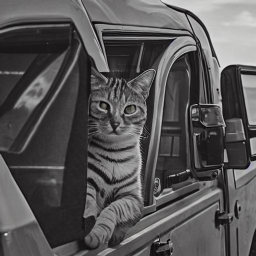}
		\includegraphics[width=0.12\linewidth]{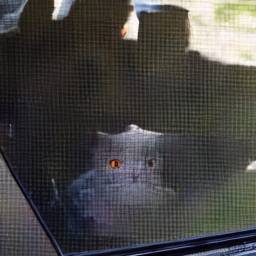}
	\end{minipage}\\
	``An orange cat looking out the windshield of a jeep.''
	\begin{minipage}{\linewidth}
		\centering
		\includegraphics[width=0.12\linewidth]{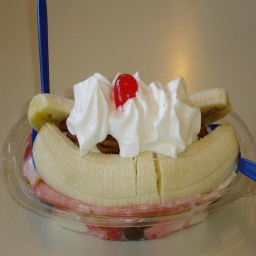}
		\includegraphics[width=0.12\linewidth]{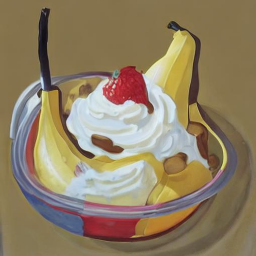}
		\includegraphics[width=0.12\linewidth]{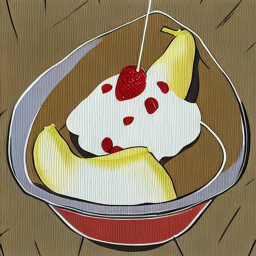}
		\includegraphics[width=0.12\linewidth]{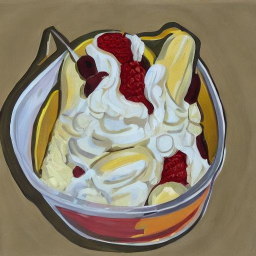}
		\includegraphics[width=0.12\linewidth]{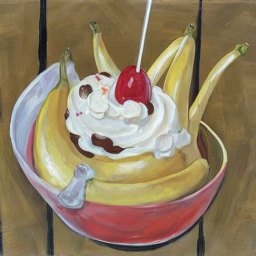}
		\includegraphics[width=0.12\linewidth]{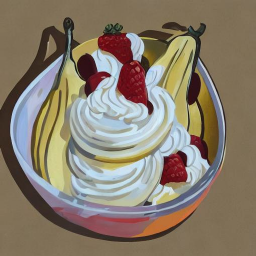}
		\includegraphics[width=0.12\linewidth]{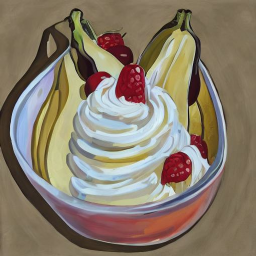}
		\includegraphics[width=0.12\linewidth]{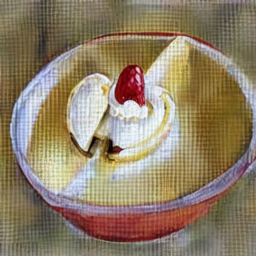}
	\end{minipage}\\
	``A depiction of a banana split in a plastic bowl on a canvas.''
	\begin{minipage}{\linewidth}
		\centering
		\includegraphics[width=0.12\linewidth]{./ivVisual/ori/000000545007.pdf}
		\includegraphics[width=0.12\linewidth]{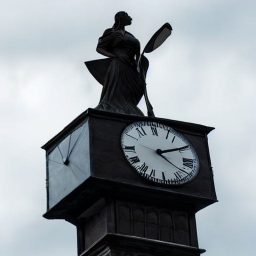}
		\includegraphics[width=0.12\linewidth]{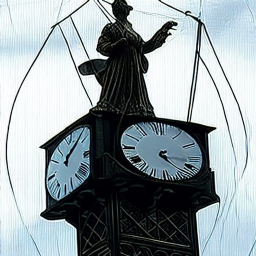}
		\includegraphics[width=0.12\linewidth]{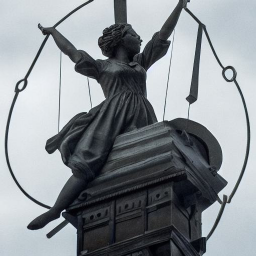}
		\includegraphics[width=0.12\linewidth]{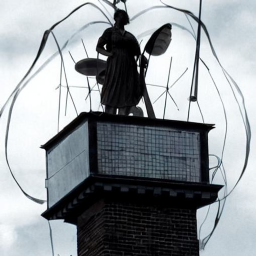}
		\includegraphics[width=0.12\linewidth]{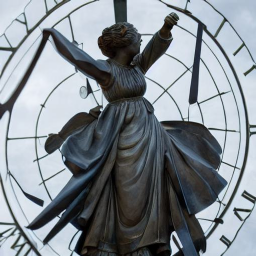}
		\includegraphics[width=0.12\linewidth]{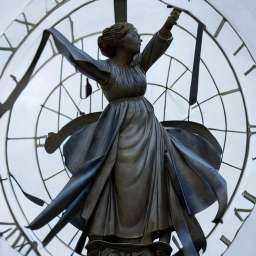}
		\includegraphics[width=0.12\linewidth]{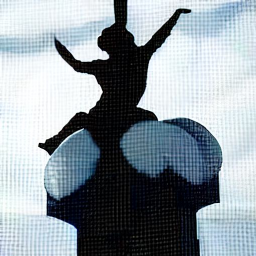}
	\end{minipage}\\
	``A statue of a woman stood atop a clock.''
	\caption{From left to right: Original images, Benign, Chain~\cite{benchmark}, SDS~\cite{sds}, AdvDM~\cite{AdvDM}, Mist (0)~\cite{mist}, Mist (1)~\cite{mist}, AdvDM$_{800}^{900}$ (Ours). View this figure in color. Use zoom-in to check the texture.}
\end{figure*}
\newpage
\subsection{Attacking SD-v1-5}
\begin{figure*}[!h]
	\centering
	\begin{minipage}{\linewidth}
		\centering
		\includegraphics[width=0.12\linewidth]{./ivVisual/ori/000000048564.pdf}
		\includegraphics[width=0.12\linewidth]{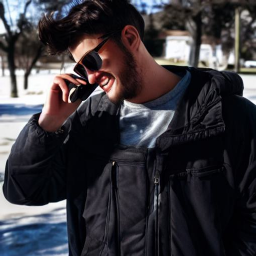}
		\includegraphics[width=0.12\linewidth]{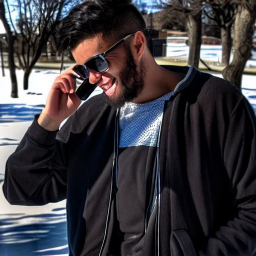}
		\includegraphics[width=0.12\linewidth]{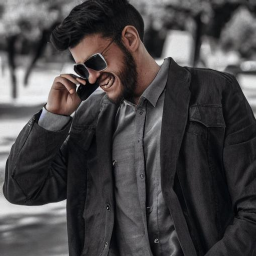}
		\includegraphics[width=0.12\linewidth]{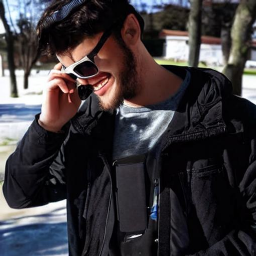}
		\includegraphics[width=0.12\linewidth]{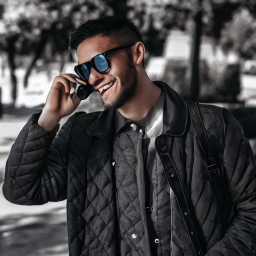}
		\includegraphics[width=0.12\linewidth]{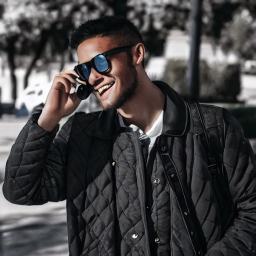}
		\includegraphics[width=0.12\linewidth]{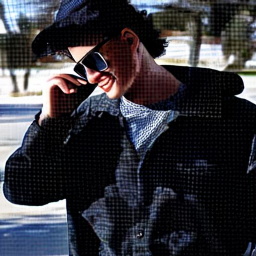}
	\end{minipage}\\
	``A young man with sunglasses talking on a cell phone on a sunny day.''
	\begin{minipage}{\linewidth}
		\centering
		\includegraphics[width=0.12\linewidth]{./ivVisual/ori/000000545958.pdf}
		\includegraphics[width=0.12\linewidth]{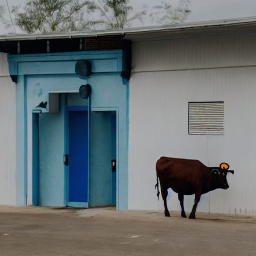}
		\includegraphics[width=0.12\linewidth]{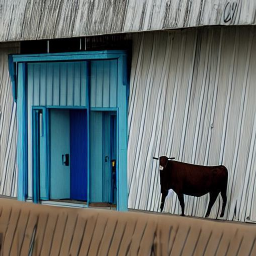}
		\includegraphics[width=0.12\linewidth]{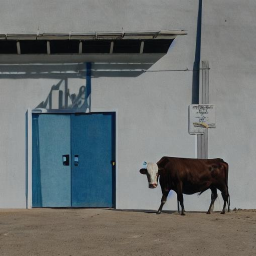}
		\includegraphics[width=0.12\linewidth]{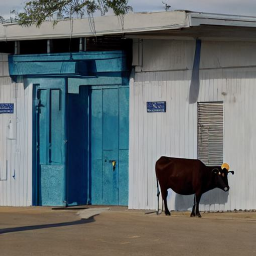}
		\includegraphics[width=0.12\linewidth]{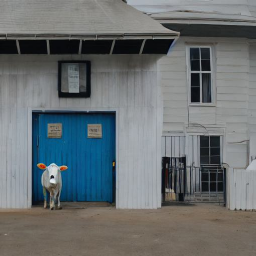}
		\includegraphics[width=0.12\linewidth]{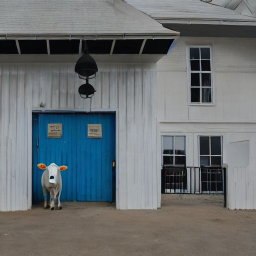}
		\includegraphics[width=0.12\linewidth]{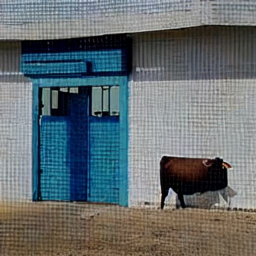}
	\end{minipage}\\
	``A white building with a blue entrance has a cow standing outside of it.''
	\begin{minipage}{\linewidth}
		\centering
		\includegraphics[width=0.12\linewidth]{./ivVisual/ori/000000537802.pdf}
		\includegraphics[width=0.12\linewidth]{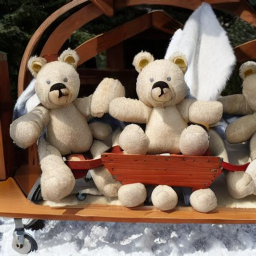}
		\includegraphics[width=0.12\linewidth]{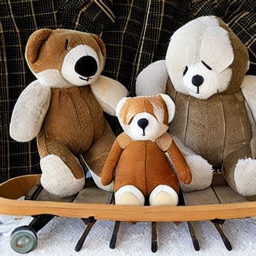}
		\includegraphics[width=0.12\linewidth]{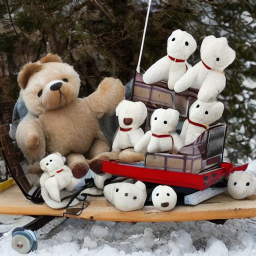}
		\includegraphics[width=0.12\linewidth]{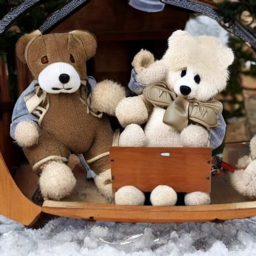}
		\includegraphics[width=0.12\linewidth]{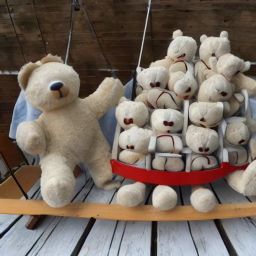}
		\includegraphics[width=0.12\linewidth]{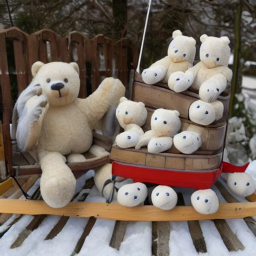}
		\includegraphics[width=0.12\linewidth]{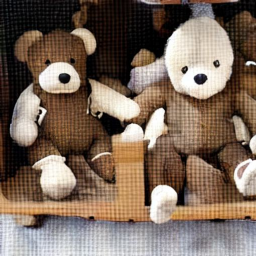}
	\end{minipage}\\
	``A set of plush toy teddy bears can be seen sitting in a sled.''
	\begin{minipage}{\linewidth}
		\centering
		\includegraphics[width=0.12\linewidth]{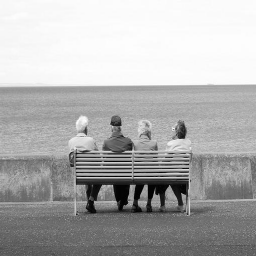}
		\includegraphics[width=0.12\linewidth]{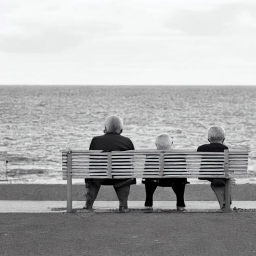}
		\includegraphics[width=0.12\linewidth]{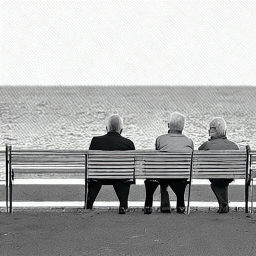}
		\includegraphics[width=0.12\linewidth]{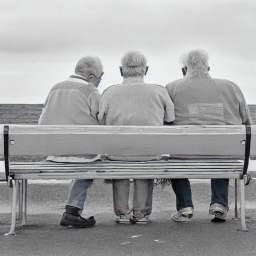}
		\includegraphics[width=0.12\linewidth]{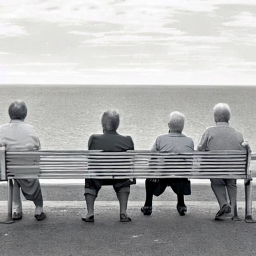}
		\includegraphics[width=0.12\linewidth]{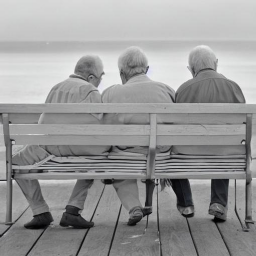}
		\includegraphics[width=0.12\linewidth]{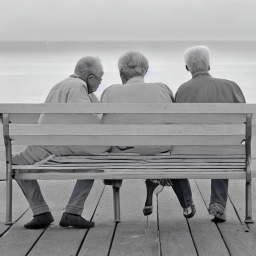}
		\includegraphics[width=0.12\linewidth]{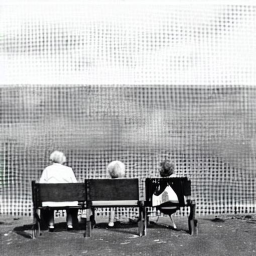}
	\end{minipage}\\
	``Four elders relax on a park bench, gazing out at the sea.''
	\begin{minipage}{\linewidth}
		\centering
		\includegraphics[width=0.12\linewidth]{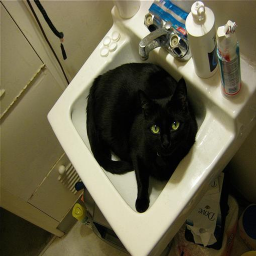}
		\includegraphics[width=0.12\linewidth]{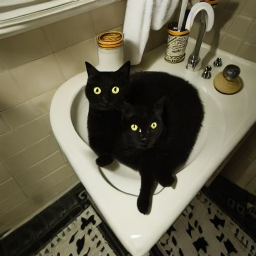}
		\includegraphics[width=0.12\linewidth]{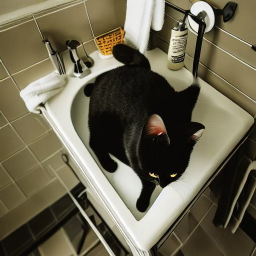}
		\includegraphics[width=0.12\linewidth]{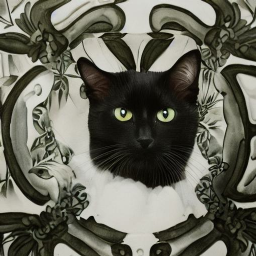}
		\includegraphics[width=0.12\linewidth]{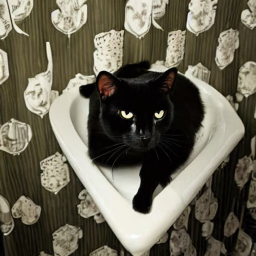}
		\includegraphics[width=0.12\linewidth]{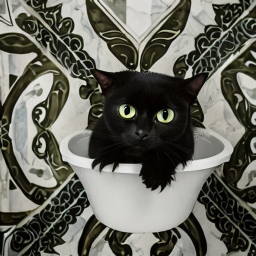}
		\includegraphics[width=0.12\linewidth]{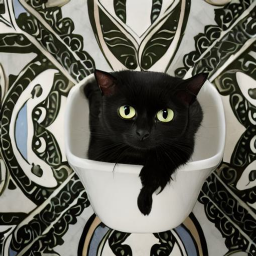}
		\includegraphics[width=0.12\linewidth]{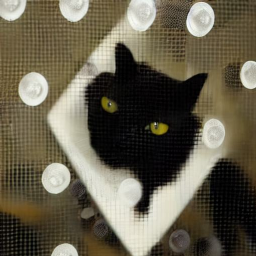}
	\end{minipage}\\
	``A black cat rests inside a white bathroom sink.''
	\begin{minipage}{\linewidth}
		\centering
		\includegraphics[width=0.12\linewidth]{./ivVisual/ori/000000283412.pdf}
		\includegraphics[width=0.12\linewidth]{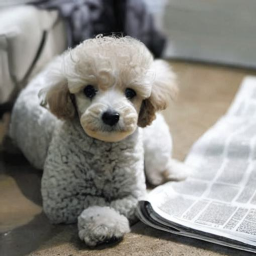}
		\includegraphics[width=0.12\linewidth]{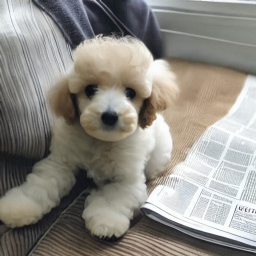}
		\includegraphics[width=0.12\linewidth]{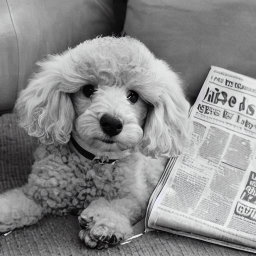}
		\includegraphics[width=0.12\linewidth]{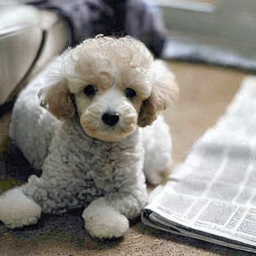}
		\includegraphics[width=0.12\linewidth]{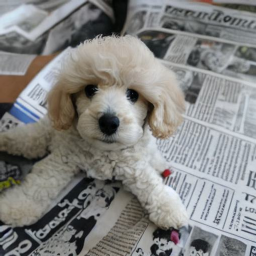}
		\includegraphics[width=0.12\linewidth]{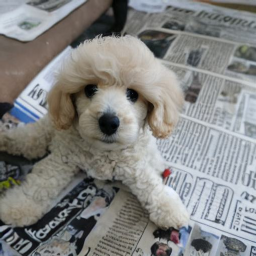}
		\includegraphics[width=0.12\linewidth]{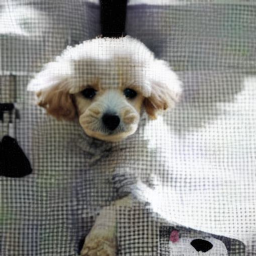}
	\end{minipage}\\
	``You can see a small poodle puppy lying near a newspaper with an expression of guilt on its face.''
	\begin{minipage}{\linewidth}
		\centering
		\includegraphics[width=0.12\linewidth]{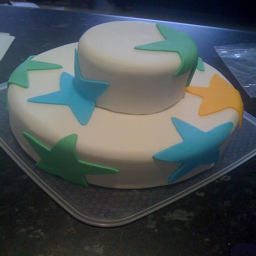}
		\includegraphics[width=0.12\linewidth]{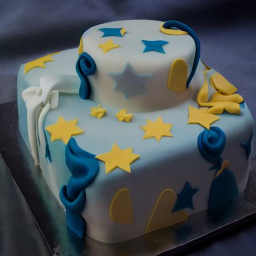}
		\includegraphics[width=0.12\linewidth]{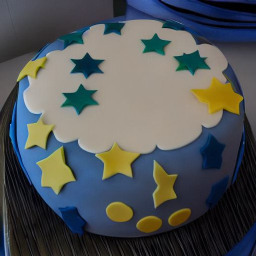}
		\includegraphics[width=0.12\linewidth]{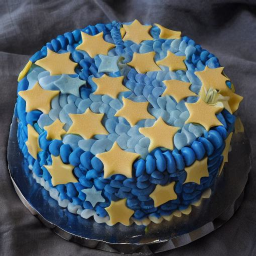}
		\includegraphics[width=0.12\linewidth]{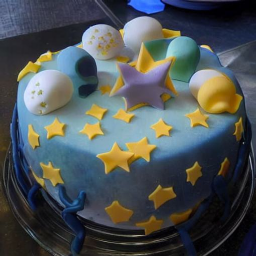}
		\includegraphics[width=0.12\linewidth]{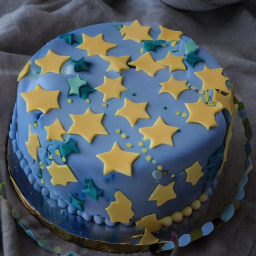}
		\includegraphics[width=0.12\linewidth]{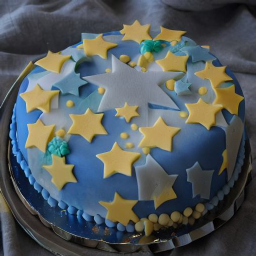}
		\includegraphics[width=0.12\linewidth]{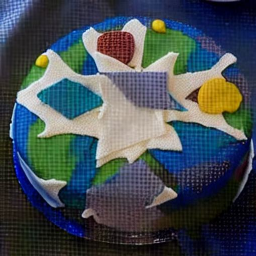}
	\end{minipage}\\
	``The cake is embellished with fondant stars in shades of blue, yellow, and green.''
	\begin{minipage}{\linewidth}
		\centering
		\includegraphics[width=0.12\linewidth]{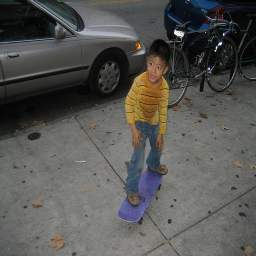}
		\includegraphics[width=0.12\linewidth]{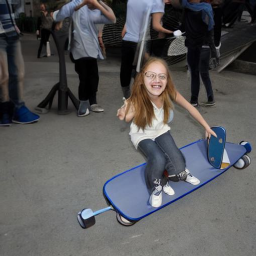}
		\includegraphics[width=0.12\linewidth]{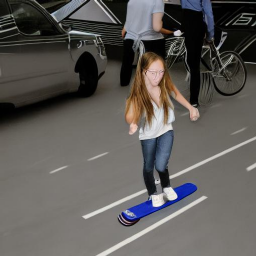}
		\includegraphics[width=0.12\linewidth]{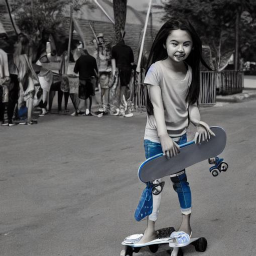}
		\includegraphics[width=0.12\linewidth]{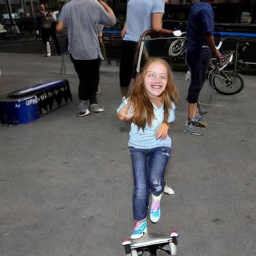}
		\includegraphics[width=0.12\linewidth]{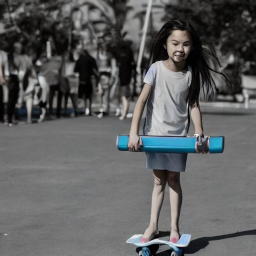}
		\includegraphics[width=0.12\linewidth]{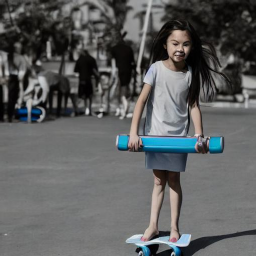}
		\includegraphics[width=0.12\linewidth]{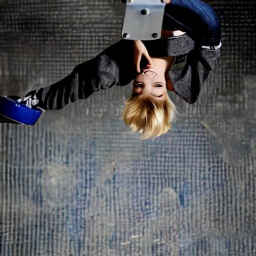}
	\end{minipage}\\
	``A young girl with a silver shirt is on a blue skateboard.''
	\caption{From left to right: Original images, Benign, Chain~\cite{benchmark}, SDS~\cite{sds}, AdvDM~\cite{AdvDM}, Mist (0)~\cite{mist}, Mist (1)~\cite{mist}, AdvDM$_{800}^{900}$ (Ours). View this figure in color. Use zoom-in to check the texture.}
\end{figure*}
\newpage
\subsection{Attacking SD-v2-1}
\begin{figure*}[!h]
	\centering
	\begin{minipage}{\linewidth}
		\centering
		\includegraphics[width=0.12\linewidth]{./ivVisual/ori/000000061658.pdf}
		\includegraphics[width=0.12\linewidth]{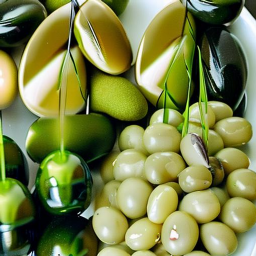}
		\includegraphics[width=0.12\linewidth]{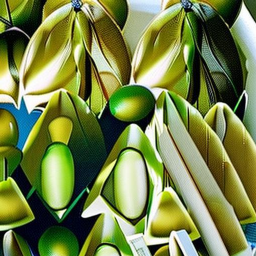}
		\includegraphics[width=0.12\linewidth]{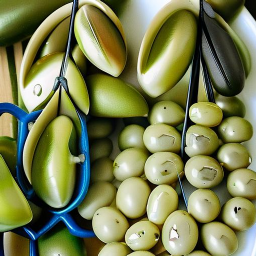}
		\includegraphics[width=0.12\linewidth]{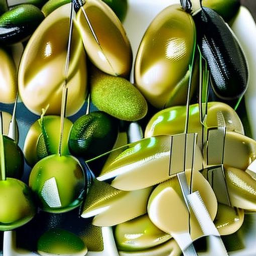}
		\includegraphics[width=0.12\linewidth]{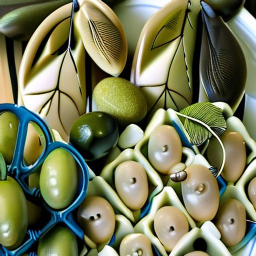}
		\includegraphics[width=0.12\linewidth]{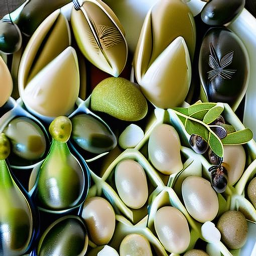}
		\includegraphics[width=0.12\linewidth]{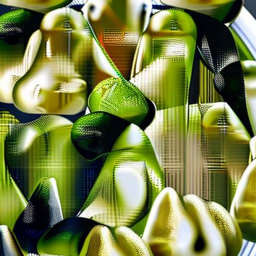}
	\end{minipage}\\
	``A mixture of black and green olives on a white platter.''
	\begin{minipage}{\linewidth}
		\centering
		\includegraphics[width=0.12\linewidth]{./ivVisual/ori/000000217219.pdf}
		\includegraphics[width=0.12\linewidth]{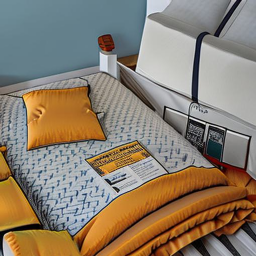}
		\includegraphics[width=0.12\linewidth]{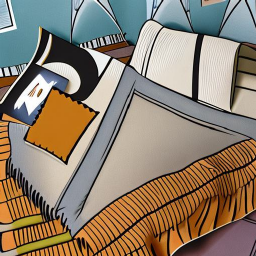}
		\includegraphics[width=0.12\linewidth]{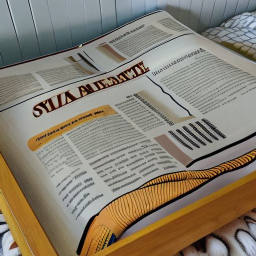}
		\includegraphics[width=0.12\linewidth]{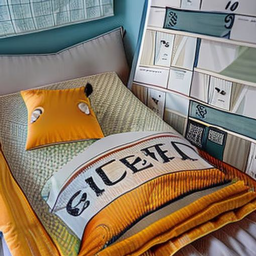}
		\includegraphics[width=0.12\linewidth]{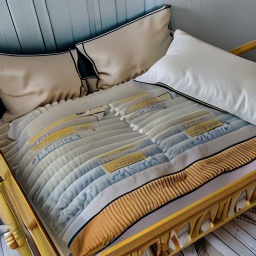}
		\includegraphics[width=0.12\linewidth]{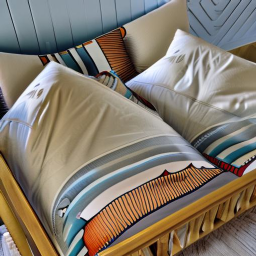}
		\includegraphics[width=0.12\linewidth]{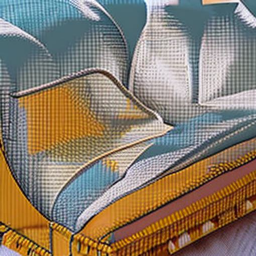}
	\end{minipage}\\
	``A gazette sprawled on a neat bed with a duvet and bolster.''
	\begin{minipage}{\linewidth}
		\centering
		\includegraphics[width=0.12\linewidth]{./ivVisual/ori/000000096493.pdf}
		\includegraphics[width=0.12\linewidth]{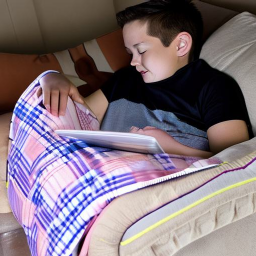}
		\includegraphics[width=0.12\linewidth]{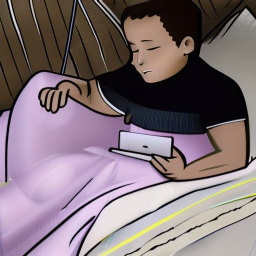}
		\includegraphics[width=0.12\linewidth]{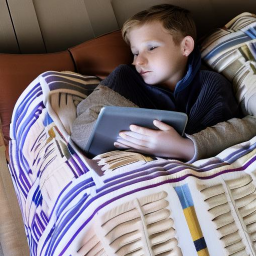}
		\includegraphics[width=0.12\linewidth]{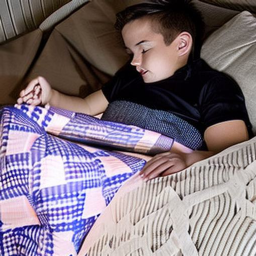}
		\includegraphics[width=0.12\linewidth]{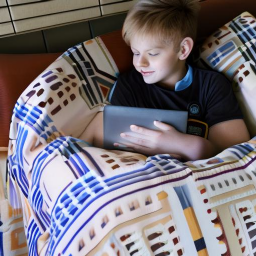}
		\includegraphics[width=0.12\linewidth]{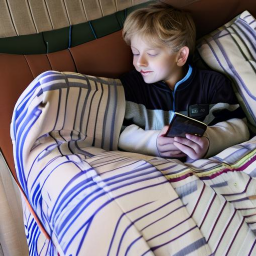}
		\includegraphics[width=0.12\linewidth]{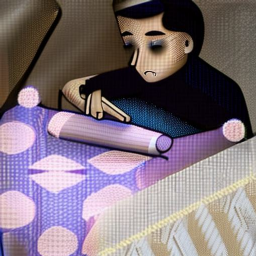}
	\end{minipage}\\
	``A boy wrapped in a comforter holds a tablet on a couch.''
	\begin{minipage}{\linewidth}
		\centering
		\includegraphics[width=0.12\linewidth]{./ivVisual/ori/000000105264.pdf}
		\includegraphics[width=0.12\linewidth]{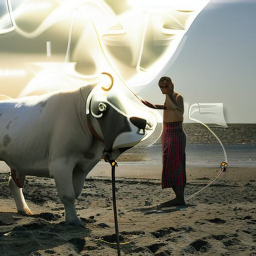}
		\includegraphics[width=0.12\linewidth]{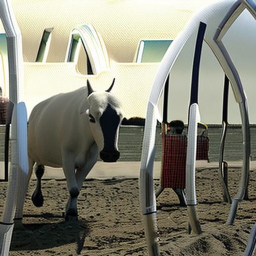}
		\includegraphics[width=0.12\linewidth]{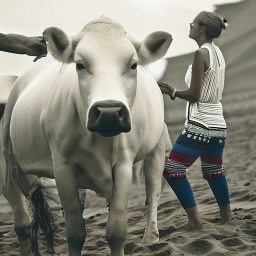}
		\includegraphics[width=0.12\linewidth]{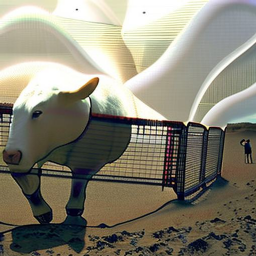}
		\includegraphics[width=0.12\linewidth]{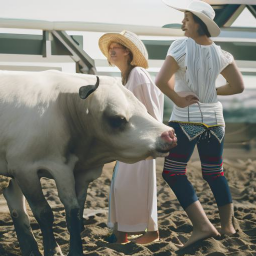}
		\includegraphics[width=0.12\linewidth]{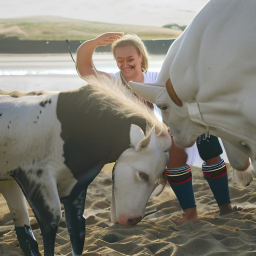}
		\includegraphics[width=0.12\linewidth]{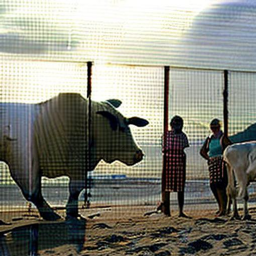}
	\end{minipage}\\
	``People on the beach playing with a white cow.''
	\begin{minipage}{\linewidth}
		\centering
		\includegraphics[width=0.12\linewidth]{./ivVisual/ori/000000125850.pdf}
		\includegraphics[width=0.12\linewidth]{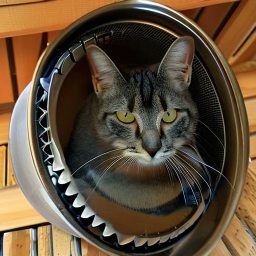}
		\includegraphics[width=0.12\linewidth]{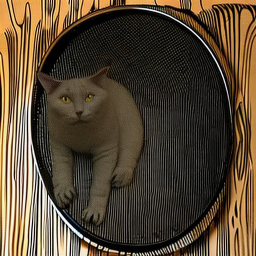}
		\includegraphics[width=0.12\linewidth]{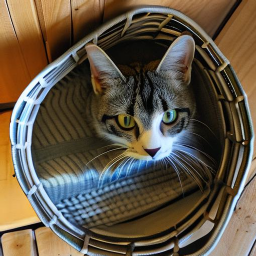}
		\includegraphics[width=0.12\linewidth]{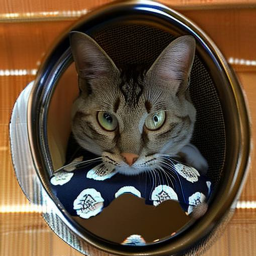}
		\includegraphics[width=0.12\linewidth]{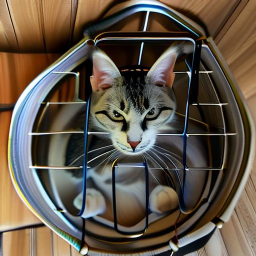}
		\includegraphics[width=0.12\linewidth]{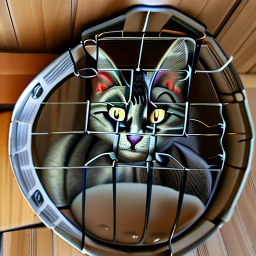}
		\includegraphics[width=0.12\linewidth]{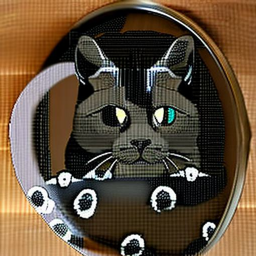}
	\end{minipage}\\
	``Curled up and dozing off, the cat has made a round tin its new bed.''
	\begin{minipage}{\linewidth}
		\centering
		\includegraphics[width=0.12\linewidth]{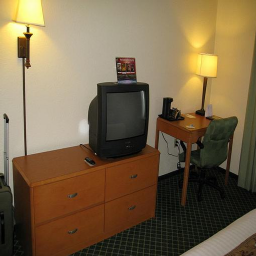}
		\includegraphics[width=0.12\linewidth]{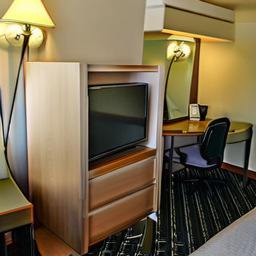}
		\includegraphics[width=0.12\linewidth]{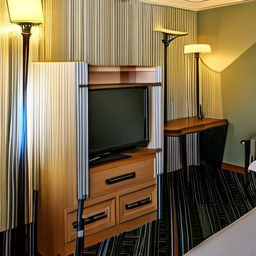}
		\includegraphics[width=0.12\linewidth]{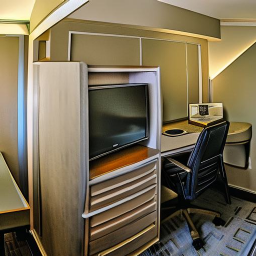}
		\includegraphics[width=0.12\linewidth]{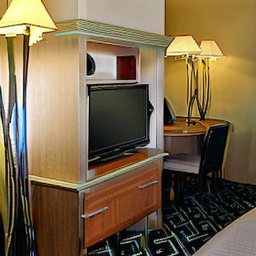}
		\includegraphics[width=0.12\linewidth]{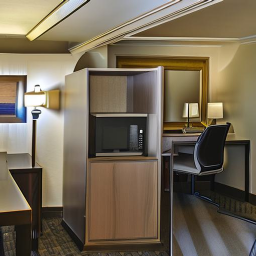}
		\includegraphics[width=0.12\linewidth]{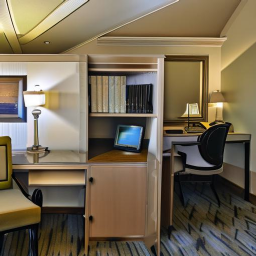}
		\includegraphics[width=0.12\linewidth]{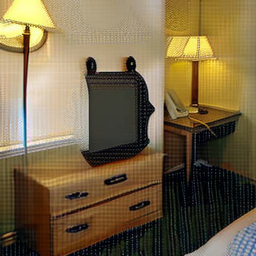}
	\end{minipage}\\
	``A suite with a little TV and a desk for work.''
	\begin{minipage}{\linewidth}
		\centering
		\includegraphics[width=0.12\linewidth]{./ivVisual/ori/000000157390.pdf}
		\includegraphics[width=0.12\linewidth]{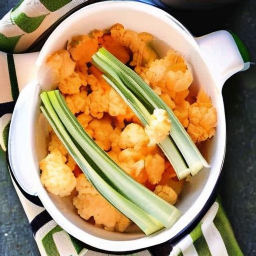}
		\includegraphics[width=0.12\linewidth]{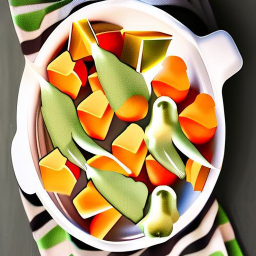}
		\includegraphics[width=0.12\linewidth]{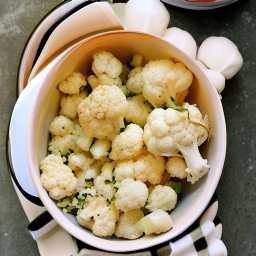}
		\includegraphics[width=0.12\linewidth]{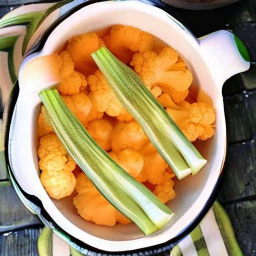}
		\includegraphics[width=0.12\linewidth]{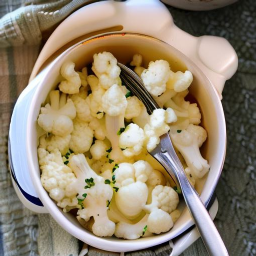}
		\includegraphics[width=0.12\linewidth]{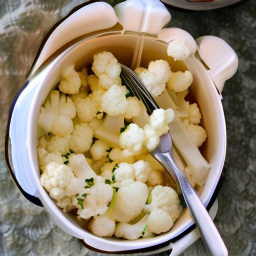}
		\includegraphics[width=0.12\linewidth]{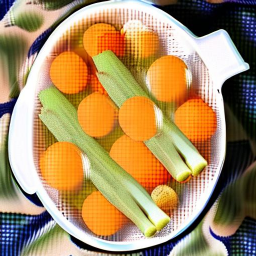}
	\end{minipage}\\
	``A bowl of chopped celery and cauliflower.''
	\begin{minipage}{\linewidth}
		\centering
		\includegraphics[width=0.12\linewidth]{./ivVisual/ori/000000168458.pdf}
		\includegraphics[width=0.12\linewidth]{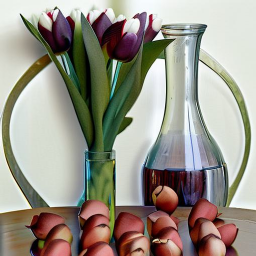}
		\includegraphics[width=0.12\linewidth]{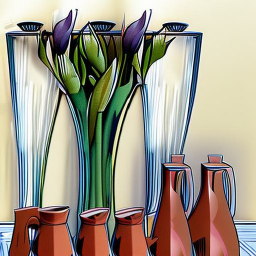}
		\includegraphics[width=0.12\linewidth]{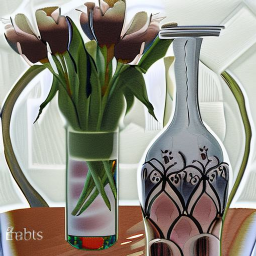}
		\includegraphics[width=0.12\linewidth]{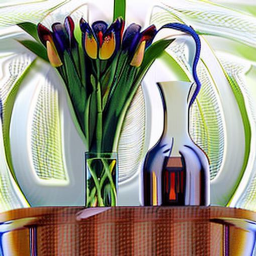}
		\includegraphics[width=0.12\linewidth]{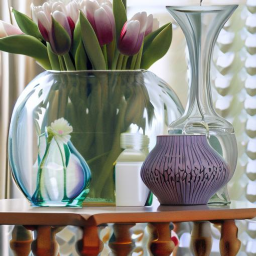}
		\includegraphics[width=0.12\linewidth]{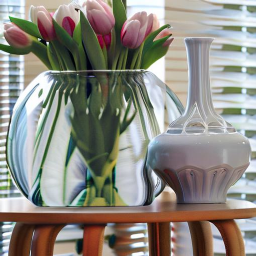}
		\includegraphics[width=0.12\linewidth]{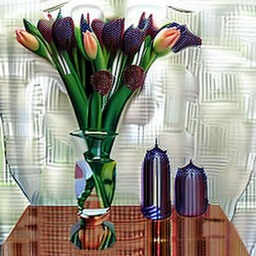}
	\end{minipage}\\
	``A vase full of tulips with a carafe on an end table.''
	\caption{From left to right: Original images, Benign, Chain~\cite{benchmark}, SDS~\cite{sds}, AdvDM~\cite{AdvDM}, Mist (0)~\cite{mist}, Mist (1)~\cite{mist}, AdvDM$_{800}^{900}$ (Ours). View this figure in color. Use zoom-in to check the texture.}
\end{figure*}
\newpage
\subsection{Attacking InstructPix2Pix}
\begin{figure*}[!h]
	\centering
	\begin{minipage}{\linewidth}
		\centering
		\includegraphics[width=0.12\linewidth]{./ivVisual/ori/000000048564.pdf}
		\includegraphics[width=0.12\linewidth]{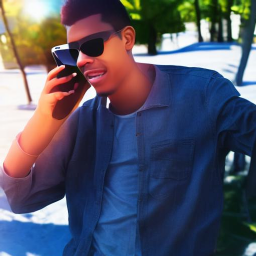}
		\includegraphics[width=0.12\linewidth]{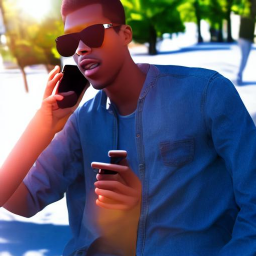}
		\includegraphics[width=0.12\linewidth]{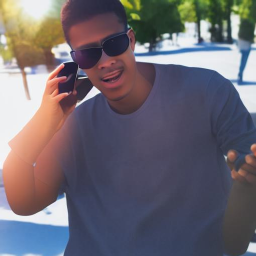}
		\includegraphics[width=0.12\linewidth]{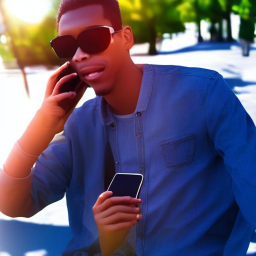}
		\includegraphics[width=0.12\linewidth]{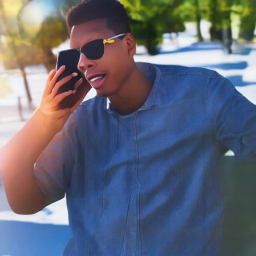}
		\includegraphics[width=0.12\linewidth]{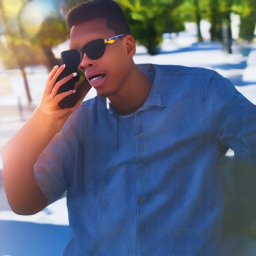}
		\includegraphics[width=0.12\linewidth]{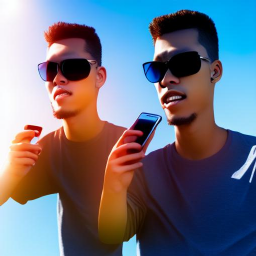}
	\end{minipage}\\
	``A young man with sunglasses talking on a cell phone on a sunny day.''
	\begin{minipage}{\linewidth}
		\centering
		\includegraphics[width=0.12\linewidth]{./ivVisual/ori/000000146457.pdf}
		\includegraphics[width=0.12\linewidth]{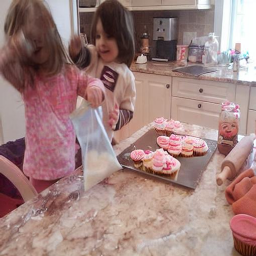}
		\includegraphics[width=0.12\linewidth]{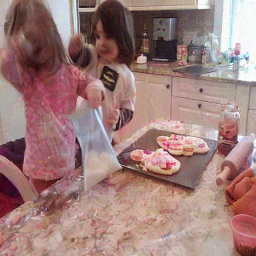}
		\includegraphics[width=0.12\linewidth]{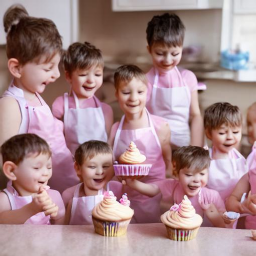}
		\includegraphics[width=0.12\linewidth]{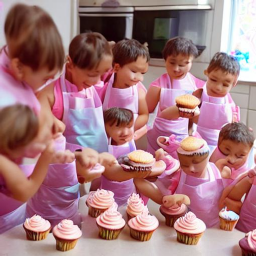}
		\includegraphics[width=0.12\linewidth]{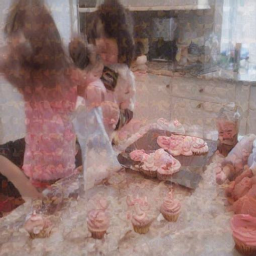}
		\includegraphics[width=0.12\linewidth]{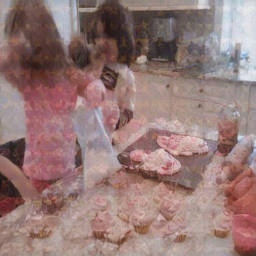}
		\includegraphics[width=0.12\linewidth]{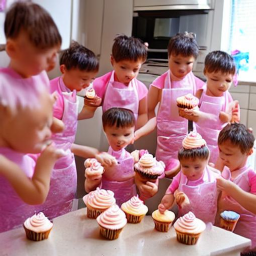}
	\end{minipage}\\
	``Three children being entertained by baking cupcakes at home.''
	\begin{minipage}{\linewidth}
		\centering
		\includegraphics[width=0.12\linewidth]{./ivVisual/ori/000000157418.pdf}
		\includegraphics[width=0.12\linewidth]{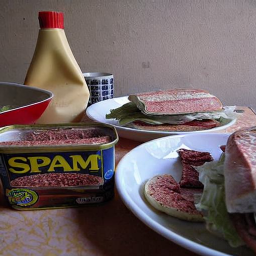}
		\includegraphics[width=0.12\linewidth]{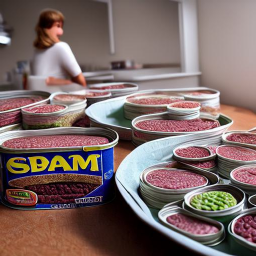}
		\includegraphics[width=0.12\linewidth]{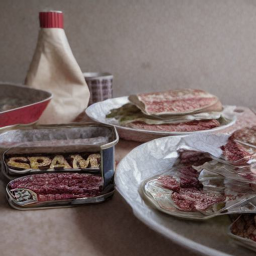}
		\includegraphics[width=0.12\linewidth]{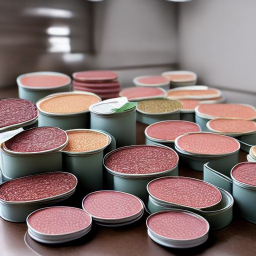}
		\includegraphics[width=0.12\linewidth]{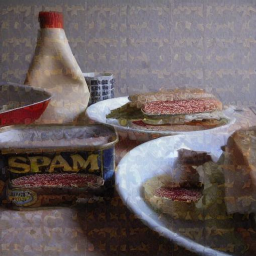}
		\includegraphics[width=0.12\linewidth]{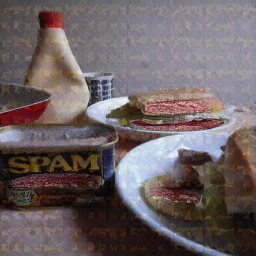}
		\includegraphics[width=0.12\linewidth]{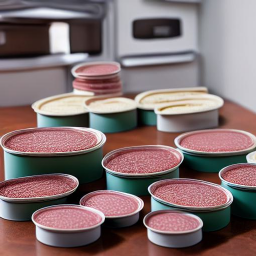}
	\end{minipage}\\
	``Trays with wraps next to a can of potted meat.''
	\begin{minipage}{\linewidth}
		\centering
		\includegraphics[width=0.12\linewidth]{./ivVisual/ori/000000166664.pdf}
		\includegraphics[width=0.12\linewidth]{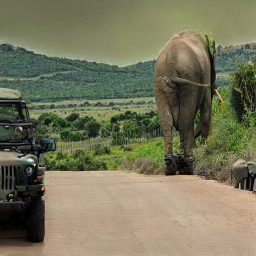}
		\includegraphics[width=0.12\linewidth]{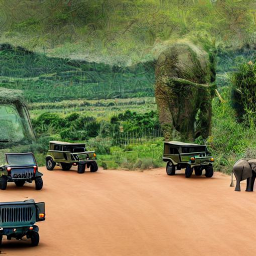}
		\includegraphics[width=0.12\linewidth]{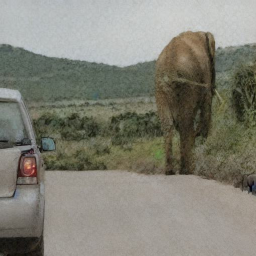}
		\includegraphics[width=0.12\linewidth]{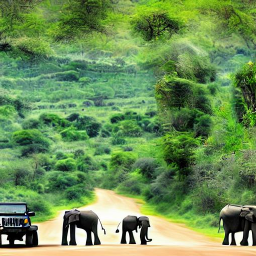}
		\includegraphics[width=0.12\linewidth]{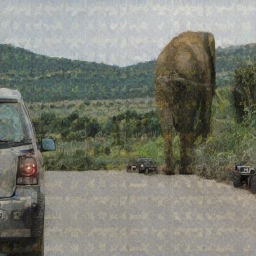}
		\includegraphics[width=0.12\linewidth]{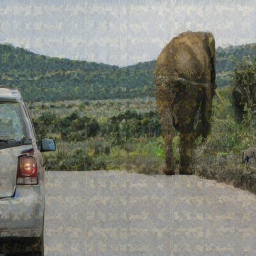}
		\includegraphics[width=0.12\linewidth]{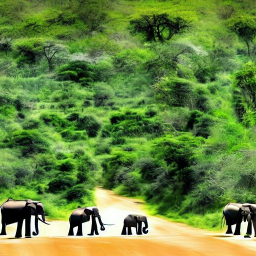}
	\end{minipage}\\
	``An elephant walking through the jungle, with a jeep next to it.''
	\begin{minipage}{\linewidth}
		\centering
		\includegraphics[width=0.12\linewidth]{./ivVisual/ori/000000463842.pdf}
		\includegraphics[width=0.12\linewidth]{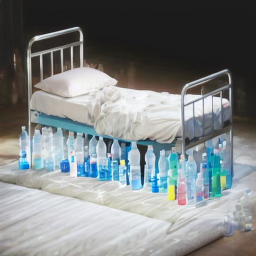}
		\includegraphics[width=0.12\linewidth]{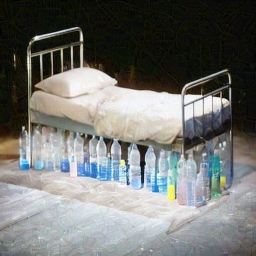}
		\includegraphics[width=0.12\linewidth]{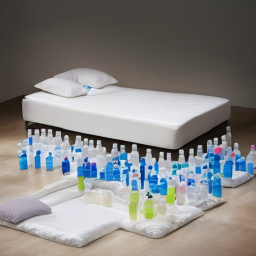}
		\includegraphics[width=0.12\linewidth]{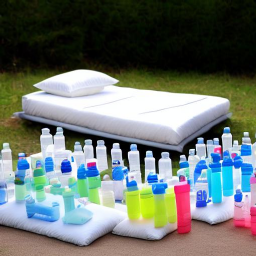}
		\includegraphics[width=0.12\linewidth]{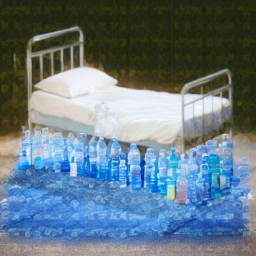}
		\includegraphics[width=0.12\linewidth]{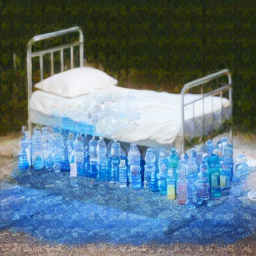}
		\includegraphics[width=0.12\linewidth]{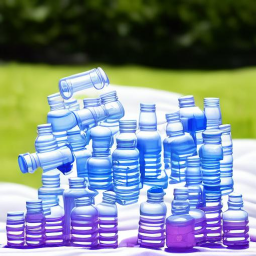}
	\end{minipage}\\
	``A mattress with a pile of juice bottles gathered around it.''
	\begin{minipage}{\linewidth}
		\centering
		\includegraphics[width=0.12\linewidth]{./ivVisual/ori/000000186449.pdf}
		\includegraphics[width=0.12\linewidth]{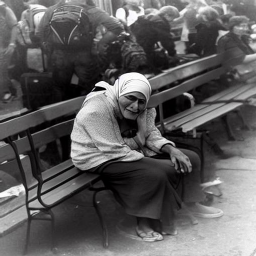}
		\includegraphics[width=0.12\linewidth]{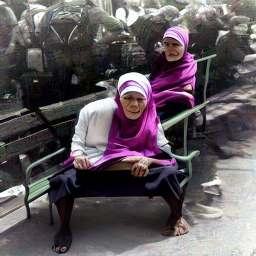}
		\includegraphics[width=0.12\linewidth]{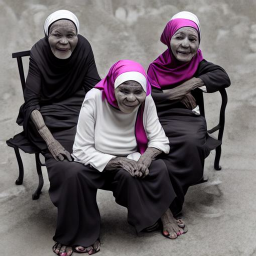}
		\includegraphics[width=0.12\linewidth]{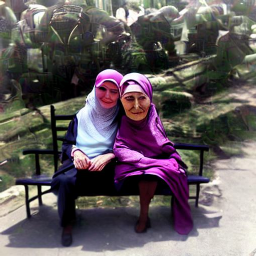}
		\includegraphics[width=0.12\linewidth]{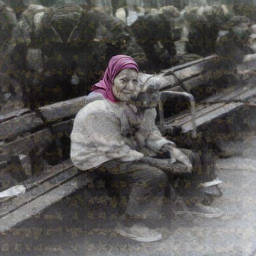}
		\includegraphics[width=0.12\linewidth]{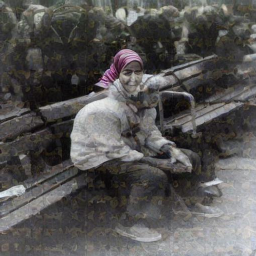}
		\includegraphics[width=0.12\linewidth]{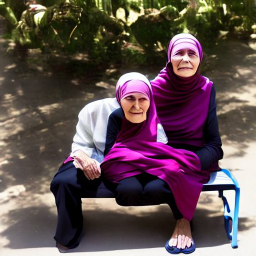}
	\end{minipage}\\
	``A senior lady with a headscarf perched on the bench.''
	\begin{minipage}{\linewidth}
		\centering
		\includegraphics[width=0.12\linewidth]{./ivVisual/ori/000000232244.pdf}
		\includegraphics[width=0.12\linewidth]{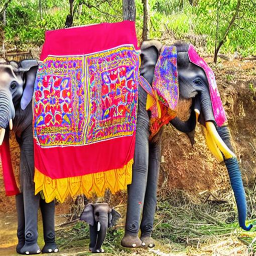}
		\includegraphics[width=0.12\linewidth]{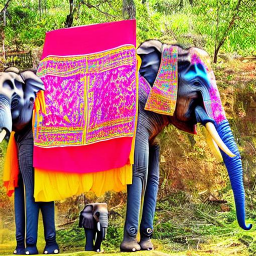}
		\includegraphics[width=0.12\linewidth]{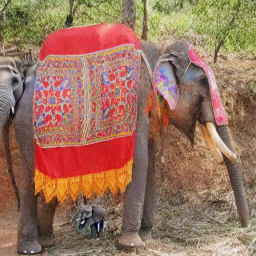}
		\includegraphics[width=0.12\linewidth]{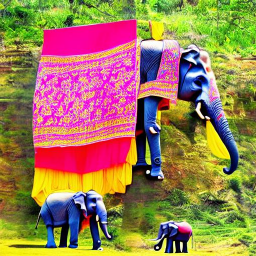}
		\includegraphics[width=0.12\linewidth]{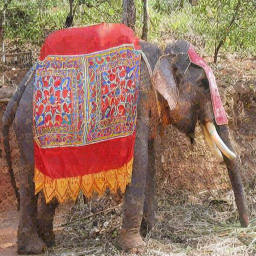}
		\includegraphics[width=0.12\linewidth]{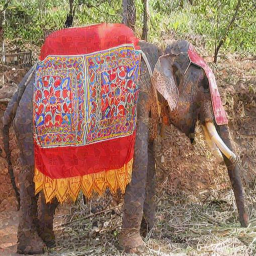}
		\includegraphics[width=0.12\linewidth]{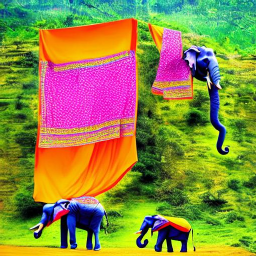}
	\end{minipage}\\
	``an elephant standing elegantly while wearing a bright, colorful cloth in the picturesque countryside.''
	\begin{minipage}{\linewidth}
		\centering
		\includegraphics[width=0.12\linewidth]{./ivVisual/ori/000000370900.pdf}
		\includegraphics[width=0.12\linewidth]{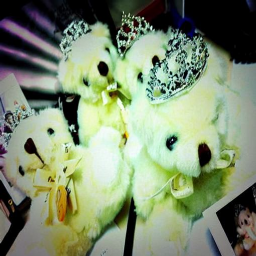}
		\includegraphics[width=0.12\linewidth]{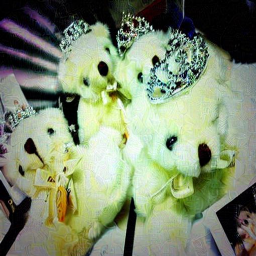}
		\includegraphics[width=0.12\linewidth]{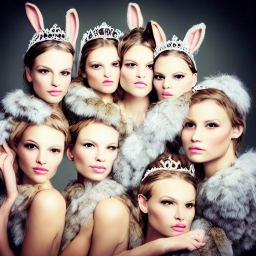}
		\includegraphics[width=0.12\linewidth]{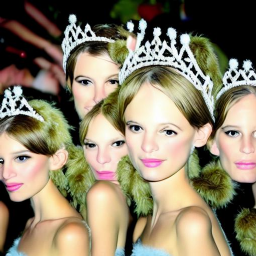}
		\includegraphics[width=0.12\linewidth]{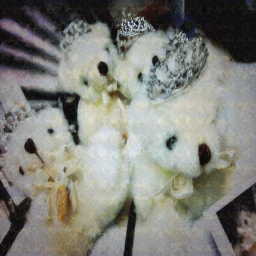}
		\includegraphics[width=0.12\linewidth]{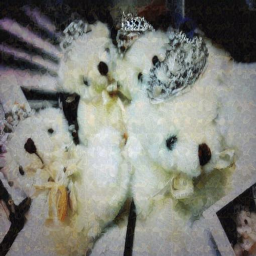}
		\includegraphics[width=0.12\linewidth]{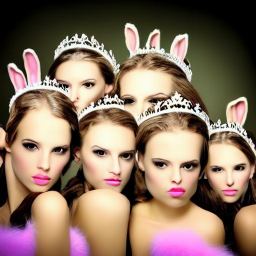}
	\end{minipage}\\
	``A bunch of fuzzy bunnies with tiaras on.''
	\caption{From left to right: Original images, Benign, Chain~\cite{benchmark}, SDS~\cite{sds}, AdvDM~\cite{AdvDM}, Mist (0)~\cite{mist}, Mist (1)~\cite{mist}, AdvDM$_{800}^{900}$ (Ours). View this figure in color. Use zoom-in to check the texture.}
\end{figure*}

\newpage
\subsection{Attacking SD-Inpainting}
\begin{figure*}[!h]
	\centering
	\begin{minipage}{\linewidth}
		\centering
		\includegraphics[width=0.12\linewidth]{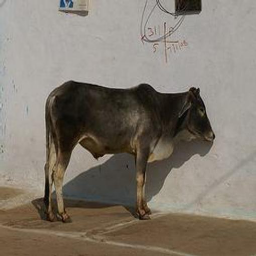}
		\includegraphics[width=0.12\linewidth]{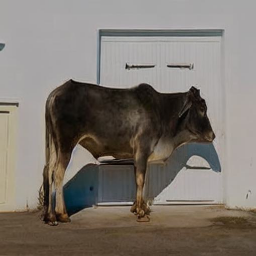}
		\includegraphics[width=0.12\linewidth]{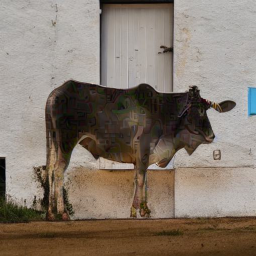}
		\includegraphics[width=0.12\linewidth]{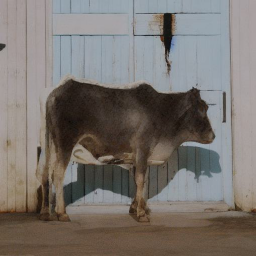}
		\includegraphics[width=0.12\linewidth]{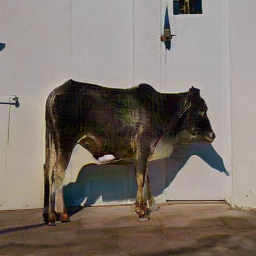}
		\includegraphics[width=0.12\linewidth]{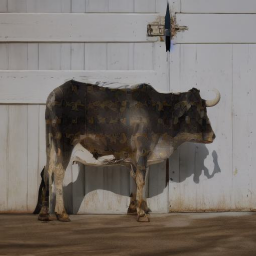}
		\includegraphics[width=0.12\linewidth]{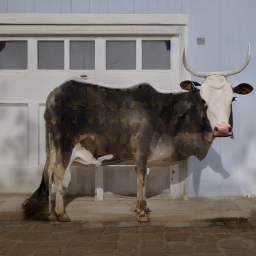}
		\includegraphics[width=0.12\linewidth]{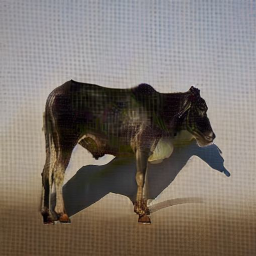}
	\end{minipage}\\
	``A white wall and bright blue door is where a cow is standing next to.''
	\begin{minipage}{\linewidth}
		\centering
		\includegraphics[width=0.12\linewidth]{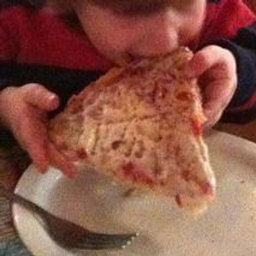}
		\includegraphics[width=0.12\linewidth]{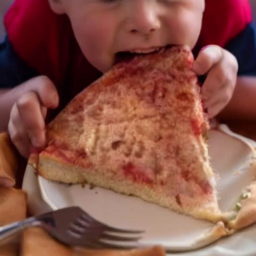}
		\includegraphics[width=0.12\linewidth]{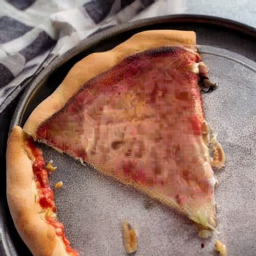}
		\includegraphics[width=0.12\linewidth]{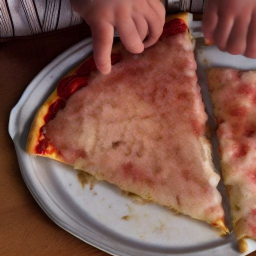}
		\includegraphics[width=0.12\linewidth]{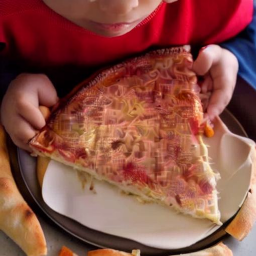}
		\includegraphics[width=0.12\linewidth]{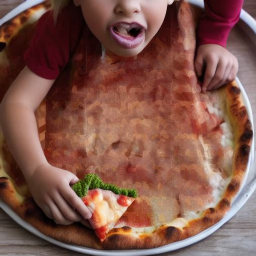}
		\includegraphics[width=0.12\linewidth]{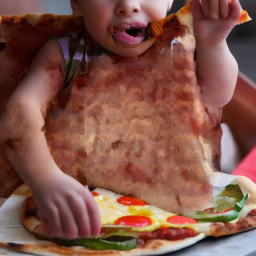}
		\includegraphics[width=0.12\linewidth]{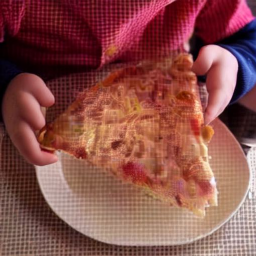}
	\end{minipage}\\
	``A kid chomping on a pizza slice resting on a plate.''
	\begin{minipage}{\linewidth}
		\centering
		\includegraphics[width=0.12\linewidth]{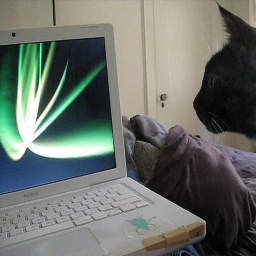}
		\includegraphics[width=0.12\linewidth]{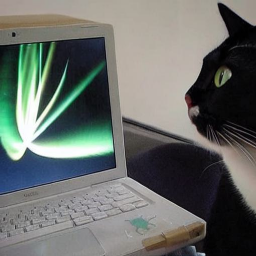}
		\includegraphics[width=0.12\linewidth]{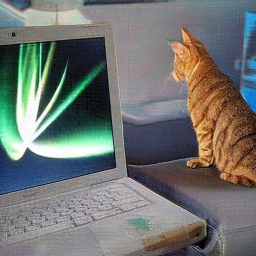}
		\includegraphics[width=0.12\linewidth]{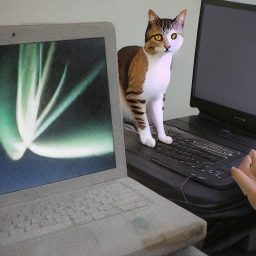}
		\includegraphics[width=0.12\linewidth]{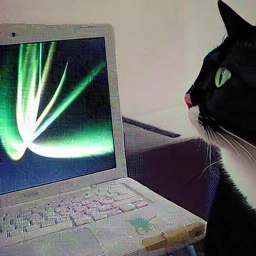}
		\includegraphics[width=0.12\linewidth]{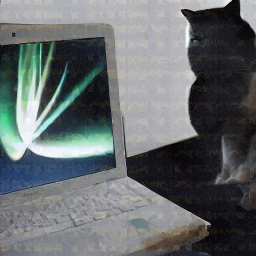}
		\includegraphics[width=0.12\linewidth]{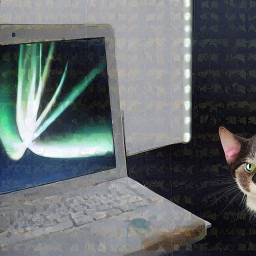}
		\includegraphics[width=0.12\linewidth]{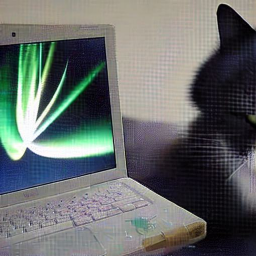}
	\end{minipage}\\
	``The screensaver on a laptop computer screen is being viewed by a cat.''
	\begin{minipage}{\linewidth}
		\centering
		\includegraphics[width=0.12\linewidth]{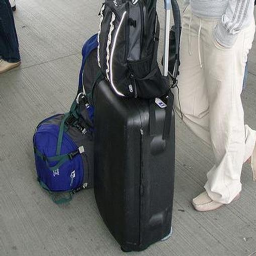}
		\includegraphics[width=0.12\linewidth]{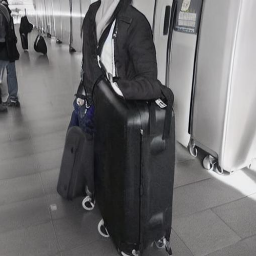}
		\includegraphics[width=0.12\linewidth]{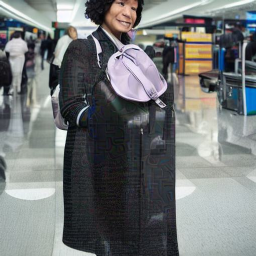}
		\includegraphics[width=0.12\linewidth]{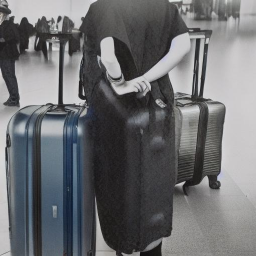}
		\includegraphics[width=0.12\linewidth]{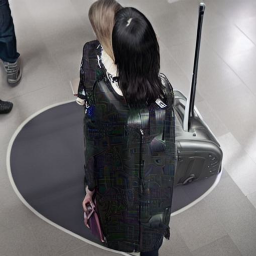}
		\includegraphics[width=0.12\linewidth]{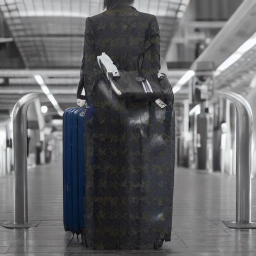}
		\includegraphics[width=0.12\linewidth]{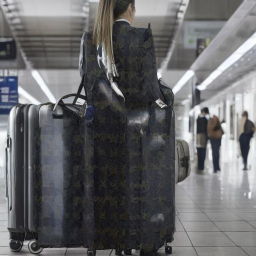}
		\includegraphics[width=0.12\linewidth]{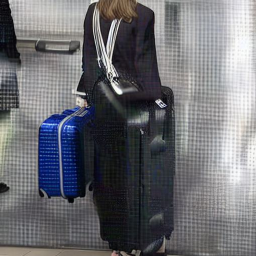}
	\end{minipage}\\
	``Next to her luggage, a woman stands on the platform.''
	\begin{minipage}{\linewidth}
		\centering
		\includegraphics[width=0.12\linewidth]{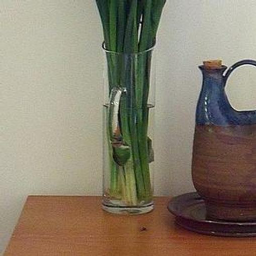}
		\includegraphics[width=0.12\linewidth]{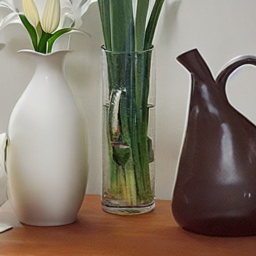}
		\includegraphics[width=0.12\linewidth]{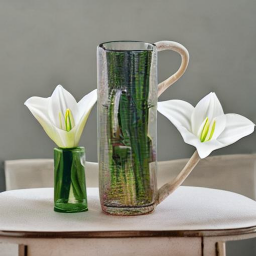}
		\includegraphics[width=0.12\linewidth]{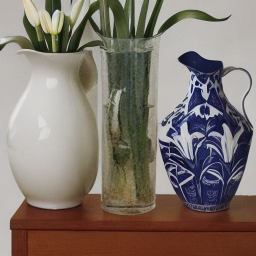}
		\includegraphics[width=0.12\linewidth]{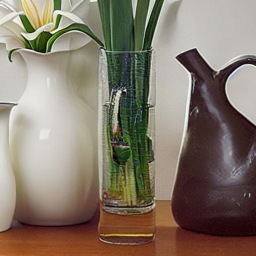}
		\includegraphics[width=0.12\linewidth]{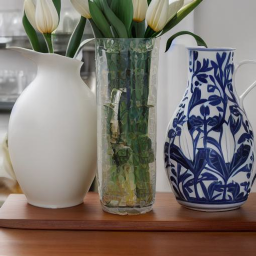}
		\includegraphics[width=0.12\linewidth]{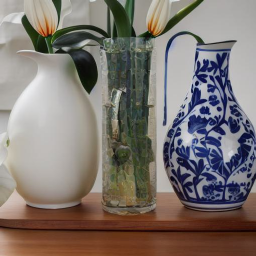}
		\includegraphics[width=0.12\linewidth]{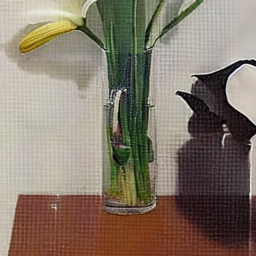}
	\end{minipage}\\
	``A vase with white lilies sitting next to a ceramic jug.''
	\begin{minipage}{\linewidth}
		\centering
		\includegraphics[width=0.12\linewidth]{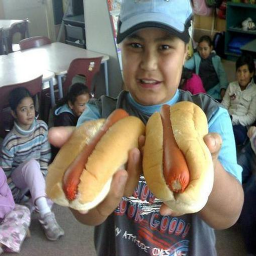}
		\includegraphics[width=0.12\linewidth]{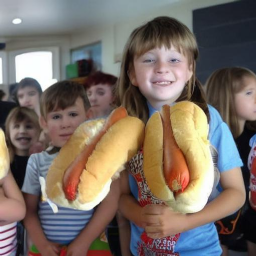}
		\includegraphics[width=0.12\linewidth]{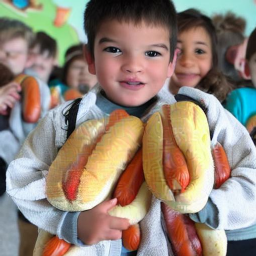}
		\includegraphics[width=0.12\linewidth]{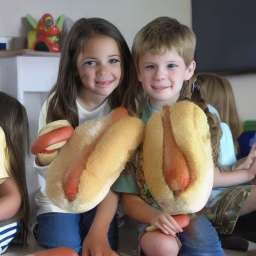}
		\includegraphics[width=0.12\linewidth]{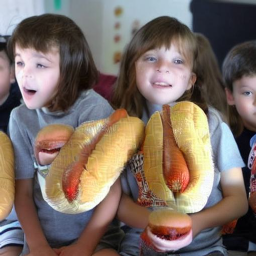}
		\includegraphics[width=0.12\linewidth]{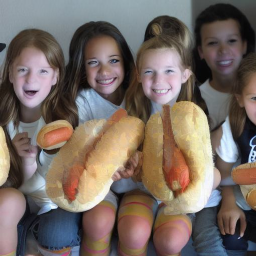}
		\includegraphics[width=0.12\linewidth]{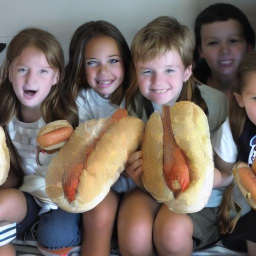}
		\includegraphics[width=0.12\linewidth]{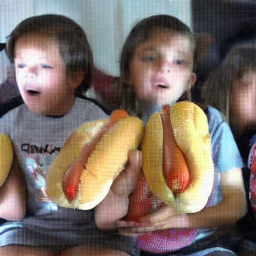}
	\end{minipage}\\
	``In a room of kids, a kid is holding two hot dogs. ''
	\begin{minipage}{\linewidth}
		\centering
		\includegraphics[width=0.12\linewidth]{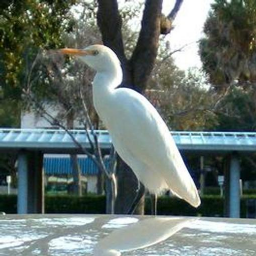}
		\includegraphics[width=0.12\linewidth]{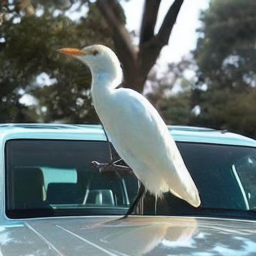}
		\includegraphics[width=0.12\linewidth]{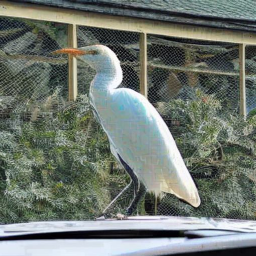}
		\includegraphics[width=0.12\linewidth]{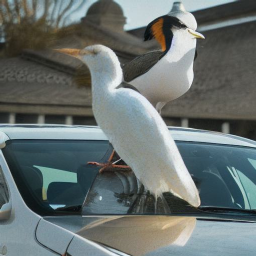}
		\includegraphics[width=0.12\linewidth]{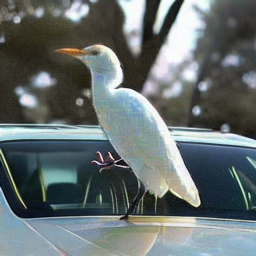}
		\includegraphics[width=0.12\linewidth]{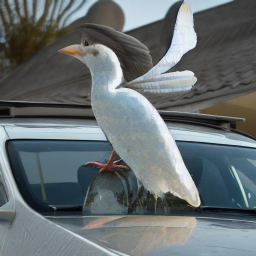}
		\includegraphics[width=0.12\linewidth]{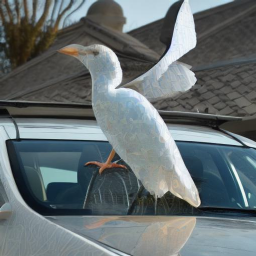}
		\includegraphics[width=0.12\linewidth]{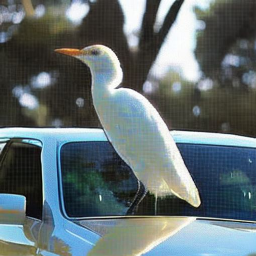}
	\end{minipage}\\
	``A bird standing on a car roof close to a covered pavilion.''
	\begin{minipage}{\linewidth}
		\centering
		\includegraphics[width=0.12\linewidth]{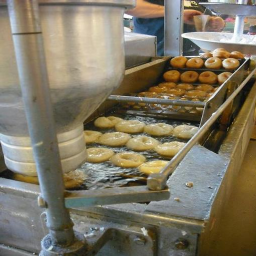}
		\includegraphics[width=0.12\linewidth]{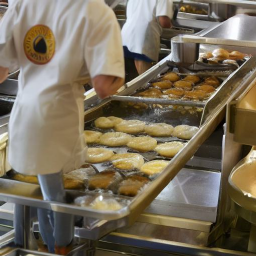}
		\includegraphics[width=0.12\linewidth]{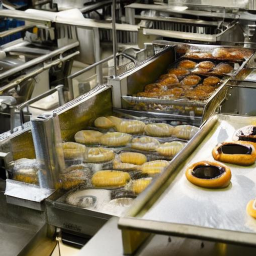}
		\includegraphics[width=0.12\linewidth]{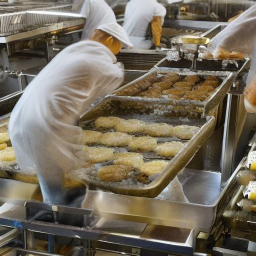}
		\includegraphics[width=0.12\linewidth]{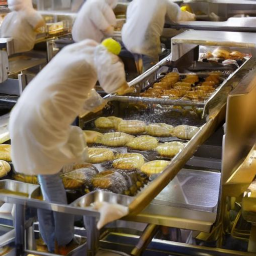}
		\includegraphics[width=0.12\linewidth]{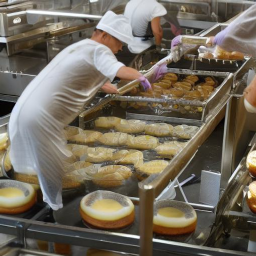}
		\includegraphics[width=0.12\linewidth]{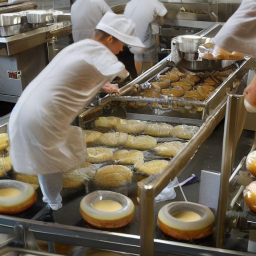}
		\includegraphics[width=0.12\linewidth]{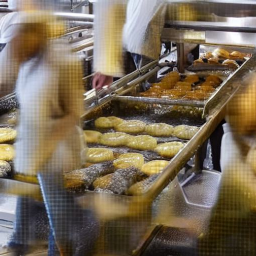}
	\end{minipage}\\
	``At the fryer, rows and rows of doughnuts move along the assembly line.''
	\caption{From left to right: Original images, Benign, Chain~\cite{benchmark}, SDS~\cite{sds}, AdvDM~\cite{AdvDM}, Mist (0)~\cite{mist}, Mist (1)~\cite{mist}, AdvDM$_{800}^{900}$ (Ours). View this figure in color. Use zoom-in to check the texture.}
\end{figure*}

\newpage
\subsection{Attacking SD-2-Inpainting}
\begin{figure*}[!h]
	\centering
	\begin{minipage}{\linewidth}
		\centering
		\includegraphics[width=0.12\linewidth]{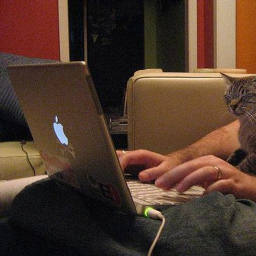}
		\includegraphics[width=0.12\linewidth]{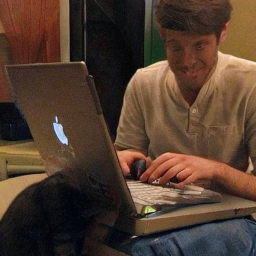}
		\includegraphics[width=0.12\linewidth]{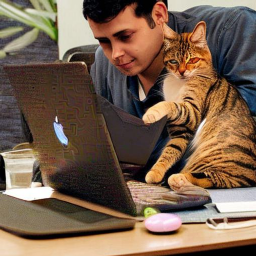}
		\includegraphics[width=0.12\linewidth]{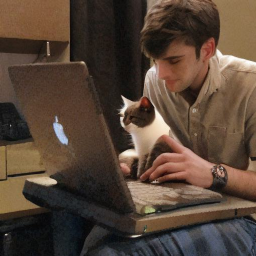}
		\includegraphics[width=0.12\linewidth]{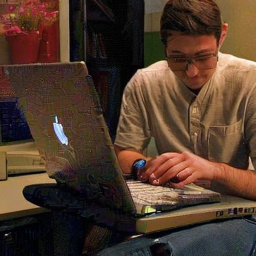}
		\includegraphics[width=0.12\linewidth]{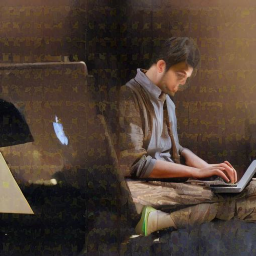}
		\includegraphics[width=0.12\linewidth]{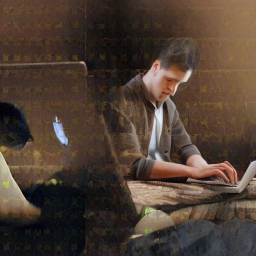}
		\includegraphics[width=0.12\linewidth]{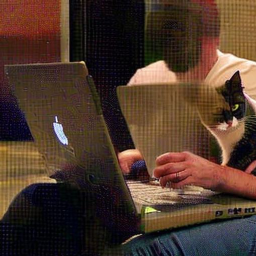}
	\end{minipage}\\
	``While typing on the laptop, the man has a cat sitting on his lap.''
	\begin{minipage}{\linewidth}
		\centering
		\includegraphics[width=0.12\linewidth]{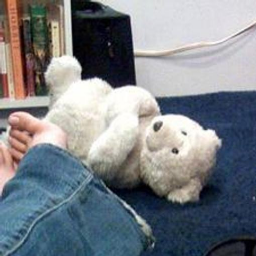}
		\includegraphics[width=0.12\linewidth]{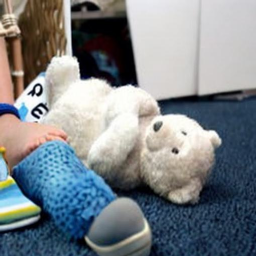}
		\includegraphics[width=0.12\linewidth]{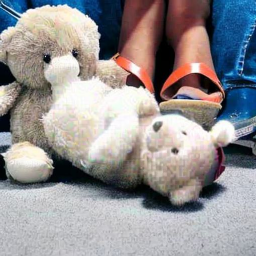}
		\includegraphics[width=0.12\linewidth]{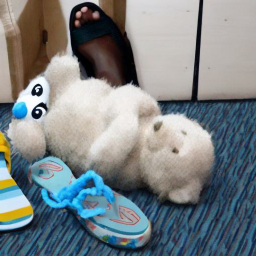}
		\includegraphics[width=0.12\linewidth]{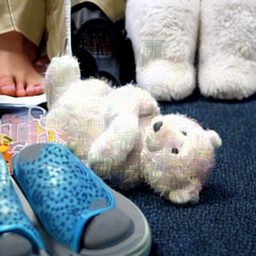}
		\includegraphics[width=0.12\linewidth]{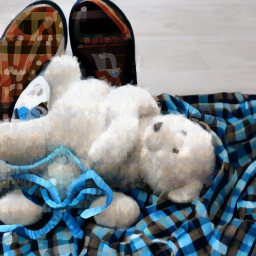}
		\includegraphics[width=0.12\linewidth]{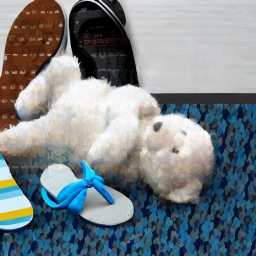}
		\includegraphics[width=0.12\linewidth]{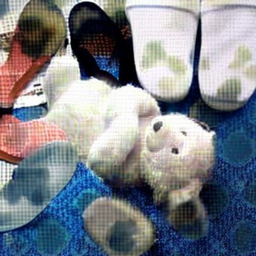}
	\end{minipage}\\
	``Sitting on the floor beside flip flops and a teddy bear is a person.''
	\begin{minipage}{\linewidth}
		\centering
		\includegraphics[width=0.12\linewidth]{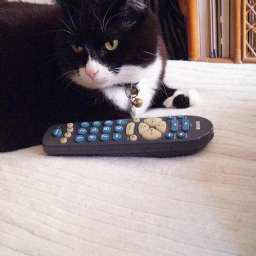}
		\includegraphics[width=0.12\linewidth]{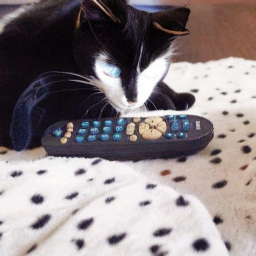}
		\includegraphics[width=0.12\linewidth]{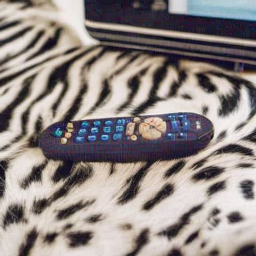}
		\includegraphics[width=0.12\linewidth]{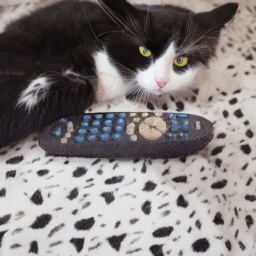}
		\includegraphics[width=0.12\linewidth]{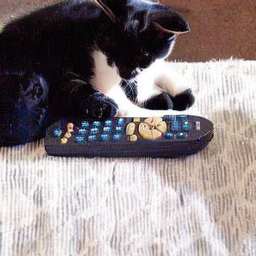}
		\includegraphics[width=0.12\linewidth]{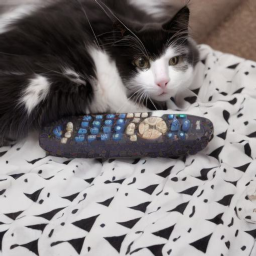}
		\includegraphics[width=0.12\linewidth]{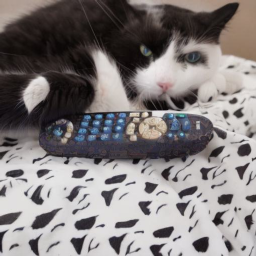}
		\includegraphics[width=0.12\linewidth]{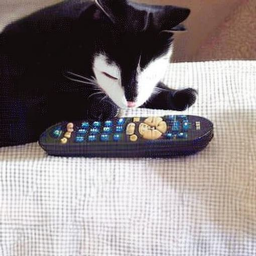}
	\end{minipage}\\
	``A remote control positioned beside a cat with black and white fur.''
	\begin{minipage}{\linewidth}
		\centering
		\includegraphics[width=0.12\linewidth]{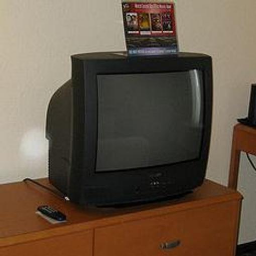}
		\includegraphics[width=0.12\linewidth]{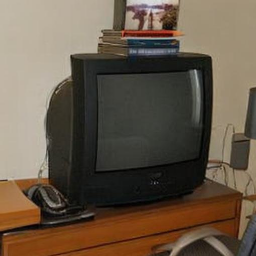}
		\includegraphics[width=0.12\linewidth]{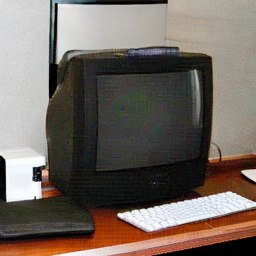}
		\includegraphics[width=0.12\linewidth]{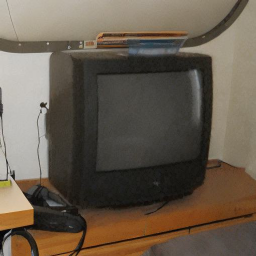}
		\includegraphics[width=0.12\linewidth]{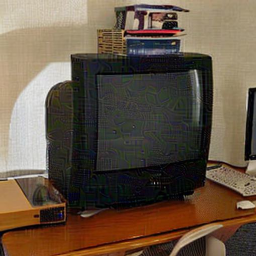}
		\includegraphics[width=0.12\linewidth]{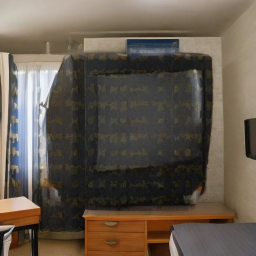}
		\includegraphics[width=0.12\linewidth]{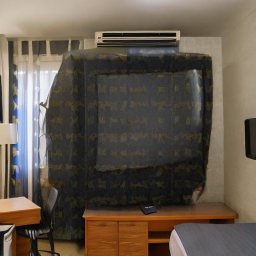}
		\includegraphics[width=0.12\linewidth]{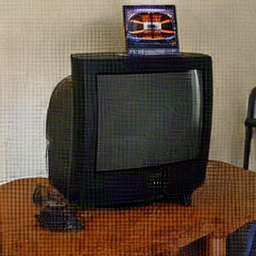}
	\end{minipage}\\
	``An accommodation with a diminutive television and a workstation.''
	\begin{minipage}{\linewidth}
		\centering
		\includegraphics[width=0.12\linewidth]{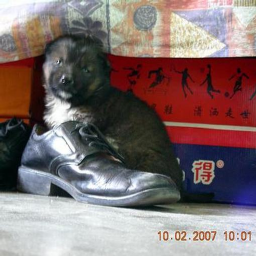}
		\includegraphics[width=0.12\linewidth]{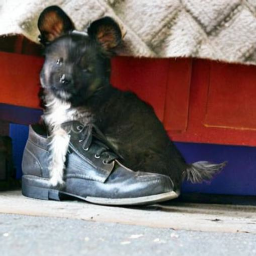}
		\includegraphics[width=0.12\linewidth]{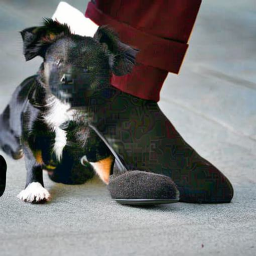}
		\includegraphics[width=0.12\linewidth]{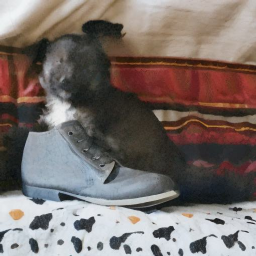}
		\includegraphics[width=0.12\linewidth]{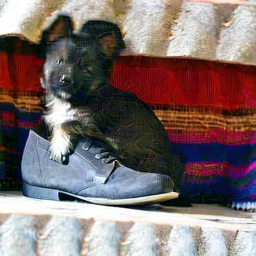}
		\includegraphics[width=0.12\linewidth]{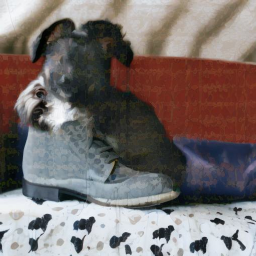}
		\includegraphics[width=0.12\linewidth]{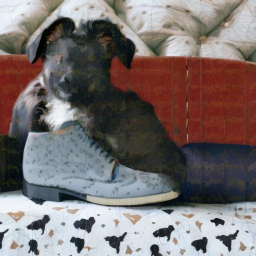}
		\includegraphics[width=0.12\linewidth]{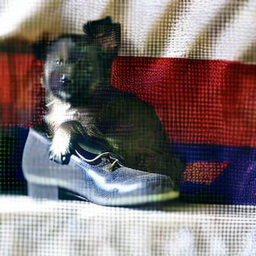}
	\end{minipage}\\
	``A pair of shoes is accompanied by a small black dog.''
	\begin{minipage}{\linewidth}
		\centering
		\includegraphics[width=0.12\linewidth]{./iiVisual/ori/000000411774.pdf}
		\includegraphics[width=0.12\linewidth]{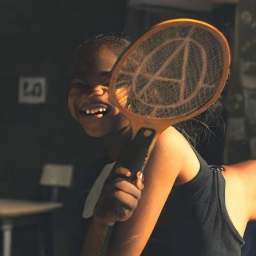}
		\includegraphics[width=0.12\linewidth]{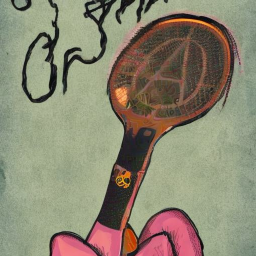}
		\includegraphics[width=0.12\linewidth]{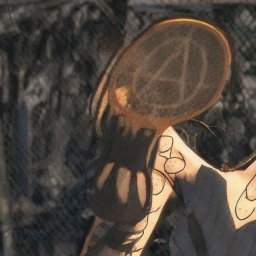}
		\includegraphics[width=0.12\linewidth]{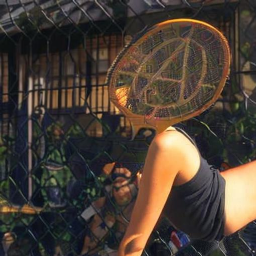}
		\includegraphics[width=0.12\linewidth]{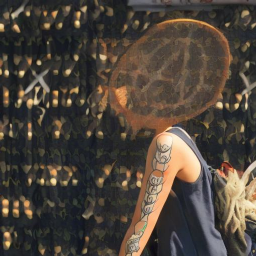}
		\includegraphics[width=0.12\linewidth]{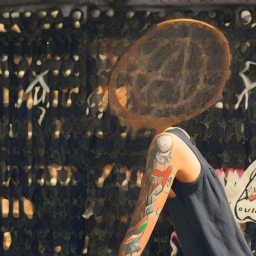}
		\includegraphics[width=0.12\linewidth]{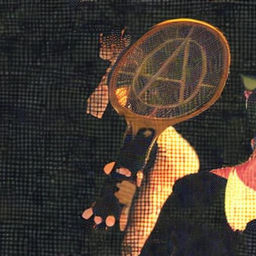}
	\end{minipage}\\
	``A racket-wielding girl grins, showing off her tattoo.''
	\begin{minipage}{\linewidth}
		\centering
		\includegraphics[width=0.12\linewidth]{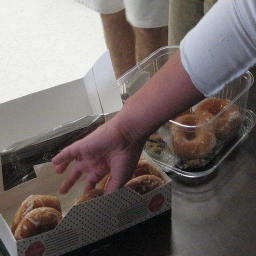}
		\includegraphics[width=0.12\linewidth]{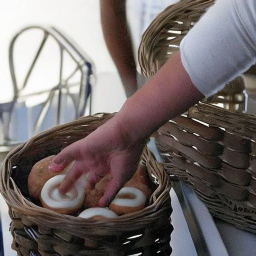}
		\includegraphics[width=0.12\linewidth]{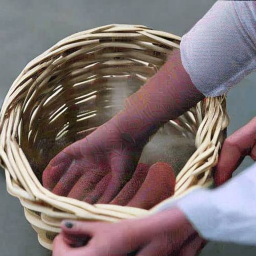}
		\includegraphics[width=0.12\linewidth]{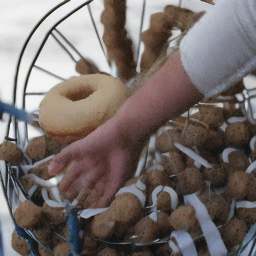}
		\includegraphics[width=0.12\linewidth]{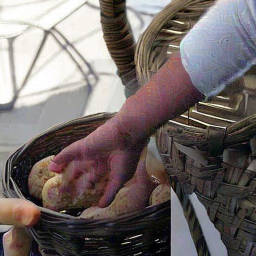}
		\includegraphics[width=0.12\linewidth]{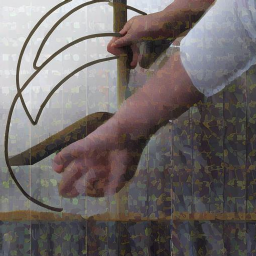}
		\includegraphics[width=0.12\linewidth]{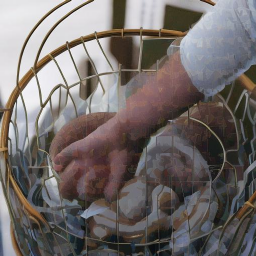}
		\includegraphics[width=0.12\linewidth]{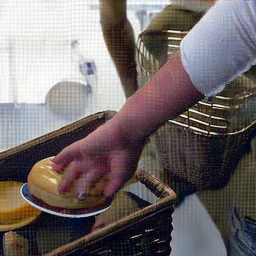}
	\end{minipage}\\
	``In the image, a person is reaching for a doughnut from a basket.''
	\begin{minipage}{\linewidth}
		\centering
		\includegraphics[width=0.12\linewidth]{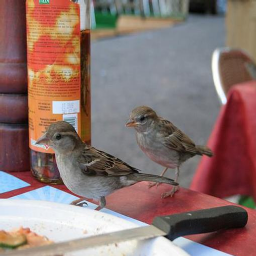}
		\includegraphics[width=0.12\linewidth]{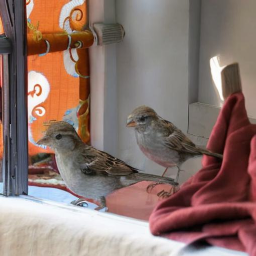}
		\includegraphics[width=0.12\linewidth]{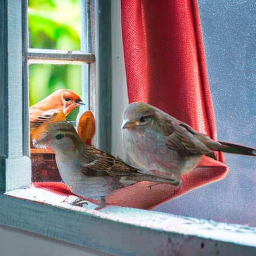}
		\includegraphics[width=0.12\linewidth]{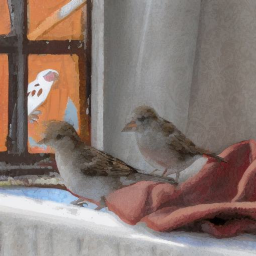}
		\includegraphics[width=0.12\linewidth]{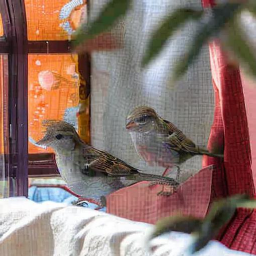}
		\includegraphics[width=0.12\linewidth]{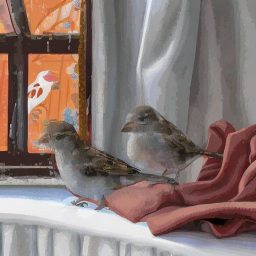}
		\includegraphics[width=0.12\linewidth]{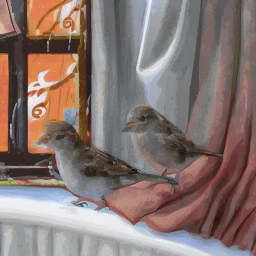}
		\includegraphics[width=0.12\linewidth]{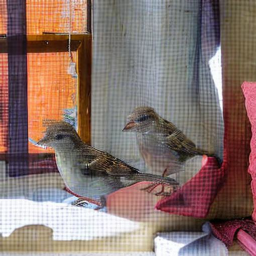}
	\end{minipage}\\
	``Two sparrows sit on a windowsill with a red curtain in a house.''
	\caption{From left to right: Original images, Benign, Chain~\cite{benchmark}, SDS~\cite{sds}, AdvDM~\cite{AdvDM}, Mist (0)~\cite{mist}, Mist (1)~\cite{mist}, AdvDM$_{800}^{900}$ (Ours). View this figure in color. Use zoom-in to check the texture.}
\end{figure*}

\end{document}